\documentclass[11pt]{article}

\usepackage{latexsym,multirow}
\usepackage{rotating,tabularx}
\usepackage{floatrow}
\usepackage{xr}
\usepackage{amssymb,amsmath, bm}
\usepackage{graphicx}
\usepackage[table,xcdraw]{xcolor}
\usepackage{booktabs,lscape}
\usepackage{enumerate}
\usepackage{capt-of}
\usepackage[ruled, lined, linesnumbered, commentsnumbered, longend]{algorithm2e}
\usepackage{algpseudocode}
\usepackage{setspace}
\usepackage{comment}

\usepackage{natbib}
\usepackage[colorlinks,allcolors=blue]{hyperref}
\usepackage{colortbl}
\usepackage{subcaption,cleveref}
\newfloatcommand{capbtabbox}{table}[][\FBwidth]
\usepackage{dcolumn}
\newcolumntype{.}{D{.}{.}{-1}}
\newcolumntype{d}[1]{D{.}{.}{#1}}
\usepackage{amsthm}
\usepackage{floatrow}

\theoremstyle{definition}

\newtheorem{definition}{Definition}

\newtheorem{theorem}{Theorem}

\newcommand{\ind}{\mbox{$\perp\!\!\!\perp$}}

\newcommand{\argmax}{\operatornamewithlimits{argmax}}

\usepackage{rotating}
\newcommand{\E}{\ensuremath{\mathbb{E}}}
\newcommand{\bx}{\bm{x}}
\newcommand{\bX}{\bm{X}}
\newcommand{\btheta}{\boldsymbol{\theta}}
\newcommand{\bTheta}{\boldsymbol{\Theta}}

\usepackage[compact]{titlesec}

\allowdisplaybreaks
\newcommand\spacingset[1]{\renewcommand{\baselinestretch}%
  {#1}\small\normalsize}

\title{Bayesian Safe Policy Learning with Chance Constrained
  Optimization: Application to Military Security Assessment during the
  Vietnam War\thanks{We acknowledge the partial support from Cisco
    Systems, Inc. (CG\# 2370386), National Science Foundation
    (SES--2051196), and Sloan Foundation (Economics Program; 2020--13946).}}

\author{Zeyang Jia\thanks{Ph.D. Student, Department of Statistics, Harvard University. 1 Oxford Street, Cambridge MA 02138.Email: \href{mailto:zeyangjia@fas.harvard.edu}{zeyangjia@fas.harvard.edu}} \and Eli Ben-Michael\thanks{Assistant Professor, Department of Statistics \& Data Science and Heinz College of Information Systems \& Public Policy, Carnegie Mellon University. 4800 Forbes Avenue, Hamburg Hall, Pittsburgh PA 15213.  Email: \href{mailto:ebenmichael@cmu.edu}{ebenmichael@cmu.edu} URL:
\href{https://ebenmichael.github.io}{ebenmichael.github.io}} \and Kosuke Imai\thanks{Professor,
      Department of Government and Department of Statistics, Harvard
      University.  1737 Cambridge Street, Institute for Quantitative
      Social Science, Cambridge MA 02138.  Email:
      \href{mailto:imai@harvard.edu}{imai@harvard.edu} URL:
      \href{https://imai.fas.harvard.edu}{https://imai.fas.harvard.edu}}}

\begin{document}

\maketitle
\begin{abstract}
  Algorithmic decisions and recommendations are used in many
  high-stakes decision-making settings such as criminal justice,
  medicine, and public policy. We investigate whether it would have
  been possible to improve a security assessment algorithm employed
  during the Vietnam War, using outcomes measured immediately after
  its introduction in late 1969. This empirical application raises
  several methodological challenges that frequently arise in
  high-stakes algorithmic decision-making. First, before implementing
  a new algorithm, it is essential to characterize and control the
  risk of yielding worse outcomes than the existing algorithm. Second,
  the existing algorithm is deterministic, and learning a new
  algorithm requires transparent extrapolation. Third, the existing
  algorithm involves discrete decision tables that are difficult to
  optimize over.

  To address these challenges, we introduce the Average Conditional
  Risk (ACRisk), which first quantifies the risk that a new
  algorithmic policy leads to worse outcomes for subgroups of
  individual units and then averages this risk over the distribution
  of subgroups. We also propose a Bayesian policy learning framework
  that maximizes the posterior expected value while controlling the
  posterior expected ACRisk. This framework separates the estimation
  of heterogeneous treatment effects from policy optimization,
  enabling flexible estimation of effects and optimization over
  complex policy classes. We characterize the resulting
  chance-constrained optimization problem as a constrained linear
  programming problem. Our analysis shows that compared to the actual
  algorithm used during the Vietnam War, the learned algorithm
  assesses most regions as more secure and emphasizes economic and
  political factors over military factors.

  \bigskip

  
\end{abstract}

\clearpage
\spacingset{1.375}

\section{Introduction}
 
Algorithmic decisions and recommendations have long been used in areas
as diverse as credit markets \citep{Lauer2017_credit} and war
\citep{Daddis2012_vietnam}.  They are now increasingly integral to
many aspects of today's society, including online advertising
\citep[e.g.,][]{li2010contextual,tang2013automatic,schwartz2017customer},
medicine \citep[e.g.,][]{kamath2001model,nahum2018just}, and criminal
justice \citep[e.g.,][]{imai2020experimental,greiner2020randomized}.
A primary challenge when applying data-driven policies to
consequential decision-making is to characterize and control the risk
associated with any new policies learned from the data.  Stakeholders
in fields such as medicine, public policy, and the military may be
concerned that the adoption of new data-derived policies could
inadvertently lead to worse outcomes for some individuals in certain
settings.

In this paper, we consider a particularly high-stakes setting,
analyzing a United States (US) military security assessment policy
that saw active use in the Vietnam War.  During the war, the US
military developed a data-driven scoring system called the Hamlet
Evaluation System (HES) to produce a security score for each region
\citep{CHECO}; commanders used these scores to make air strike
decisions.  A recent analysis based on a regression discontinuity
design shows that the airstrikes had significantly negative effects on
development outcomes including regional safety, economic, and civic
society measures, and so were broadly counter-productive
\citep{dell2018nation}.

We consider whether it would have been possible to improve the HES
using contemporaneous data collected by the US military and related
agencies to optimize for various development outcomes, while
practically controlling the risk of worsening these outcomes in too
many regions.  In particular, the original HES was composed of various
``sub-model scores'' that measured different aspects of each region
(e.g., economic variables, local administration, enemy military
presence) based on survey responses.  It then combined these into a
single security score through a three-level hierarchical aggregation
using pre-defined decision tables.  The security scores were presented
to Air Force commanders, who used them to make targeting decisions.

We focus on modifying these underlying decision tables while
maintaining the transparency and interpretability of the original HES.
Even though the system was in use over a half-century ago, this
analysis serves three purposes.  First, we show how to address
methodological problems that are also common in other algorithmic
decisions and recommendations.  Second, because we limit to using data
that were available at the time, our analysis serves as a case study
on the development of optimal algorithm-assisted decision-making tools
in high-stakes settings.  Lastly, investigating the potential
improvement gives insight into ways, in which the HES fell short,
providing a historical lesson.

This empirical analysis poses several methodological challenges that
are commonly encountered in high-stakes data-driven decision-making
settings.  First, we want to characterize and control the risk that a
new learned decision, classification, or recommendation policy may
lead to worse outcomes for some regions.  Second, the HES is a
deterministic function of the input data, implying that extrapolation
is necessary to learn new policies. Third, the security score is
produced via a series of aggregations using decision tables, which is
discrete and difficult to optimize over. Indeed, such decision tables
are widely used in many public policy and medical decision-making
settings \citep[e.g., risk scores in the US criminal justice
system][]{greiner2020randomized,imai2020experimental}.

To address these challenges, we introduce a risk measure, the Average
Conditional Risk (ACRisk), that first quantifies the risk of a given
policy for groups of individual units with a specified set of
covariates and then averages this conditional risk over the
distribution of the covariates.  In contrast to existing risk measures
that characterize the uncertainty around the average performance of
the policy
\citep[e.g.,][]{delage2010percentile,vakili2015mean,bai2022pessimistic},
the ACRisk measures the extent to which a learned policy negatively
affects subgroups.  This allows us to better characterize the
potential heterogeneous risks of applying a new policy.  With this
risk measure in hand, we propose a Bayesian safe policy learning
framework that maximizes the posterior expected value given the
observed data while controlling the posterior expected ACRisk.  We
formulate this as a chance-constrained optimization problem and show
how to convert it into a linear constraint, which can be solved as a
standard optimization problem.

The primary advantages of the proposed framework are its flexibility
and practical applicability.  Because the chance-constrained
optimization problem only relies on posterior samples, one can use
popular Bayesian nonparametric regression models such as BART and
Gaussian Process regression
\citep{rasmussen2003gaussian,chipman2010bart}, while efficiently
finding an optimal policy within a complex policy class.  This is
especially helpful in settings such as ours with limited or no 
covariate overlap, where our framework allows for flexible
extrapolation through the Bayesian prior.  In contrast, frequentist
notions of safe policy learning rely on robust optimization and
require solving a minimax optimization problem over both the class of
potential models and the class of potential policies, making it
difficult to consider nonparametric models and complex policy classes
at the same time \citep{pu2020estimating,
  kallus2021minimax,ben2022policy,zhang2022safe}.

We show through simulation studies that 
controlling the posterior expected ACRisk effectively limits the
ACRrisk across various scenarios, reducing the risk of harming certain subgroups of units. We also
find that although the proposed methodology is designed to be
conservative, under some settings with a low signal-to-noise ratio, it
yields a new policy whose average value is higher than a policy
learned without the safety constraint.  This is evidence that
the proposed safety constraint can effectively regularize the policy optimization problem.

We apply the proposed methodology to find adaptations to the HES that
lead to better outcomes --- as measured by military, economic, and
social objectives --- while limiting the posterior probability that
some regions experience worse outcomes under the new system than under
the original HES.  We consider two policy learning problems --- one
where we only change the last layer of the hierarchical aggregation
that combines military, political and social economic sub-model
scores, and another where we modify all of the decision tables used in
the three-level hierarchical aggregation at the same time.  To deal
with the latter complex case, we develop a stochastic optimization
algorithm, based on random walks on partitions of directed acyclic
graphs, that is generally applicable to decision tables.

Our analysis consistently shows that the original HES is too
pessimistic---assessing regions as too insecure---and places too much
emphasis on military factors.  In contrast, our data-derived
adaptations to the HES assess regions as more secure and rely more on
economic and social factors to produce regional security scores.

\paragraph{Related literature.}

Recent years have witnessed a growing interest among statisticians and
machine learning researchers in finding optimal policies from
randomized experiments and observational studies
\citep[e.g.][]{beygelzimer2009offset,qian2011performance,dudik2011doubly,zhao2012estimating,zhang2012estimating,swaminathan2015batch,luedtke2016statistical,zhou2017residual,kitagawa2018should,kallus2018balanced,athey2021policy,zhou2022offline}.
These works typically consider the following two steps under a
frequentist framework --- first characterize the average performance,
or value, of a given policy via the CATE, and then learn an optimal
policy by maximizing the estimated value based on the observed data.

In contrast, we adopt a Bayesian perspective --- first obtain the
posterior distribution of the CATE given the observed data, and then
learn an optimal policy by maximizing the posterior expected value.
Bayesian methods have been widely used for causal inference
\citep[see][for a recent review]{li2022bayesian}.  In particular, BART
and Gaussian processes are often used to flexibly estimate the CATE
\citep{hill2011bayesian, branson2019nonparametric,
  taddy2016nonparametric,hahn2020bayesian}.  However, it appears that
the Bayesian approach has been rarely applied to policy learning \citep{kasy2018optimal}.  Our
proposed framework takes advantage of these popular Bayesian
nonparametric methods for safe policy learning.

There is also a growing literature on policy learning
in scenarios where the CATE is unidentifiable \citep{manski2007minimax,pu2020estimating,yata2021optimal,kallus2021minimax,ben2021safe,zhang2022safe,ben2022policy}.  These include
observational studies with unmeasured confounders
\citep{kallus2021minimax}, studies with noncompliance or 
an instrumental variable \citep{pu2020estimating}, studies that lack overlap
due to deterministic  treatment rules
\citep{ben2021safe,zhang2022safe}, and utility functions that involve the joint
set of potential outcomes \citep{ben2022policy}.  These works first
partially identify the value of a given policy then
find the policy that maximizes the worst-case value using robust optimization.
Our approach differs in that we decouple estimation and policy optimization by 
only relying on posterior samples for policy learning.

In the Reinforcement Learning (RL) literature, various notions of
safety have been studied (e.g., safe reinforcement learning, risk
averse reinforcement learning, pessimistic reinforcement learning; see
\cite{garcia2015comprehensive}).  For example, \cite{geibel2005risk}
control the risk of the agent visiting a ``dangerous state'' by
explicitly imposing a risk constraint when finding an optimal policy.
In contrast, \cite{sato2001td} and \cite{vakili2015mean} use the
variance of the return as a penalty term in the objective when finding
an optimal policy with a high expected return and low variance.  While
this RL literature focuses on online settings where the algorithm is
designed to avoid risks during exploration, we study the risk of
applying data-driven policies in offline settings.

We also extend existing work by developing the notion of ACRisk and using it
as a constraint in optimizing the posterior expected value of a new
policy.  A related literature is \textit{pessimistic offline RL}, which
quantifies the risk of a given policy using the lower confidence bound
(LCB) of the value, and finds a policy that has the best LCB
\citep{jin2021pessimism,buckman2020importance,zanette2021provable,xie2021bellman,chen2022offline,rashidinejad2021bridging,yin2021towards,shi2022pessimistic,yan2022efficacy,uehara2021pessimistic,bai2022pessimistic,jin2022policy}.
In contrast, the proposed ACRisk measures the extent to which a new
policy negatively affects some groups of individuals when compared to
the baseline policy.

Finally, our work is also related to chance constrained optimization,
which is widely used in the analysis of decision making under
uncertainty \citep[e.g.,][]{schwarm1999chance,filar1995percentile,delage2007percentile,delage2010percentile,FARINA201653}.
For example, \citet{delage2010percentile} consider
chance constrained control for Markov Decision Processes. They assume
a Gaussian model for the reward distribution and use chance
constrained optimization to find a policy that achieves low regret
with high posterior probability.  In contrast, our method considers a
more general setup beyond the Gaussian model and uses the posterior
expected value of the ACRisk as a constraint.

\paragraph{Outline of the paper.}
The remainder of this paper is organized as follows.
Section~\ref{sec:hes} describes U.S. military security assessment in the Vietnam War, the HES, and the related empirical policy learning problem.
Section~\ref{sec:setup} introduces a formal setup, and 
Section~\ref{sec:framework} describes the Bayesian safe policy learning framework and the chance-constrained optimization procedure, as well as implementation via Gaussian Processes and Bayesian Causal Forests.
Section~\ref{sec:numerical} presents numerical experiments evaluating our proposal.
Section~\ref{sec:application} applies the Bayesian safe policy learning method to the Military security assessment problem.
Section~\ref{sec:discussion} concludes and discusses limitations and future directions.

\section{Military Security Assessment during the Vietnam War}
\label{sec:hes}

During the Vietnam War, the United States Air Force (USAF) conducted
numerous air strike campaigns.  One factor guiding USAF commanders in
making targeting decisions was a data-driven scoring system called the
Hamlet Evaluation System (HES) whose goal was to provide a metric for
regional security based on survey data \citep{CHECO}. We briefly
describe the HES and its aggregation rules, as well as the policy
learning problem we consider.  Specifically, we focus on the second
version of the HES, which was developed in 1969 and was originally
analyzed by \cite{dell2018nation}.

\subsection{The Hamlet Evaluation System}

The HES was based on a total of 169 military, political, and
socioeconomic indicators collected quarterly by Civil Operations and
Revolutionary Development Support (CORDS), a joint civilian-military
agency. These indicators were then categorized and summarized as 20
continuous ``sub-model'' scores.\footnote{The HES actually produces 19
  different sub-model scores from the 169 indicators, but one
  sub-model score is used twice in the later aggregation. For
  simplicity and clarity, consider these to be 20 scores.}  Each
sub-model score ranges from 1 to 5 and measures a different aspect of
a given region, such as \textit{Enemy Military Activity},
\textit{Economic Activity}, and \textit{Public Health}.  The HES
aggregates these 20 continuous-valued sub-model scores into a single
integer-valued security score ranging between 1 and 5.
The sub-model scores are all semantically ordered so
that lower values indicate that a region is ``worse'' in the metric.
The final HES score is ordered so that regions with a score of 1
should get the highest priority from USAF commanders and those with a
score of 5 should get the lowest.
After calculating the security score for each
region at the beginning of a quarter, USAF commanders used these
scores to make air strike decisions during that quarter.

Although several other factors contributed to the determination of
eventual air strike targets, \cite{dell2018nation} show that regions
which fell just above a security score threshold --- and hence had a
higher security score --- were less likely to be subject to
air strikes than those regions just below the threshold.  Using a
regression discontinuity design, the authors find that the air strike
campaign was largely counter-productive, increasing insurgency
activities and decreasing civic engagements.

Given this finding, we investigate whether it is possible to improve
the original HES security evaluation with the contemporaneous data
available during the war.  We focus on the first quarter of the HES
deployment (Q4 1969) and use the military impact assessments taken at
the beginning of the succeeding quarter (January 1970) as the outcome
to measure the efficacy of the HES.

\subsection{Aggregation via Decision Tables}

\begin{figure}[t!]
  \centering \spacingset{1}
  \includegraphics[width=\textwidth]{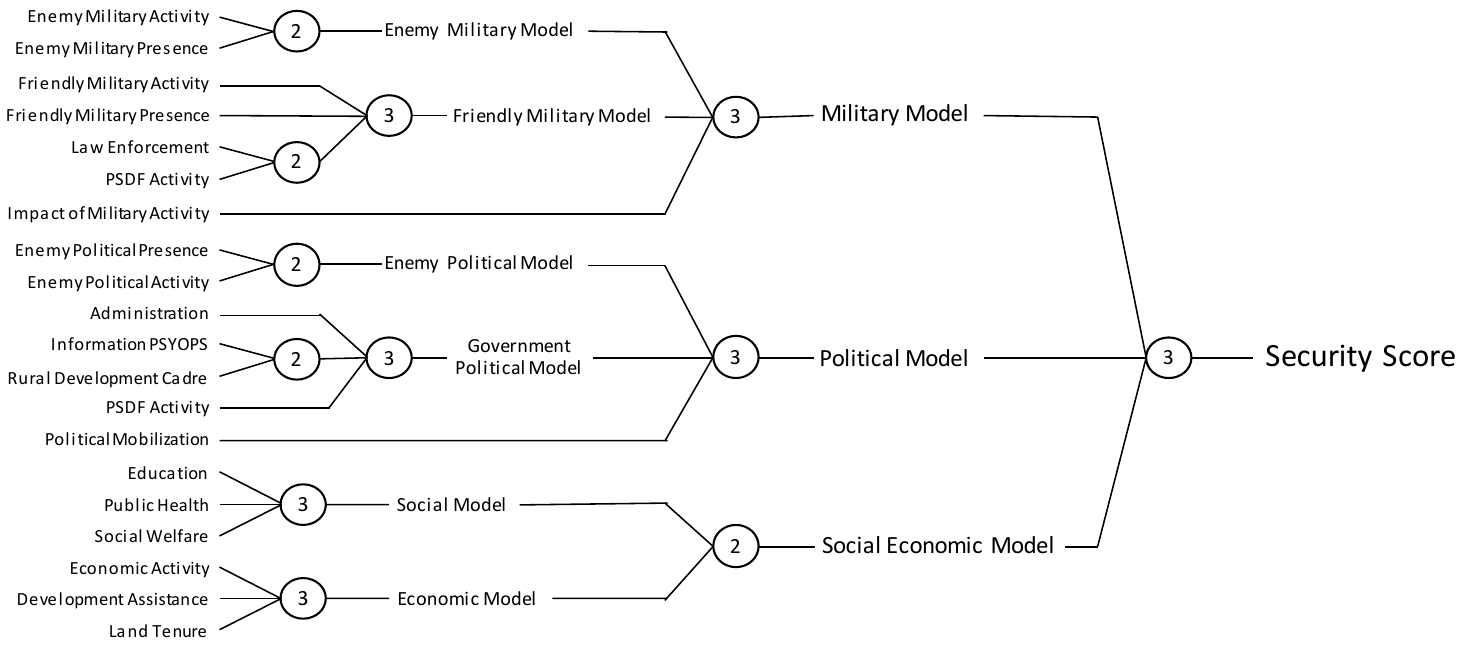}
  \caption{Aggregation of 20 sub-model scores. The Hamlet Evaluation
    System (HES) uses 20 sub-model scores as inputs, and aggregates them using
    two-way and three-way decision tables. Each circle corresponds to one aggregation based on the two-way or three-way decision table, and the decision tables used in different circles are the same. }
  \label{fig:hes_agg}
\end{figure}

The HES aggregates 20 continuous sub-model scores to a single
integer-valued security score that take a value in $\{1,2,3,4,5\}$, as
shown in Figure~\ref{fig:hes_agg}.  This hierarchical aggregation
process starts by rounding the 20 continuous sub-model scores to the
nearest integer in $\{1,2,3,4,5\}$.  The security score then is
produced through the following three steps.  First, 18 out of these 20
rounded sub-model scores are aggregated to six integer-valued model
scores relating to Enemy Military, Friendly Military, Enemy Political,
Government Political, Social, and Economic model scores.  Second, the
resulting six model scores, along with two remaining inputs (impact of
military activity and political mobilization), are aggregated to three
higher tier integer-valued scores --- Military, Political, and Social
Economic model scores.  Finally, these three scores are further
aggregated to the integer-valued single security score.

At each step of aggregation, the HES uses a two-way or three-way
decision table indicated by the number shown in each circle of
Figure~\ref{fig:hes_agg}.  These tables take two or three
integer-valued scores, ranging from 1 to 5, and return a new score,
which takes a value in $\{1,2,3,4,5\}$.
Figures~\ref*{fig:hes_table}~and~\ref*{fig:3waytable} in the appendix
show the two-way and three-way decision tables used in the HES,
respectively. For example, the two-way decision table maps an input of
$(2,3)$ to an output of $2$.  In the HES, the same two-way and
three-way tables are used across all levels of aggregation.

In our analysis, we modify the underlying two-way and three-way
decision tables.  We do not modify the hierarchical structure itself
or using black-box machine learning tools, because the hierarchical
structure contains substantial domain knowledge, and we wish to retain
the transparency of the original system.  This also facilitates easier
comparisons between our data-derived modifications and the original
HES system.  As we discuss and explore in
Section~\ref{sec:three_layer}, altering the underlying decision tables
alters the implicit influence of each of the 20 factors in
constructing the overall HES score.  We note, however, that our
methodological development in the next sections can also be used to
adjust the hierarchy itself or to use black box methods, although
there may be additional computational considerations.

\subsection{Evidence-based Policy Learning}

To measure the impact of air strikes and and learn a new decision
policy, we analyze three binary outcomes that reflect regional
security, economy, and civic society. These outcomes are based on
survey responses from multiple sources collected in January 1970,
after the period of air strikes in the final quarter of 1969
\citep{dell2018nation}.  During this period, USAF commanders targeted
1024 of the 1954 regions defined by the HES.\footnote{See Appendix
  Table~\ref{tab:summary} and Figure~\ref{fig:mean_submodel_scores}
  for other summary statistics as well as the average values of the
  security score and various sub-model scores} Our objective is to
develop a new decision rule that generates appropriate security scores
based on the 20 continuous sub-model scores.  It is worth noting that
our decision variable is the output security score from the HES rather
than the air strike decision itself, which is made by commanders. We
aim to optimize the way, in which the final security score is produced
from the 20 sub-model scores to achieve the best overall outcomes in
regional security, economy, and civic society (marginalizing over
commanders' airstrike decisions).

This application yields several methodological challenges. First, the
stakes of this decision are high and there is potential heterogeneity
across regions.  This motivates a safe policy learning approach that
limits the possibility of generating worse outcomes under a new,
learned policy for some regions than under the existing policy.
Second, the existing policy is deterministic, rather than stochastic.
Therefore, due to the lack of overlap between regions that received
different HES scores, we must extrapolate when estimating the
potential outcomes under a new policy.  Lastly, the existing policy is
based on multiple decision tables.  Thus, to keep the existing
structure of the policy, we must solve a complex optimization problem.
We now develop a new Bayesian safe policy framework that addresses
these challenges.

\section{Preliminaries}
\label{sec:setup}

Before we develop our Bayesian policy learning framework, we describe
the problem setup and notation.  In addition, we provide a brief
review of Bayesian policy learning.

\subsection{Setup and Notation}

We consider an individualized treatment rule or policy, which
deterministically maps a set of $p$ pre-treatment covariates
$\bX\in \mathcal{X}\subseteq\mathbb{R}^p$ to a multivalued decision
$D \in \mathcal{D}=\{0,1,\ldots,K-1\}$ where $K$ denotes the total
number of decision categories.  Formally, a policy is defined as a
function $\delta: \mathcal{X} \to \mathcal{D}$.  We have a simple
random sample of $n$ observations from the population of interest
$\mathcal{P}$, which is characterized by the joint distribution of
$\{\bX, Y(0), Y(1),\ldots,Y(K-1)\}$ where $Y(k)$ represents a
potential outcome under decision $D=k$ with $k \in \mathcal{D}$ and
the observed outcome is given by $Y_i = Y_i(D_i)$ \citep{neyman1923,
  rubin1974}.  This notation implicitly assumes no interference
between units \citep{rubin1980randomization}.

We also assume that the decision is unconfounded given the covariates,
i.e., $\left\{Y_i(k)\right\}_{k=0}^{K-1} \ind D_i\mid \bX_i$
\citep{rosenbaum1983central}.  However, we do not assume that there
exists covariate overlap for treated and controlled units --- i.e.,
$0 < \Pr(D_i = k \mid \bX_i = \bx) < 1$ for all $k \in \mathcal{D}$
and $\bx \in \mathcal{X}$.  The lack of overlap is motivated by the
fact that in many settings, the data are collected under an existing
deterministic policy \citep{ben2021safe,zhang2022safe}.

Suppose that a policy-maker defines a utility function
$u:\{0,1,...,K-1\} \times \mathcal{Y} \rightarrow \mathbb{R}$, which
maps every decision-outcome pair to a real-valued utility.  Then, the
expected utility or value of policy $\delta$ in a policy class $\Delta$
is given by,
\begin{equation}
  V(\delta)\ := \ \mathbb{E}\left[u(\delta(\bX), Y(\delta(\bX)))\right]. 
\end{equation}
Our goal is to find an optimal policy within a policy class $\Delta$
that has a high value relative to the existing policy $\tilde\delta$,
which also belongs to the same policy class $\Delta$
\citep{ben2021safe,wei2022safety}. In our application, the baseline
policy is a deterministic rule that produces the security score in the
HES.  Our goal is to derive a new policy whose average value is higher
than that of the original policy subject to a safety constraint that
will be introduced in Section~\ref{sec:framework}.

\subsection{Bayesian Policy Learning}

We next briefly introduce the basic elements of Bayesian policy
learning \citep[see][for reviews of Bayesian causal
inference]{ding2018causal,li2022bayesian}.  Under the Bayesian
framework, all parameters are regarded as random variables.  Following
the convention, we will use upper and lower case characters to
represent parameters and their realizations, respectively.

Suppose that the marginal distribution of potential outcomes is
characterized by a possibly infinite dimensional parameter $\bTheta$.
Bayesian policy learning proceeds in two steps.  First, we specify a
prior distribution $\pi(\bTheta)$ on the parameter $\bTheta$ and
compute its posterior distribution using Bayes' rule,
\begin{align}
  \pi(\bTheta\mid \{\bX_i,D_i, Y_i\}_{i=1}^n)
   \ \propto \ & \prod_{i=1}^n p(Y_i\mid \bTheta,\bX_i,D_i)\pi(\bTheta).
\end{align}

Under Bayesian policy learning, given the posterior distribution of
$\bTheta$, the optimal policy maximizes the posterior expected value,
\begin{equation}
  \delta_{opt} \ = \ \argmax_{\delta \in \Delta}\
  \mathbb{E}\left[V(\delta;\bTheta)\mid \{D_i, \bX_i,
    Y_i\}_{i=1}^n\right], \label{eq:bayesoptimal} 
\end{equation}
where
$V(\delta;\bTheta):= \mathbb{E}\left[u(\delta(\bX),Y(\delta(\bX))) \mid
  \bTheta\right]$. We explicitly use
  the notation $V(\delta;\bTheta)$ to emphasize its dependence on the
  parameter $\bTheta$ as well as the policy $\delta$.
Specifically, $V(\delta;\bTheta)$ averages over both the marginal distribution of $\bX$, and the conditional distribution of $Y(\delta(\bX))$ given $\bX$ and $\bTheta$. 
The expectation in Equation~\eqref{eq:bayesoptimal} is taken over the
posterior distribution of $\bTheta$ given the observed data.

To solve the optimization problem in Equation~\eqref{eq:bayesoptimal},
we can rewrite the expectation over parameter $\bTheta$, covariates
$\bX$, and potential outcome $Y(\delta(\bX))$ as,
\begin{align}
  &\mathbb{E}\left[V(\delta;\bTheta)\mid \{ \bX_i,D_i,
  Y_i\}_{i=1}^n\right] \ =
  & \int_{\mathcal{X}}\sum_{d=0}^{K-1}\mathbb{E}\left[u(d,Y(d))\mid \bX=\bx,\{\bX_i,D_i,Y_i\}_{i=1}^n\right]\mathbb{I}(\delta(\bx)=d) dF(\bx),\label{eq:expec}
\end{align}
where the inside expectation is taken over the posterior distribution
of $\bTheta$ given the observed data, and the distribution of the
potential outcomes given $\bTheta$ and $\bX$. The outside expectation
is taken over the distribution of $\bX$, where $F$ is the CDF of
$\bX$. For simplicity, we will assume the distribution of $X$ in the
target population is known.

Thus, the original problem in Equation~\eqref{eq:bayesoptimal} can
consist of two steps: (i) estimate the posterior expected utility of
decision $d$ given a covariate profile $\bx$,
$\mathbb{E}\left[u(d,Y(d))\bigr|\bX=\bx,\{\bX_i,D_i,Y_i\}_{i=1}^n\right]$,
and (ii) find the optimal policy $\delta$ by solving the optimization
\eqref{eq:expec} with the estimated values.  This formulation
separates the problem of computing the posterior distribution from
that of finding an optimal policy.

\section{The Proposed Methodology}
\label{sec:framework}

We now describe our Bayesian safe policy learning framework.  We begin
by introducing a new risk metric, the Average Conditional Risk
(ACRisk), that represents the probability that a new policy yields a
worse expected utility conditional on covariates than the baseline
policy.  We then propose to maximize the posterior expected value of
the new policy while limiting the posterior expected ACRisk.  Our
methodology consists of two steps: first estimating the conditional
average treatment effect (CATE) using a flexible Bayesian model, and
then finding a policy within the policy class that maximizes the
posterior expected value while controlling the posterior expected
ACRisk.  Finally, we show that this chance-constrained optimization
problem can be written as a standard Bayesian policy learning problem
with linear constraints, avoiding additional computational complexity.

\subsection{Average Conditional Risk (ACRisk)}

In the existing literature, the risk of a policy is typically measured
through uncertainty about its value.  We then seek a policy that has a
high value with a small estimation uncertainty.  One limitation of
this common approach is its failure to account for potential
heterogeneity in risk across different groups of individuals.  A
policy that performs well on average may not benefit everyone.  In our
empirical application, a security assessment rule that is effective on
average may negatively impact some regions.  While the risk due to
estimation uncertainty will become small as the sample size increases,
this inherent risk due to heterogeneous treatment effects will remain,
even in large samples.

To address this, we introduce a new risk measure, the Average
Conditional Risk (ACRisk), which represents the probability that,
conditional on the covariates, a policy yields a worse expected
utility than the baseline policy.
\begin{definition}[Average Conditional Risk (ACRisk)] \spacingset{1}
  For a given policy $\delta$, the Average Conditional Risk with
  respect to a baseline policy $\tilde\delta$ is
  $$R(\delta, \tilde \delta;\btheta) := \mathbb{P}\left(
    \mathbb{E}\Big[u( \delta(\bX),Y( \delta(\bX))) \mid
    \bX,\bTheta=\btheta \Big]< \mathbb{E}\left[u(\tilde
      \delta(\bX),Y(\tilde \delta(\bX))) \mid \bX,
      \bTheta=\btheta\right] \right),$$ where $\bTheta$ represents the
  possibly infinite-dimensional parameter that governs the marginal
  distributions of potential outcomes.
\end{definition}
The inner expectation in the above definition is taken over the
conditional distribution of $Y(\delta(\bX))$ given $\bX$ and $\bTheta$
while the outer probability is defined with respect to the marginal
distribution of $\bX$. We use the above notation to make it explicit
that the ACRisk depends on the parameter $\bTheta$, which governs the
conditional distribution of potential outcomes given the covariates
$\bX$. Note that under this definition,
$R(\tilde \delta, \tilde\delta; \btheta) = 0$ for any value of
$\btheta$.

The ACRisk first evaluates whether a group of individuals with a
certain set of covariates has a lower expected utility under policy
$\delta$ than under the baseline policy $\tilde\delta$.  It then
computes the proportion of such at-risk groups in the population by
averaging over the distribution of covariates.  The ACRisk is
dependent on the covariates we choose, and using different set of
covariates can result in different at-risk groups, potentially leading
to different ACRisk values.

\begin{table}[t]
  \centering \spacingset{1}
  \caption{A numerical example that illustrates the Average    Conditional Risk (ACRisk).  In this example, both    Policies~I~and~II outperform the baseline policy on average, but the    improvement of Policy~I comes at the expense of the group B.  The ACRisk shows that 50\% of the population, which is the size of the group $B$ in this example, is at risk.}
  \label{tab:example}
  \begin{tabular}{ccccc}
    \hline
                  & \multicolumn{2}{c}{Conditional expected utility}                  & \multirow{2}{*}{Value} & \multirow{2}{*}{ACRisk} \\
                  & Group A (50\%)& Group B (50\%) &       &        \\
  \hline
  Baseline policy & 0                            & 0              & 0     & 0      \\
  Policy I        & 4                            & $-2$           & 1     & 0.5    \\
  Policy II       & 1                            & 1              & 1     & 0      \\
  \hline
  \end{tabular}
\end{table}

Table~\ref{tab:example} provides a numerical example where both
Policies~I~and~II outperform the baseline policy.  While the
improvement of Policy~I comes at the expense of group B, Policy~II
performs equally well for both groups.  As a result, the ACRisk of
Policy~I is higher with 50\% of the population at risk.

We note that the ACRisk is a risk measure based on groups (defined by
covariates $\bX$).  In particular, the ACRisk differs from the following
individual measure of risk.
\begin{definition}[Average Individual Risk (AIRisk)] \spacingset{1}
  For a given policy $\delta$, its average individual risk with
  respect to a baseline policy $\tilde\delta$ is defined as,
  $$R^\ast(\delta, \tilde \delta;\lambda) := \mathbb{P}\left(u(\delta(\bX), Y(\delta(\bX)) < u(\delta(\bX),
    Y(\tilde\delta(\bX))) \mid \Lambda = \lambda \right),$$
where $\Lambda$ represents a (possibly infinite dimensional) parameter
that governs the joint distribution of potential outcomes.
\end{definition}
Unlike the ACRisk, the AIRisk is not identifiable because it depends
on the joint distribution of potential outcomes.  One possible
strategy is to control an upper bound of the AIRisk \citep{kallus2022s,ben2022policy,li2022unit}.
We leave the development of safe policy learning in terms of the AIRisk
to future research and focus on the ACRisk for the remainder of this
paper.

To control the ACRisk, we must estimate the parameter $\bTheta$.
Under our Bayesian safe policy learning framework, we consider the
posterior average of the ACRisk, which we call the Posterior Average
Conditional Risk or PACRisk.  This measure allows us to directly
incorporate estimation uncertainty into our analysis.
\begin{definition}[Posterior Average Conditional Risk
  (PACRisk)] \label{def:PACRisk} \spacingset{1} For a given policy
  $\delta$, the Posterior Average Conditional Risk with respect to a
  baseline policy $\tilde\delta$ is
  $$R_p(\delta,\tilde\delta) :=
  \mathbb{E}\left[R(\delta,\tilde\delta;\bTheta)\mid \{\bX_i, D_i, Y_i\}_{i=1}^n\right],
  $$
  where the expectation is taken over the posterior distribution of
  $\bTheta$ given the observed data $\{\bX_i, D_i, Y_i\}_{i=1}^n$.
\end{definition}

\subsection{Bayesian Safe Policy Learning with Chance Constrained Optimization}
\label{subsec:bayes_safe}

We now propose a Bayesian safe policy learning procedure that limits the
PACRisk introduced above (Definition~\ref{def:PACRisk}).
Specifically, we find a policy within a policy class $\Delta$ that
maximizes the posterior average value while limiting the PACRisk below
a specified threshold $\epsilon \in [0, 1]$,
\begin{equation}
  \begin{aligned}
    \delta_{\text{safe}} \ = \ \argmax_{\delta \in \Delta}\ \
    &\mathbb{E}\left[V(\delta;\bTheta)\mid \{\bX_i, D_i, Y_i\}_{i=1}^n\right]  \\
     \text{subject to\ \ \ } &\mathbb{E}\left[R(\delta,\tilde\delta;\bTheta)\middle | \{\bX_i, D_i, Y_i\}_{i=1}^n\right]\le \epsilon,
  \end{aligned}  \label{eq:safe_bayes}
\end{equation}
The expectation in the optimization target is taken over the posterior
distribution of the parameter $\bTheta$. We typically choose a policy
class $\Delta$ that contains the baseline policy so that a solution to
Equation~\eqref{eq:safe_bayes} always exists.  The constraint in
Equation~\eqref{eq:safe_bayes} ensures that for a randomly selected
group of individuals, $\delta_{\text{safe}}$ has an at most $\epsilon$
posterior probability of yielding a worse expected value for the group
than the baseline policy $\tilde{\delta}$.  Thus, a smaller value of
$\epsilon$ yields a safer policy that is more conservative, limiting
the extent of changes to the baseline policy.

A standard Bayesian policy maximizes the posterior expected utility,
modifying the baseline policy for subgroups where the posterior
expected CATE is large. However, the additional PACRisk constraint in
Equation~\eqref{eq:expec} prioritizes changes to the baseline policy
for those subgroups whose posterior expected value of the CATE is
large and its posterior uncertainty is small.  Consider a toy example,
in which $\bX$ is a binary variable and
$\bTheta = (\theta_0,\theta_1)$ where $\theta_0$ and $\theta_1$ are
the CATEs for $\bX=0$ and $\bX=1$, respectively.  The baseline policy
is to treat nobody.  In addition, suppose that the posterior
distributions of $\theta_0$ and $\theta_1$ are $N(10,100)$ and
$N(1,0.1)$, respectively.  Under this setting, the Bayesian optimal
policy $\delta_{opt}$ will treat everyone even though the treatment
effect for the subgroup $\bX=0$ is highly uncertain. In contrast, the
PACRisk constraint in our methodology considers the posterior
probability of the group-level positive treatment effect, and will be
more conservative when changing the baseline policy may negatively
affect some subgroups due to a high degree of posterior uncertainty of
the CATE.

One potential drawback of chance constraints such as the one shown in
Equation~\eqref{eq:safe_bayes} is its computational difficulty due to
their complex dependency on both the parameter $\bTheta$ and the
policy $\delta$.  This makes it challenging to determine how the value
of the constraint changes as a function of the policy $\delta$.  For
this reason, researchers typically impose strong assumptions on the
marginal distribution of potential outcomes and the parameter
$\bTheta$ to obtain a closed-form solution instead
\citep[e.g.,][]{delage2010percentile,vitus2015stochastic,mowbray2022safe}.

We overcome this challenge by formulating
Equation~\eqref{eq:safe_bayes} as a deterministic optimization problem
with a linear constraint that separates the estimation of CATE from
policy optimization in the manner similar to Equation~\eqref{eq:expec}
under the standard Bayesian policy learning.  The following theorem
formally establishes the equivalence between the chance constraint
optimization in Equation~\eqref{eq:safe_bayes} and a constrained
linear programming problem.
\begin{theorem}[Control of the PACRisk as a Linear Constraint]
    \label{th1} \spacingset{1}
    Define the posterior conditional benefit and risk of decision $k$ relative to
    the existing policy $\tilde\delta$ as,
    \begin{equation*}
            b_k(\bx) := \mathbb{E}\left[\tau_k(\bx,\bTheta)\mid\{D_i, \bX_i,
              Y_i\}_{i=1}^n \right],\quad
            r_k(\bx) := \mathbb{P}\left(\tau_k(\bx,\bTheta)<0\mid \{D_i,
              \bX_i, Y_i\}_{i=1}^n \right),
      \end{equation*}
      where
      $\tau_k(\bx,\btheta):=\mathbb{E}\left[u(k,Y(k))-u(\tilde\delta(\bx),Y(\tilde\delta(\bx)))\mid
      \bX=\bx,  \bTheta=\btheta \right]$.  Then, the chance-constrained
      optimization defined in Equation~\eqref{eq:safe_bayes} is
      equivalent to the following deterministic optimization problem,
    \begin{equation}
      \begin{aligned}
            \delta_{\text{safe}} \ = \ & \argmax_{\delta \in \Delta}\ \
            \sum_{k=0}^{K-1}\int  I(\delta(\bX)=k)b_k(\bX)dF(\bX)\\
            & \text{subject to \ \ } \sum_{k=0}^{K-1} \int
            I(\delta(\bX)=k)r_k(\bx)dF(\bX) \le \epsilon
            \end{aligned}
        \label{eq:th1}
      \end{equation}
      where $F(\bx)$ is the cumulative distribution function of
      covariates $\bX$.
\end{theorem}
\noindent Proof is given in Appendix~\ref{pf1}. In the theorem,
$\tau_k(x,\btheta)$ is the improvement in the conditional expected
utility when changing the decision $\tilde\delta(x)$ to $k$ where
$\{\tau_k(x,\bTheta)\}_{k=0}^{K-1}$ is fully determined by $K-1$
pairwise CATEs between $K$ decisions.  In practice, we use the
empirical CDF for $F(\bX)$.

Theorem~\ref{th1} enables us to explicitly separate the posterior of
$\bTheta$ and the policy $\delta$ in the chance-constrained
optimization. We note that this formulation is possible because the
ACRisk of a policy $\delta$ is a weighted average of its conditional
risk given $\bX$, which can be represented by
$\{r_k(\bx)\}_{k=0}^{K-1}$. We can first calculate the conditional
improvement and risk, i.e., $b_k(\bx)$ and $r_k(\bx)$, using either
closed-form calculation or samples from posterior draws. Then, like
the standard Bayesian policy learning, we can solve the constrained
policy optimization problem based on $b_k(\bx)$ and $r_k(\bx)$.
Algorithm~\ref{alg:mcmc} summarizes the procedure of our proposed
Bayesian safe policy learning based on the posterior draws of
$\bTheta$.

\begin{algorithm}[t]
  \caption{Bayesian Safe Policy Learning based on Posterior Draws}
  \label{alg:mcmc}
  \KwIn{Data $\{\bX_i, D_i, Y_i\}_{i=1}^n$; A total of $M$ posterior
    draws, i.e., $\{\bTheta^{(m)}\}_{m=1}^{M}$; Covariate
    distribution function $F(x)$}
    
    \KwOut{Bayesian safe policy $\hat\delta_{\text{safe}}: \mathcal{X}
      \longrightarrow \mathcal{D}=\{0,1,\ldots,K-1\}$}
    
Approximate the posterior conditional benefit of decision $k$, $b_k(\bx) = \mathbb{E}\left[\tau_k(\bx,\bTheta)\mid
\{\bX_i, D_i, Y_i\}_{i=1}^n \right]$, by the sample average of
posterior draws for $\bx \in \mathcal{X}$:
  $$
      \hat b_k(\bx)= \frac{1}{M} \sum_{m=1}^{M} \tau_k(\bx,\bTheta^{(m)}) 
  $$

Approximate the posterior conditional risk of decision $k$. $r_k(\bx) = \mathbb{P}\left(\tau_k(\bx,\bTheta)<0\mid\{D_i,
\bX_i, Y_i\}_{i=1}^n \right)$, by the sample average of posterior
draws for $\bx \in \mathcal{X}$:
$$
\hat r_k(\bx)= \frac{1}{M} \sum_{m=1}^{M}I\left\{\tau_k(\bx,\bTheta^{(m)})<0\right\}
$$

Solve this approximated optimization problem:
  \begin{equation*}
      \begin{aligned}
          \hat{\delta}_{\text{safe}} = &\argmax_{\delta \in \Delta}\ \ \ \sum_{k=0}^{K-1}\int  I(\delta(\bX)=k)\hat b_k(\bX)dF(\bX)\\
  &\text{subject to \ \ } \sum_{k=0}^{K-1} \int  I(\delta(\bX)=k)\hat r_k(\bx)dF(\bX) \le \epsilon
      \end{aligned}
      \label{e10}
  \end{equation*}
\end{algorithm}

\subsection{Bayesian CATE estimation}
\label{subsec:implementation}

A primary advantage of the proposed Bayesian safe policy learning
framework introduced above is separating the tasks of estimating the
CATE and optimizing the policy.  This means that our framework can
readily accommodate the use of flexible Bayesian nonparametric models
for CATE estimation.  

As an example that mirrors our empirical application, suppose that we
have an ordered decision, $k \in \{0,1,\ldots,K-1\}$ and a continuous
outcome.  We begin by modeling the marginal distribution of the
baseline potential outcome $Y_i(0)$ and then the $K-1$ CATEs between
two adjacent treatment levels.  Specifically, we assume the following
model,
\begin{equation*}
    Y_i(k)  = \sum_{d=0}^k f_d(X_i) + \epsilon_i,  \quad \forall \ k \in\{0,1,..,K-1\}
\end{equation*}
where $\epsilon_i$ is an i.i.d Gaussian random variable with mean 0
and variance $\sigma^2$, $f_0(\bx)=\mathbb{E}[Y(0)\mid \bX=\bx]$, and
$f_k(\bx)=\mathbb{E}[Y(k)-Y(k-1)\mid \bX=\bx]$ for $1\le k\le K-1$.
Here, we have an infinite dimensional parameter
$\bTheta = \left(\sigma^2, \{f_k(\cdot)\}_{k=0}^{K-1}\right)$ that
determines the marginal distribution of each potential outcome $Y(k)$
given covariates $\bX$.  If we have binary or categorical outcomes, we
use a link function to transform the mean outcome to a continuous
scale, e.g.,
\begin{equation}
  Y_i(d)\mid \bX_i \sim \text{Bernoulli}(p_d(\bX_i)) \
  \text{where} \ p_d(\bX_i) = g^{-1}\left(\sum_{k=1}^d
    f_{k-1}(\bX_i)\right) \label{eq:link}
\end{equation}
where $g^{-1}(\cdot)$ is the inverse link function.

In principle, any Bayesian method can be applied for modeling
$\{f_k(\cdot)\}_{k=0}^{K-1}$.  In the Appendix~\ref{sec:bartgp}, we
discussed two popular methods, Bayesian Additive Regression Trees
(BART) and Gaussian Processes (GP) to demonstrate how such models can
be accommodated.

\section{A Simulation Study}
\label{sec:numerical}

In this section, we conduct a simulation study to examine the
empirical performance of the proposed Bayesian safe policy learning
methodology.  To highlight the key concepts, we consider a simple setup with a binary decision. 

\subsection{Setup}

We consider the following data generating processes with the number
of observations $n$ varying from 50 to 500, i.e.,
$n\in\{50, 100, 200, 500\}$.  There are two covariates
$\bX = (X_1, X_2)$ that are independently and identically distributed
according to a uniform distribution
$X_1, X_2 \stackrel{i.i.d}{\sim} \text{Uniform}(-1,1)$.  There are two
scenarios, one with covariate overlap and the other without it:
\begin{itemize}
\item {\bf Scenario~I (with covariate overlap):} The data are
  collected through a randomized experiment with
  $\mathbb{P}\left(D=1\mid \bX = \bx\right) = 0.5, \forall \bx\in
  \mathcal{X}$. Here, the baseline policy is to treat nobody, i.e.,
  $\tilde\delta(\bx) = 0, \forall \bx\in \mathcal{X}$.

\item {\bf Scenario~II (without covariate overlap):} The data are
  collected under a deterministic policy
  $\tilde\delta(x)= I(x_1>0.5)$, which serves as the baseline policy.
  Thus, the treatment assignment is also deterministic, i.e.,
  $\mathbb{P}\left(D=1\mid \bX = \bx\right) = \tilde\delta(\bx)$.
\end{itemize}

Within each simulation scenario, we specify the outcome model as,\footnote{We also consider a data generating process with binary outcomes in Appendix~\ref{sec:addsimu}. The results are broadly similar to the continuous case.}
\begin{equation*}
Y \mid \bX  \sim N\left(X_1 + X_2 + \{4I(X_1>0,X_2>0)-2\}D|X_1||X_2|,\sigma^2\right)
\end{equation*}
where we consider a high signal-to-noise ratio case ($\sigma = 1$), a medium signal-to-noise ratio case ($\sigma=2$), and a
low signal-to-noise ratio case ($\sigma=3$).  In this setup, treatment is effective for individuals with $X_1>0,X_2>0$.  Finally, we set the utility to be
equal to the outcome, i.e., $u(d,y)=y$. 

We consider policies based on a linear separation of the covariate
space $\mathcal{X}$:
\begin{equation}
    \Delta := \left\{\delta(\cdot,\cdot): \delta(x_1,x_2) = I(ax_1+bx_2+c>0); a,b,c\in\mathbb{R}\right\}
\end{equation}
Notice that this policy class does not contain the oracle treatment
rule --- giving the treatment to individuals with $X_1>0$ and
$X_2>0$. We apply the proposed Bayesian safe policy learning methods,
and use the empirical distribution of the covariates $\bX$ as an
approximate CDF of $\bX$.

For Scenario~I with covariate overlap, we use Bayesian Causal Forests
(BCF) to estimate the CATE and the expected outcome under the control
condition, i.e., $\E(Y(0))$, with the default hyperprior parameter
specification suggested by \cite{hahn2020bayesian}. For modeling
$\mathbb{E}[Y(0)|\bX]$, we use 200 trees and hyperparameters of
$\beta=2,\eta=0.95$.  For modeling the CATE, we impose a stronger
prior and use a BART model with 50 trees, and
$\beta=3,\eta=0.25$. Here, $\beta$ is the penalization hyperparameter
for the depth of trees (larger values lead to more shallow trees) and
$\eta$ is the splitting probability hyperparameter (larger values lead
to deeper trees).

For Scenario~II without covariate overlap, we use Gaussian Process
regression (GP) to model the expected outcome under the control
condition as well as the CATE. We set the kernel of the Gaussian
process as a Matern kernel with parameter $l=0.5, 1$ or $2$, $\nu=3/2$
(see Equation~\eqref{eq:matern}).  We use a non-informative prior
where the mean function of the GP, $m(x)$, is set to $0$. For other
hyperparameters in the kernel function, we set the length scale to
$\{0.5,1,2\}$ which correspond to a \textit{non-smooth},
\textit{moderately smooth}, and \textit{highly smooth} CATE. We set
$\sigma_0^2$ to $\{1,4,16\}$ which corresponds to a \textit{strong},
\textit{medium}, and \textit{weak} prior.

For both methods, we use \texttt{Stan} \citep{carpenter2017stan} to
compute 2,000 posterior samples from two chains with a burn-in period
of 500 samples.

\subsection{Findings}

Since we know the true data generating process, we can calculate the
ACRisk for the learned policy under different policy learning setups.
Since the proposed methodology controls the PACRisk, it is of interest
to investigate how the ACRisk of the learned policy as well as the
policy value change as functions of the safety constraint $\epsilon$
in Equation~\eqref{eq:safe_bayes}.  We also examine the influence of
the prior distribution on the performance of the proposed methodology
when there is no covariate overlap.

\begin{figure}[t!]
  \centering
  \begin{subfigure}[H]{0.495\textwidth}
    \includegraphics[width=\textwidth]{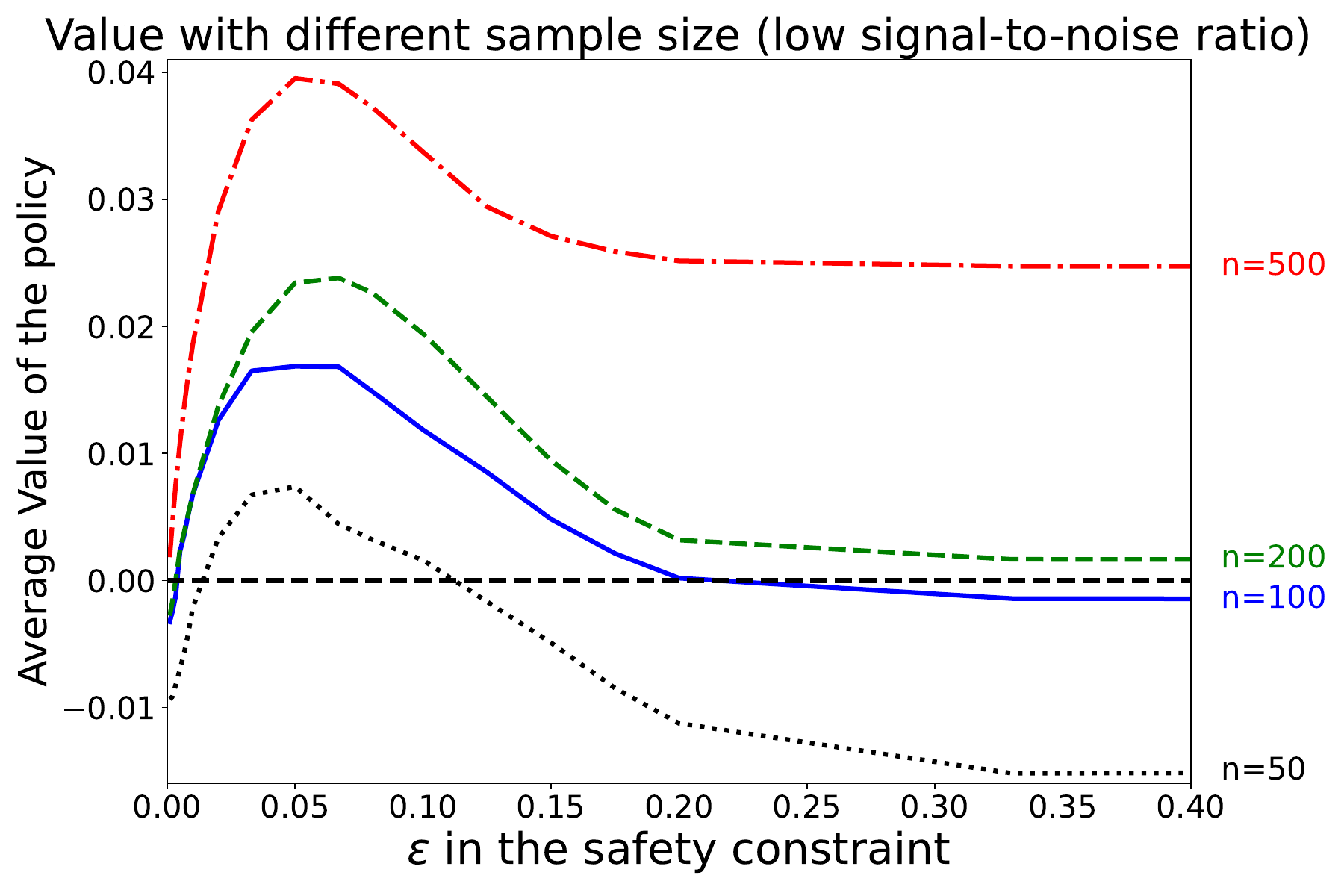}
    \caption{Average Value}
    \label{fig:simu11}
  \end{subfigure}
  \begin{subfigure}[H]{0.495\textwidth}
    \includegraphics[width=\textwidth]{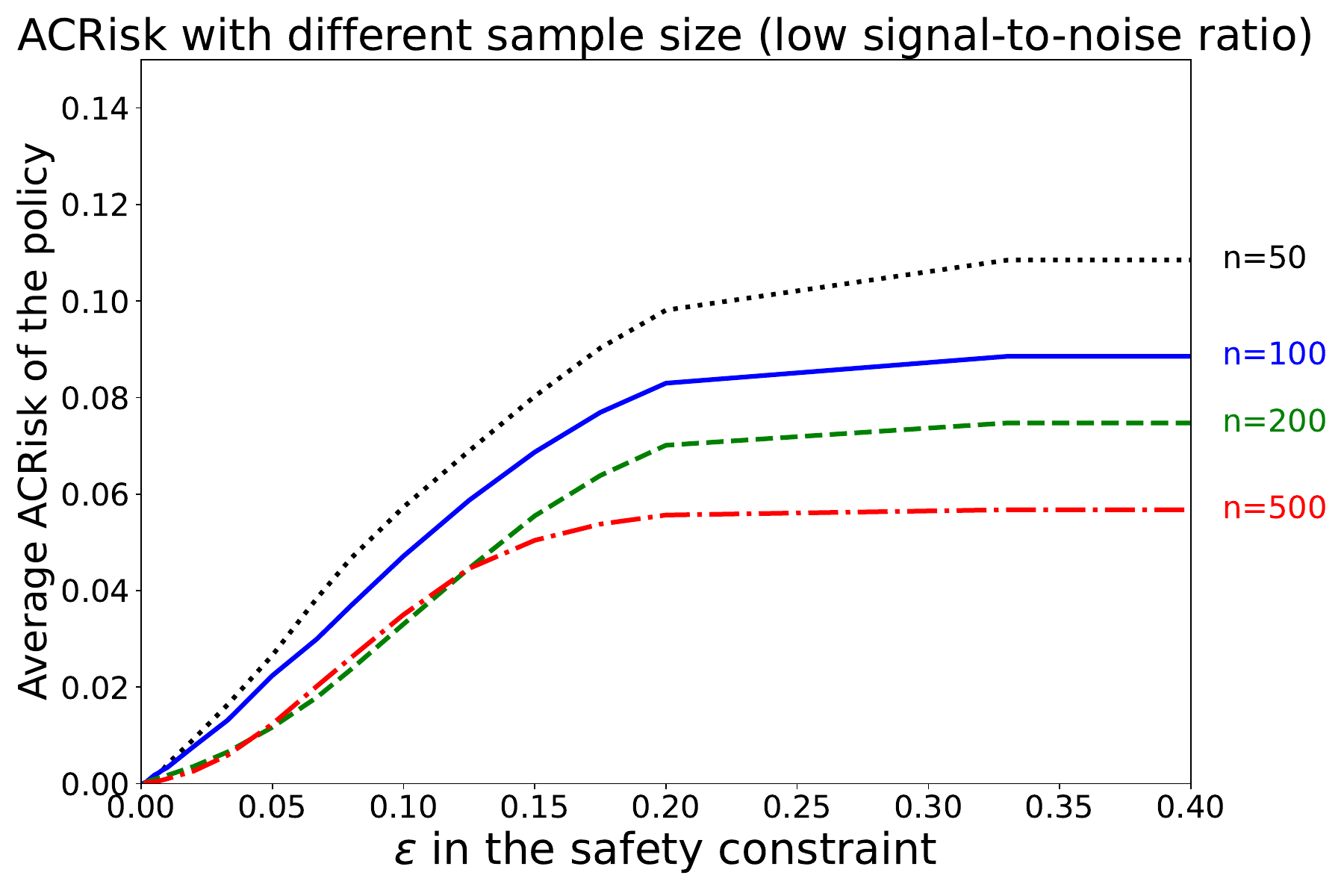}
    \caption{Average ACRisk}
    \label{fig:simu12}
  \end{subfigure}
  \caption{The average value (left panel) and ACRisk (right panel) of
    the learned policies using the data with covariate overlap,
    varying the safety constraint $\epsilon$ and sample size $n$.  } 
  \label{fig:simu1}
\end{figure}

\paragraph{Sample size and signal strength.} Figure~\ref{fig:simu1}
presents the average value (left panel) and average ACRisk (right
panel) of learned policies under different values of the safety
constraint $\epsilon$ ($x$-axis) with various sample sizes, using the
continuous outcome model.  Similarly, Figure~\ref*{fig:simu2} in the
appendix shows the results with different values of the safety
constraint $\epsilon$ and signal-to-noise ratio.  The data generating
process follows Scenario~I, and there is no extrapolation.  As
expected, we find that a weaker of safety constraint (i.e., a greater
value of $\epsilon$) leads to a greater ACRisk of the learned policy,
reaching a plateau that corresponds to the ACRisk obtained by
maximizing the posterior expected utility with no constraint.

On average, a smaller sample size and lower signal-to-noise ratio lead
to a greater ACRisk.  Interestingly, an appropriate level of the
safety constraint (around 0.05 in this simulation) achieves a greater
average value compared to maximizing the posterior expected value with
no constraint. This is especially true when the sample size is small
and/or the signal-to-noise ratio is low. In such settings, the safety
constraint regularizes the policy optimization, only modifying the
existing policy for those who are expected to benefit from a new
policy with a high degree of certainty.

\paragraph{Prior in extrapolation.} Next, we investigate the influence
of prior parameters in GPs on the performance of the proposed
methodology under Scenario~II where there is no covariate overlap.
Specifically, we consider the scale parameter $l$ and the variance
parameter $\sigma_0^2$, which determine the smoothness of CATE and
prior strength, respectively.  Recall that a greater value of $l$
implies a smoother function, while a smaller value of $\sigma_0^2$
corresponds to a stronger prior belief.  We fix the sample size to 200
and use a low signal-to-noise ratio when generating the data.

\begin{figure}[t!]
  \centering
  \begin{subfigure}[H]{0.495\textwidth}
    \includegraphics[width=\textwidth]{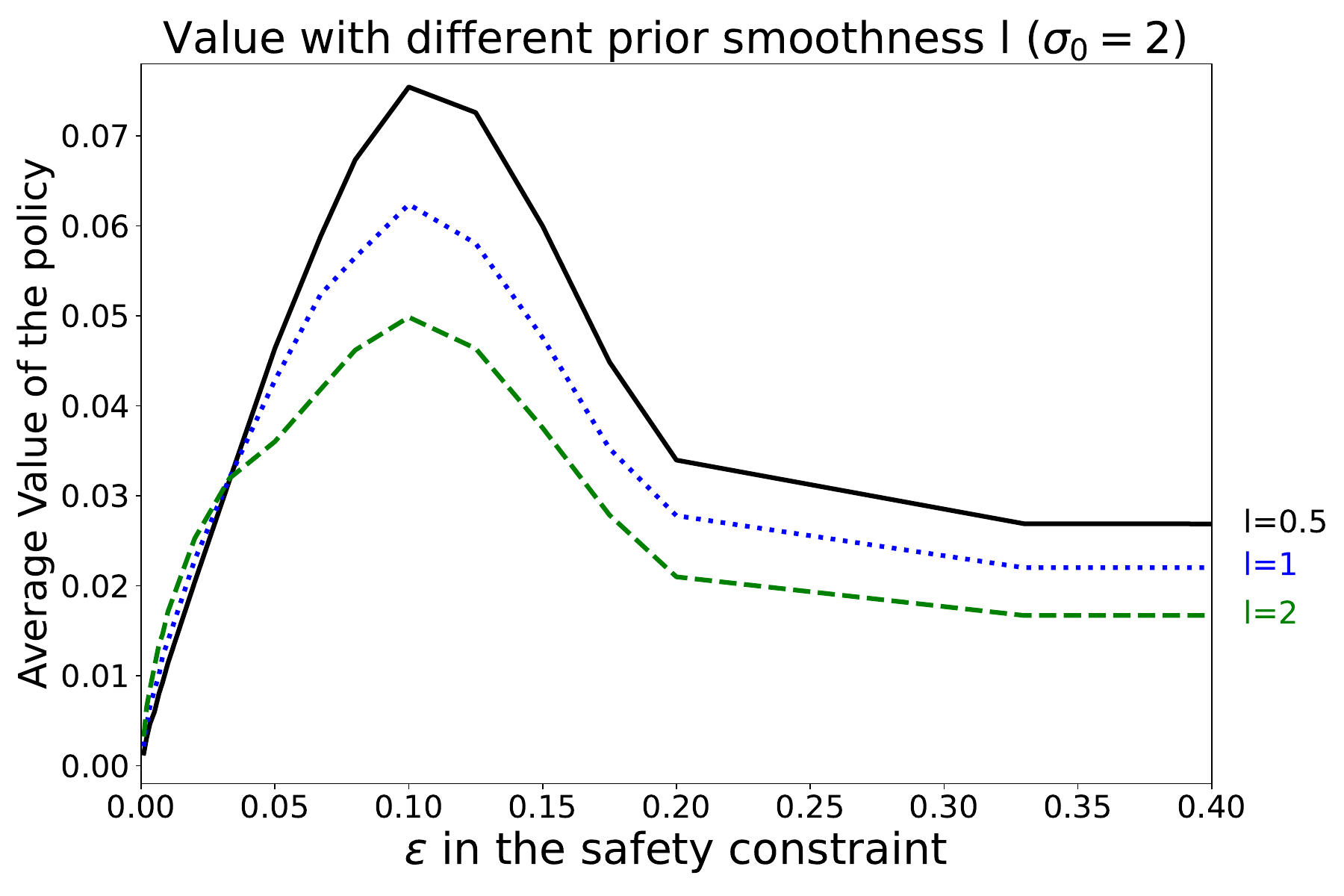}
    \caption{Average Value}
    \label{fig:simu31}
  \end{subfigure}
  \begin{subfigure}[H]{0.495\textwidth}
    \includegraphics[width=\textwidth]{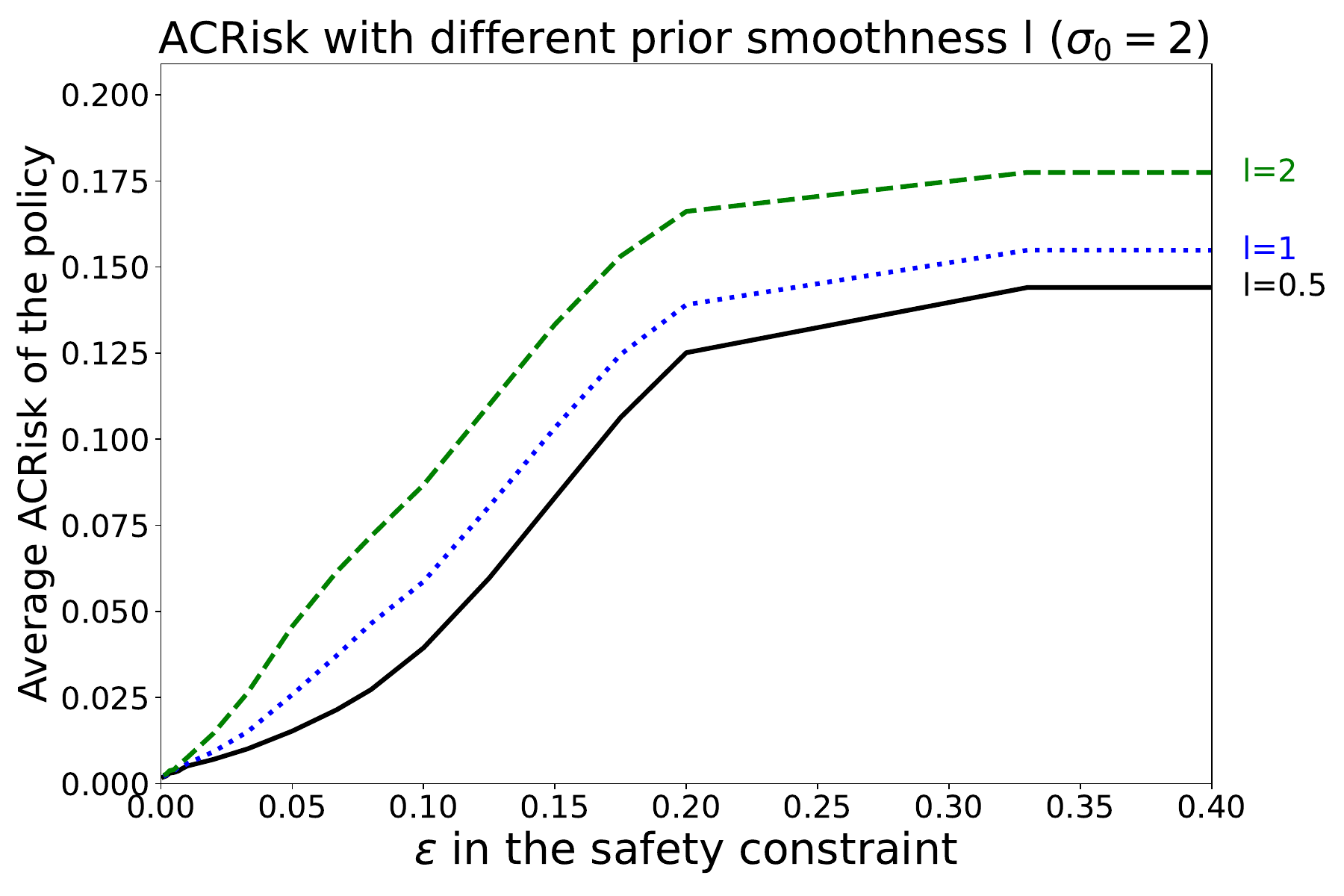}
    \caption{Average ACRisk}
    \label{fig:simu32}
  \end{subfigure}
  \caption{Average Value (left panel) and ACRisk (right panel) for
    learned policies using data without covariate overlap, varying the
    safety constraint $\epsilon$ and prior smoothness for the CATE $l$
    (a greater value corresponds to a greater degree of prior
    smoothness).}
  \label{fig:simu3}
\end{figure}

Figures~\ref{fig:simu31} shows the average value of the learned policy
under different prior smoothness.  In general, the value of learned
policy tends to decrease as the prior smoothness increases. However,
when the safety constraint is strong (i.e., $\epsilon$ is small), the
value of the learned policy with stronger prior smoothness is slightly
greater.  A small value of $\epsilon$ implies that we only wish to
change the baseline policy for a set of covariates that correspond to
large CATEs. A smoother prior leads to a more conservative estimate of
the CATE and can better identify such covariates.

Furthermore, Figure~\ref{fig:simu32} shows that as the prior for the
CATE becomes smoother, we extrapolate further, leading to an increase
in the average ACRisk. In Figure~\ref*{fig:simu4} of the appendix, we
observe a similar pattern when varying the prior strength for
estimating the CATE.

To summarize, our simulation study shows that the safety constraint on
the PACRisk can effectively control the true ACRisk in policy
learning, reducing the risk of harming specified subgroups. In
addition, a moderately binding safety constraint can induce beneficial
regularization and improve the average value of the learned policy,
especially when the sample size is small and the signal-to-noise ratio
is low.

\section{Empirical Analysis of the HES Security Assessment}
\label{sec:application}

With our methodological development in hand, we now return to the
analysis of the military security assessment described in
Section~\ref{sec:hes}.  We will focus on learning a new rule to
aggregate the 20 sub-model scores into one overall security score,
while keeping the original structure of the HES and only modifying the
two-way and three-way tables used for the aggregation.

\subsection{Setup}

To formalize our problem, let $X^{(k)}\in [1,5]$ denote the $k$-th
sub-model scores with $k \in \{1,2,\ldots,20\}$, where
$\bX \in \mathcal{X}$ denotes the whole 20-dimensional input vector.
We use $S^{(i,j)}\in\{1,2,3,4,5\}$ to represent the $i$th score at the
$j$th level where $i \in \{1,2,\ldots,n_j\}$ and $j\in \{1,2,3\}$ with
$n_j$ representing the total number of scores aggregated at the $j$-th
level.  We note that $X^{(k)}\not=S^{(1,k)}$ because the raw input
score $X^{(k)}$ is rounded to obtain the first level score
$S^{(1,k)}$. To apply the Bayesian safe policy learning method, we use
the empirical CDF of $\mathbf{X}$ as an approximate of its population
CDF.

The output security score is denoted by
$D\in \mathcal{D}=\{1,2,3,4,5\}$.  We use $Y \in \{0,1\}$ to represent
one of three binary regional development outcomes measured in January,
1970: \textit{regional safety}, \textit{regional economy}, and
\textit{regional civic society}. These variables are calculated in
\cite{dell2018nation} with a latent class analysis on data from the
HES, armed forces administrative records, and public opinion surveys.
We analyze each outcome separately.

We consider a policy $\delta$, which is a deterministic function that
maps the inputs $\bx \in \mathcal{X}$ (the 20 sub-model scores) to an
output score $d \in \mathcal{D}$.  We define two-way and three-way
decision tables as the following functions that map integer-valued
input scores to an integer-valued output score.  Recall that each
input/output score takes an integer value, ranging from 1 to 5.
\begin{equation}
    \begin{aligned}
        T_2(\cdot, \cdot)&: \{1,2,3,4,5\}^2\rightarrow\{1,2,3,4,5\},\\
        T_3(\cdot, \cdot, \cdot)&: \{1,2,3,4,5\}^3\rightarrow\{1,2,3,4,5\}.
\end{aligned}
\end{equation}

Let $\widetilde{T}_{2}$ and $\widetilde{T}_{3}$ represent the baseline
decision tables used in the existing HES.  These existing rules are
\emph{monotonic}, meaning that a decision table with a greater input
score never yields a smaller output score, holding the other input
scores constant.  Formally, a two-way decision table $T_2$ is
monotonic if and only if we have $\ T_2(i_1,j_1) \le T_2(i_2,j_2)$ for
all $i_1,j_1,i_2,j_2\in \{1,2,3,4,5\}$ and $i_1\le i_2, j_1\le j_2$.
Similarly, a three-way decision table $T_3$ is monotonic if and only
if we have $T_3(i_1,j_1,k_1) \le T_3(i_2,j_2,k_2)$ for all
$i_1,j_1,k_1,i_2,j_2,k_2\in \{1,2,3,4,5\}$ and
$i_1\le i_2, j_1\le j_2,k_1\le k_2$.  Recall that the sub-model scores
are ordered so that lower values are ``worse.''  Thus, the
monotonicity condition is reasonable because higher sub-model scores
should not make a region less secure.  We will require that our
learned decision tables also satisfy this monotonicity condition.

\subsection{Learning to aggregate military, economic, and political sub-models}
\label{sec:one_layer}

We begin by learning a single decision table while keeping the other
tables of the HES.  In particular, we consider learning a new
third-level three-way decision table $T_3$ that aggregates three
scores --- the Military, Political, and Social Economic models --- and
produces the final output security score $D$ (see
Figure~\ref{fig:hes_agg}).

We use separate GPs with a logit link function $g$ to model the
expected outcome under different decisions (see
Section~\ref{subsec:implementation}).  Specifically, we assume the
model given in Equation~\eqref{eq:link} where $d=1,2,3,4,5$.  We use
$f_0$ to model the expected outcome under the decision $D=1$, whereas
$f_k$ with $k=1,2,3,4$ is used to model the effect between two
adjacent decisions.

We set the prior mean function for all the GPs to zero, i.e.,
$m(x)=0$, implying that all potential outcomes are distributed as
Bernoulli with probability $1/2$.  For the kernel function of the GP,
we use a Matern kernel with $\nu=3/2, l=1,\sigma_0^2 = 4$ (see
Equation~\eqref{eq:matern}). This corresponds to a weak prior,
implying that with more than $90\%$ probability, $f_k(\cdot)$ is no
less smooth than a Lipschitz function with the Lipschitz constant of
$10.95$.  Appendix~\ref{sec:smoothness} presents a sensitivity
analysis where we use larger/smaller values of $l$ to induce more/less
extrapolation. Overall, these results are consistent with our main
findings that are presented below.

Optimizing a single decision table in our application is not
computationally demanding.  Therefore, we use the Gurobi solver to
optimize Equation~\eqref{eq:safe_bayes} over three-way monotonic
decision tables.  We use the military impacts on regional
\textit{safety}, \textit{economy}, and \textit{civic society} as
separate binary outcomes and set the utility function to $u(D,Y) =
Y$. Our utility only considers the outcome and does not penalize the
use of a lower or higher security score.

\begin{figure}[t!]
  \centering
  \begin{subfigure}[t]{0.48\textwidth}
    \includegraphics[width = \textwidth]{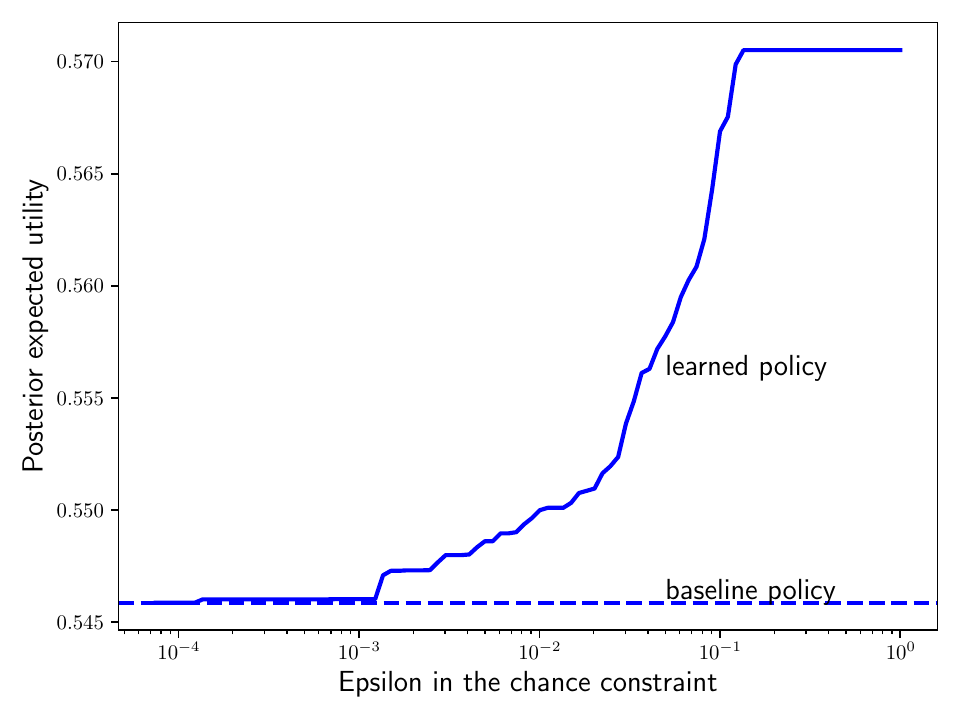}
    \caption{Posterior expected utility}
    \label{f811}
  \end{subfigure}
  \begin{subfigure}[t]{0.48\textwidth}
    \includegraphics[width = \textwidth]{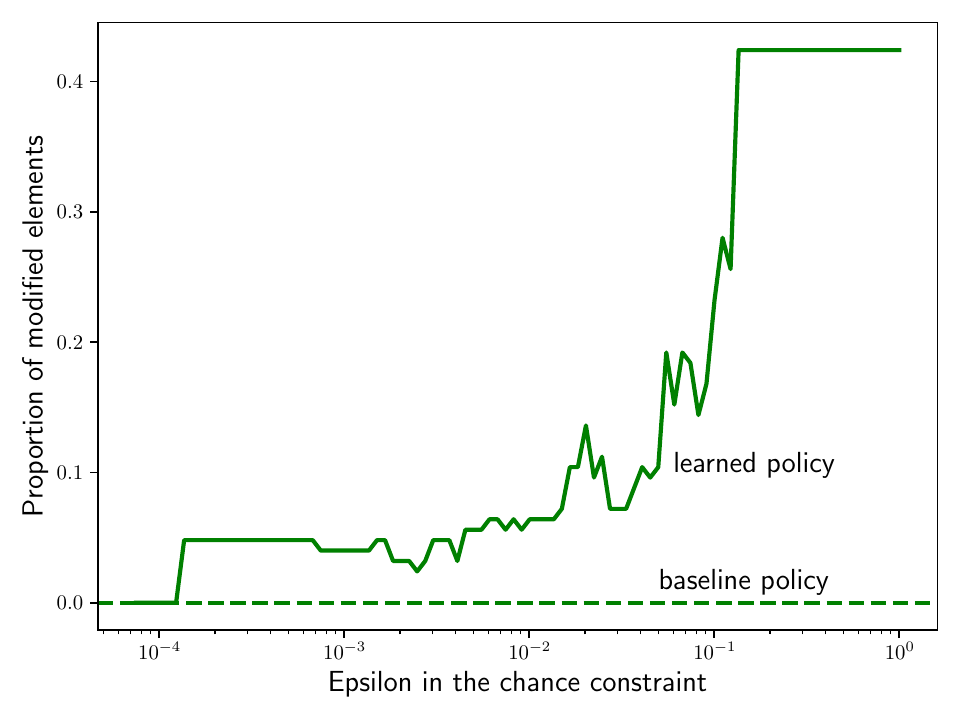}
    \caption{Proportion of changed elements}
    \label{f812}
  \end{subfigure}
  \caption{The posterior expected utility of the learned policy (left
    panel) and the proportion of elements in the three-way table
    changed by the learned policy (right panel) under different values
    of $\epsilon$, when regional \textit{safety} development is the
    outcome. A weaker safety constraint (i.e., a greater value of
    $\epsilon$) leads to a greater difference between the baseline and
    learned policies.  The posterior expected utility also becomes
    greater. } \label{fig:regional_security}
\end{figure}

Figure~\ref{fig:regional_security} shows how the results of our
Bayesian safe policy learning procedure change with the safety
constraint $\epsilon$ when the outcome is regional
\textit{safety}. Figure~\ref{f811} shows the posterior expected
utility of the learned policy, while Figure~\ref{f812} presents the
proportion of changed elements in the three-way table.  We find that a
weaker safety constraint (i.e., a greater value of $\epsilon$) leads
to a greater difference between the learned and baseline policies in
terms of the changed elements of the three-way table.  The learned
policy also has a higher posterior expected utility when the safety
constraint is weaker.

\begin{figure}[t!]
  \centering
  \includegraphics[width = 0.55\textwidth]{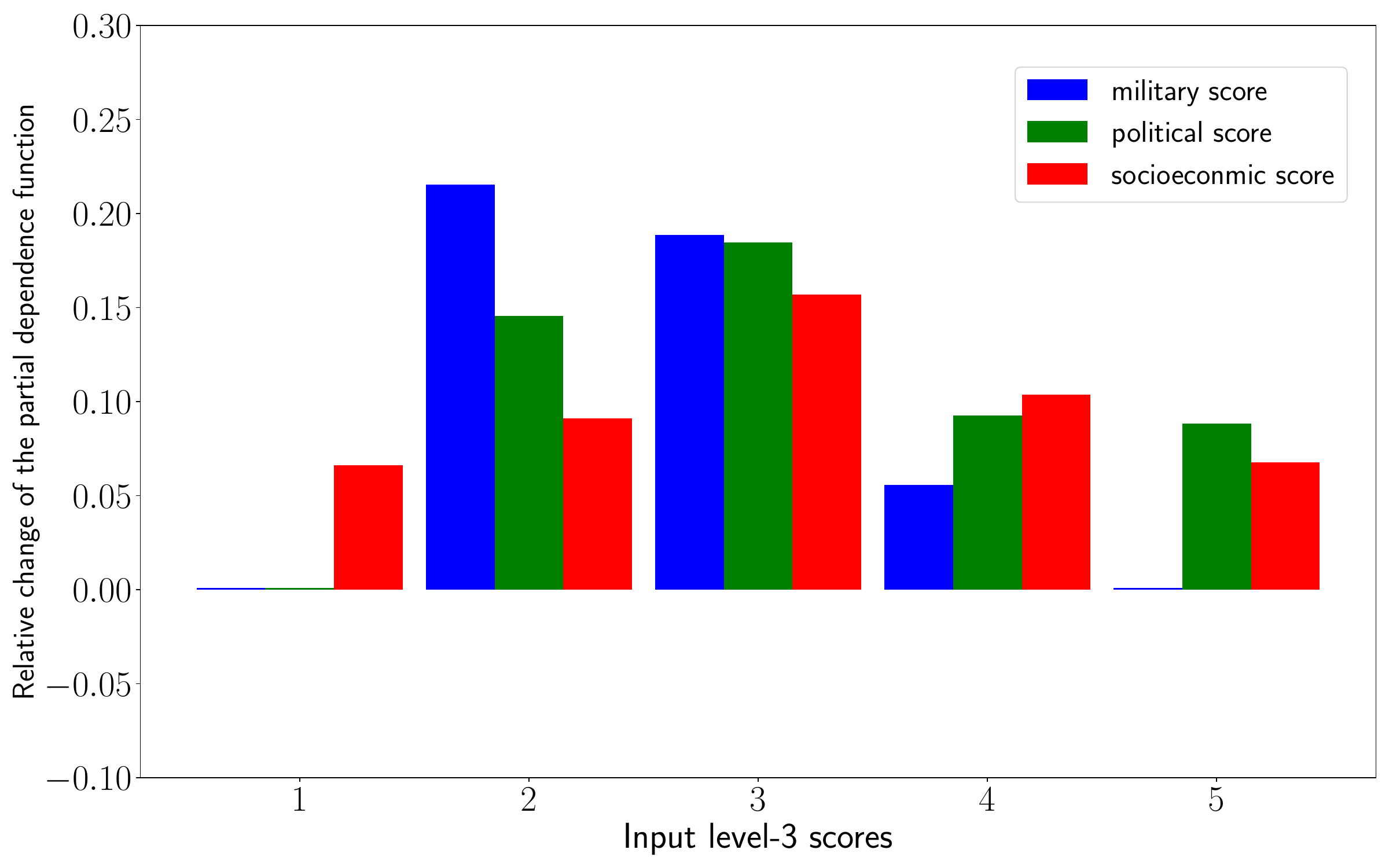}
  \caption{The relative change of the PD function from the baseline
    policy to the learned policy for $\epsilon = 0.1$. Each block
    corresponds to a different input of the PD function, and different
    colors corresponds to different level-3 scores. For example, the
    first blue bar in the first block corresponds to
    $\{I_{military}(1;T_3)-I_{military}(1;\tilde
    T_3)\}/I_{military}(1;\tilde T_3)$, where $T_3$ and $\tilde T_3$
    are the learned and baseline policies,
    respectively. } \label{fig:partial_dependence}
\end{figure}

Figure~\ref{fig:partial_dependence} presents the relative difference
in the partial dependence (PD) of the baseline and learned policies
with respect to the three level-3 scores.  The PD function measures
the dependence of the output security score on the level-3 scores,
i.e., $S^{(3,1)},S^{(3,2)},S^{(3,3)}$, by computing the marginal
expected output of the policy given certain input scores
\citep{friedman2001greedy,greenwell2018simple}.  For example, the PD
function for policy $\delta(\cdot;T_3)$ with respect to level-3
military score $S^{(3,1)}$ is computed as,
\begin{equation*}
        I_{military}(x; T_3)
        \ = \ \frac{1}{n}\sum_{i=1}^n T_3(x,S_i^{(3,2)},S_i^{(3,3)}).
\end{equation*}

We find the percent change in the PD function between the learned and
baseline policies are mostly positive.  That is, the expected output
security score under the learned policy is generally higher than the
baseline policy for all three level-3 input scores.  Recall that a
higher security score signifies that a region should not be targeted,
so increasing the scores would be likely to decrease the frequency of
airstrikes, and potentially improve the regional outcome.  This
finding is consistent with that of \citet{dell2018nation}: airstrikes
increased the military and political activities of the communist
insurgency, decreasing regional safety and the regional economy.

We further compute the PD importance \citep{greenwell2018simple} of
the military, political, and socioeconomic level-3 scores under the
learned policy as a function of the safety constraint $\epsilon$,
using the three different outcomes.  The PD importance measures the
degree to which a PD function is sensitive to the input.  A small PD
importance means that the output score does not strongly depend on the
input.  For ease of interpretation, we scale the PD importance of
level-3 scores for each policy so that they sum to one, allowing us to
compare the PD importance of level-3 scores across different policies.

\begin{figure}[t!]
    \centering
    \begin{subfigure}[b]{0.495\textwidth}
        \includegraphics[width=\textwidth]{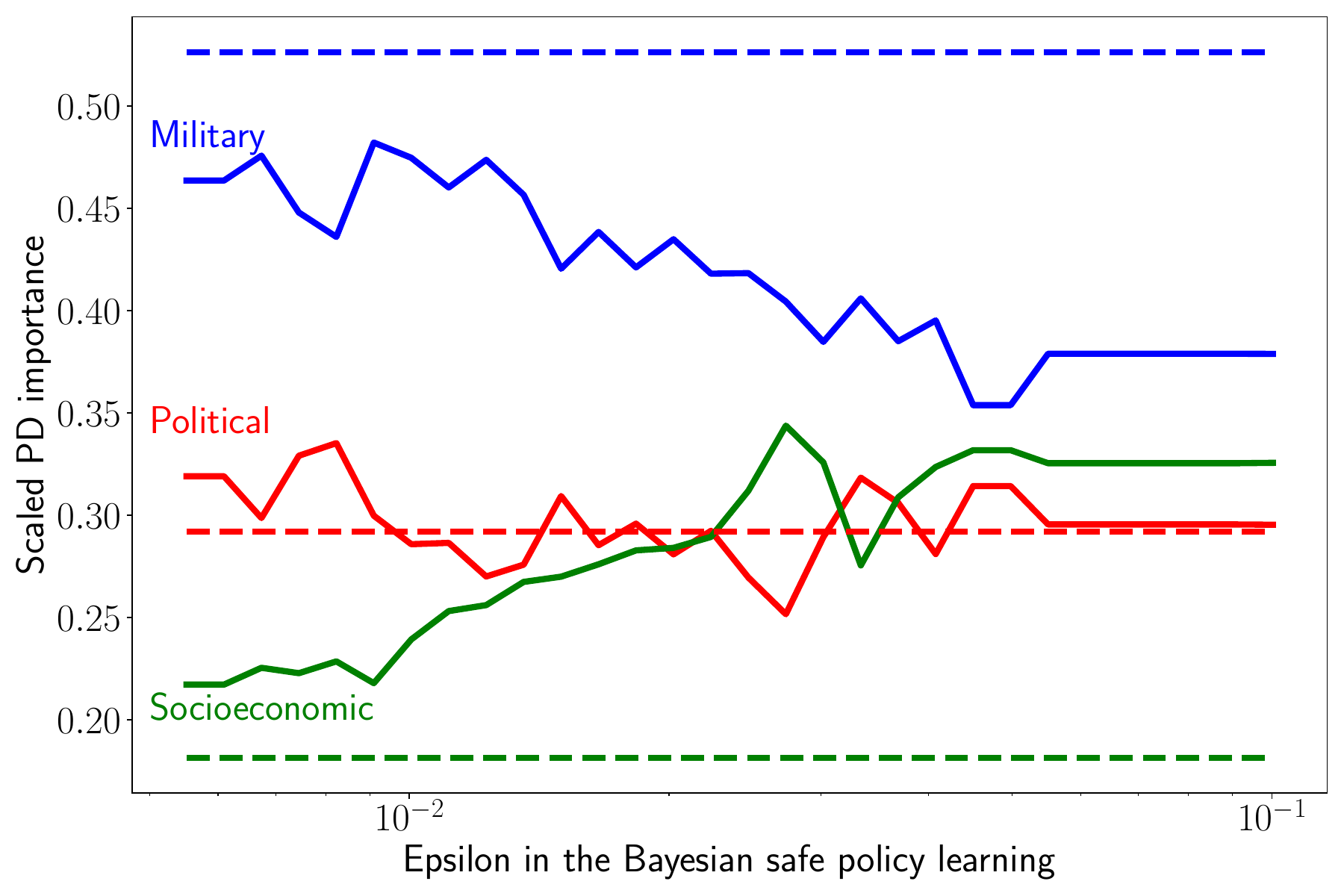}
        \caption{Regional economy as outcome}
        \label{f816}
    \end{subfigure}     
   \begin{subfigure}[b]{0.495\textwidth}
        \includegraphics[width=\textwidth]{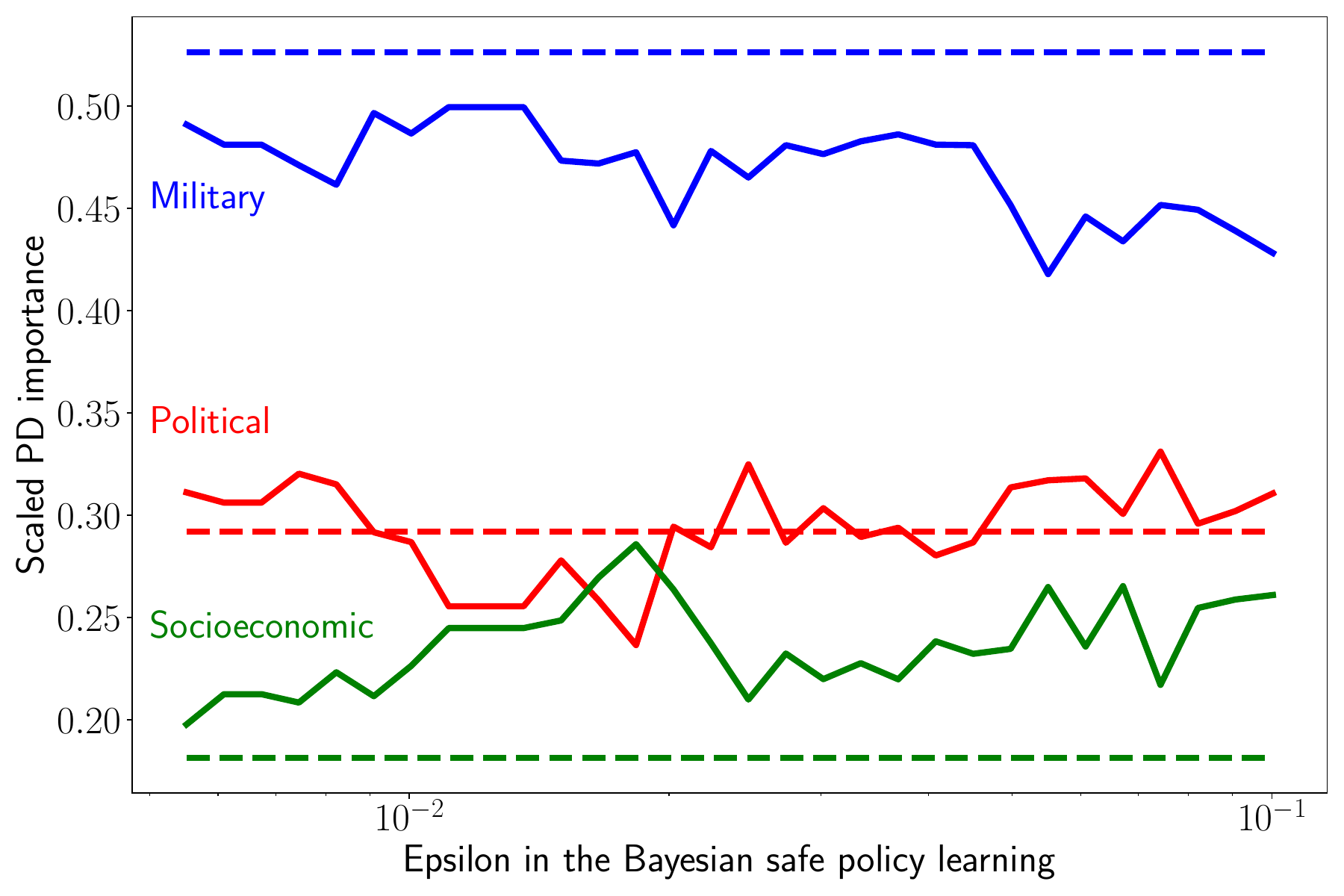}
        \caption{Regional safety as outcome}
        \label{f815}
    \end{subfigure}
    \caption{The scaled Partial Dependence (PD) importance of level-3
      scores of the learned policy, as a function of the $\epsilon$. The
      solid line corresponds to the learned policy, and the dashed
      line indicates the baseline policy. Lines with different colors
      shows the PD importance of different level-3
      scores. }
    \label{f8156}
\end{figure}

Figure~\ref{f8156} compares the scaled PD importance of each factor
(denoted by different colors) between the baseline (dotted lines) and
learned (solid lines) policies as a function of the strength of the
safety constraint $\epsilon$.  The figure shows a consistent pattern,
in which the new learned policies upweight the socioeconomic model and
downweight the military model, when both the regional economy and
regional safety are used as outcomes. Figure~\ref*{f817} in the
appendix further investigates the PD importance of different factors
in learned policy with civic society outcomes.  Overall, our analysis
suggests that the HES over-emphasized the importance of the military
model for all three types of outcomes.

\subsection{Learning the entire scoring system}
\label{sec:three_layer}

Next, we consider learning both the two-way and three-way tables
through all three levels of aggregation.  In other words, we consider
a policy class with the same aggregation structure as in the original
HES, but allows the two-way and three-way tables to be modified.
Formally, we consider the following policy class,
$\Delta_3 = \{\delta(\cdot;T_2,T_3): T_2,T_3 \ \text{is monotonic}\}.$
We use the same posterior CATE draws obtained in the previous
analysis.

Unlike Section~\ref{sec:one_layer}, this complex policy class makes it
difficult to directly optimize the objective. In
Appendix~\ref{sec:opt}, we develop an optimization method based on
directed acyclic graph (DAG) partitioning to solve this specific
problem. As in the previous analysis, we compute the PD importance
measure for each sub-model score.

\begin{figure}[p]
    \begin{subfigure}[H]{0.8\textwidth}
        \includegraphics[width=\textwidth]{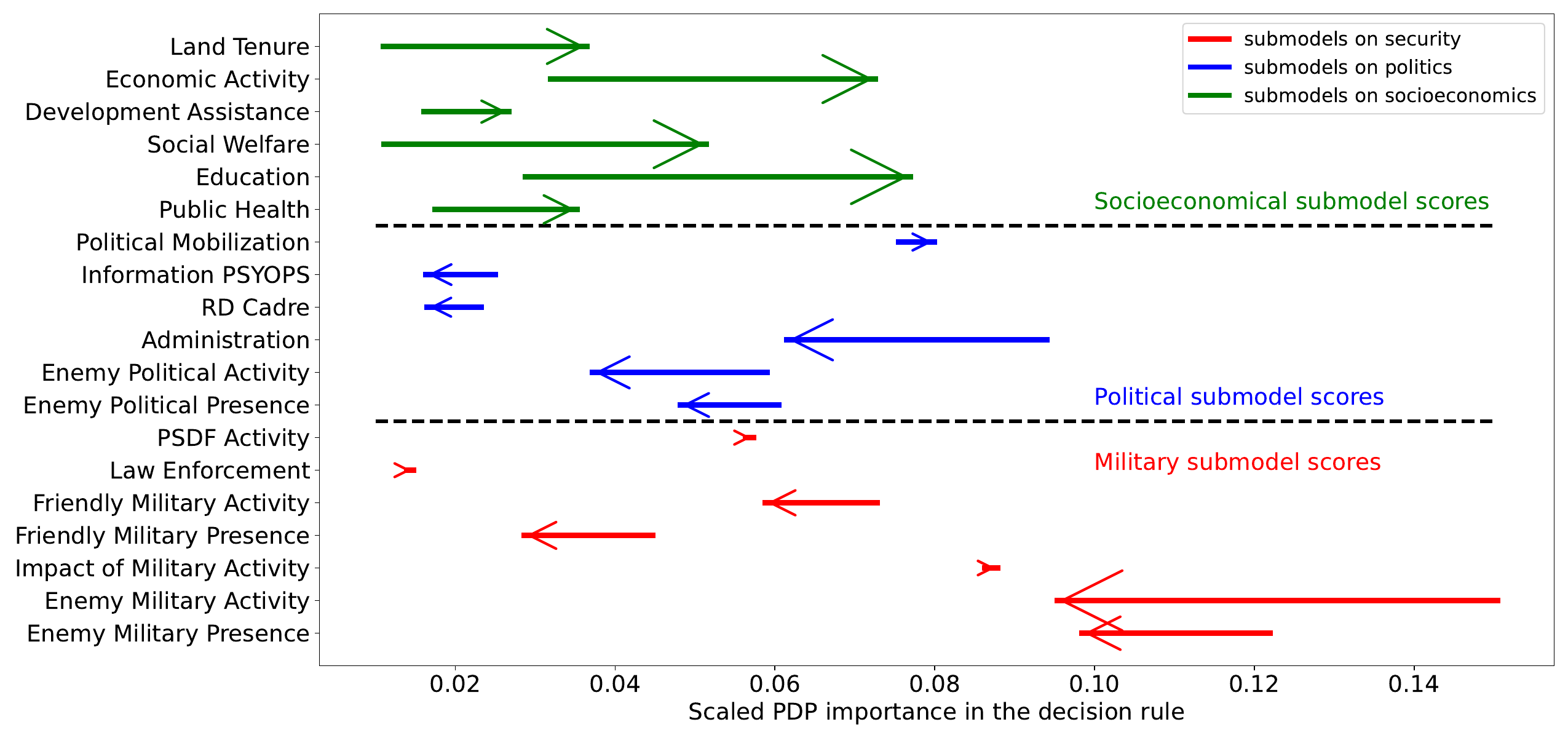}
        \caption{Regional economy as outcome}
        \label{f821-2}
        \end{subfigure}
        \begin{subfigure}[H]{0.8\textwidth}
            \includegraphics[width=\textwidth]{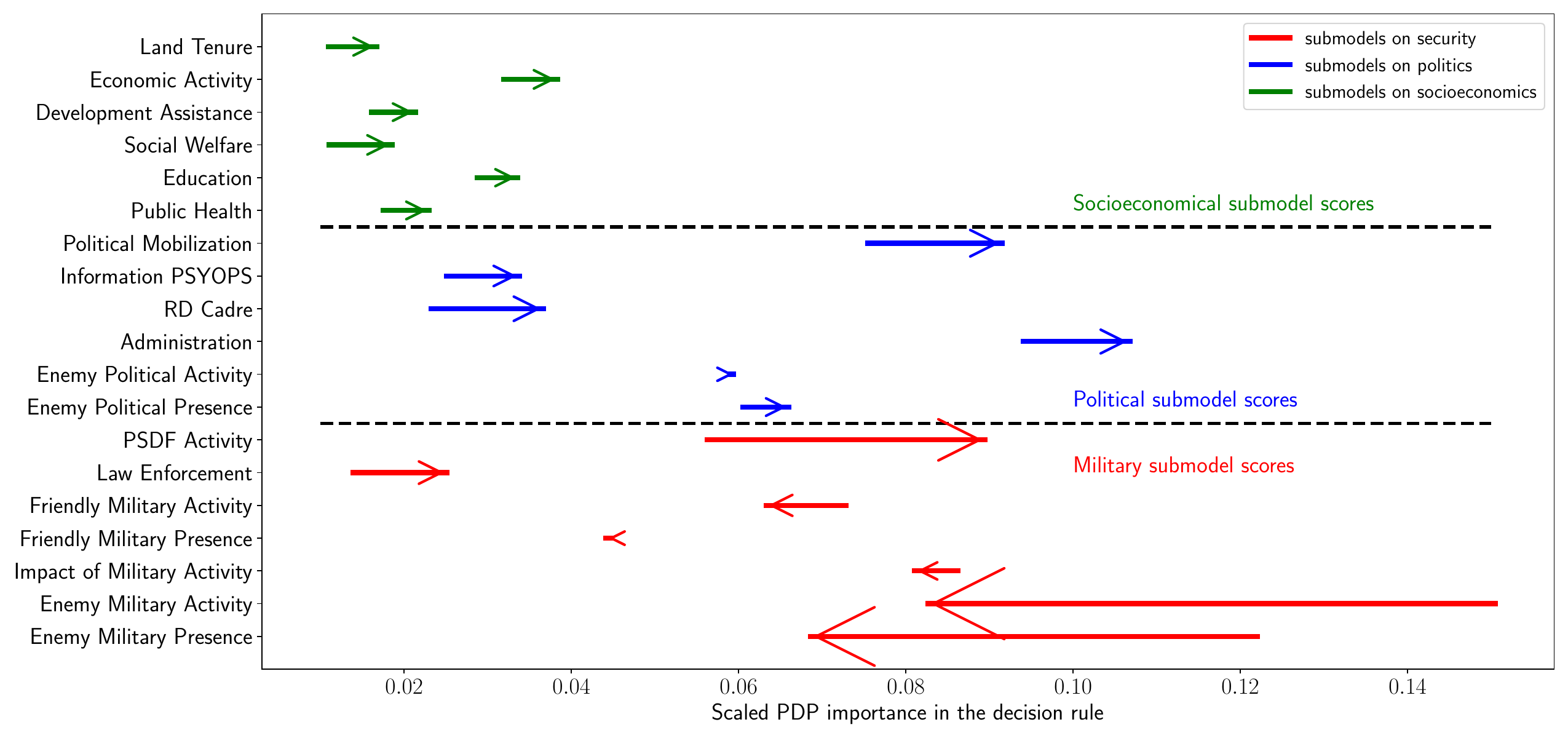}
            \caption{Regional civic society as outcome}
            \label{f821-3}
        \end{subfigure}
    \begin{subfigure}[H]{0.8\textwidth}
    \includegraphics[width=\textwidth]{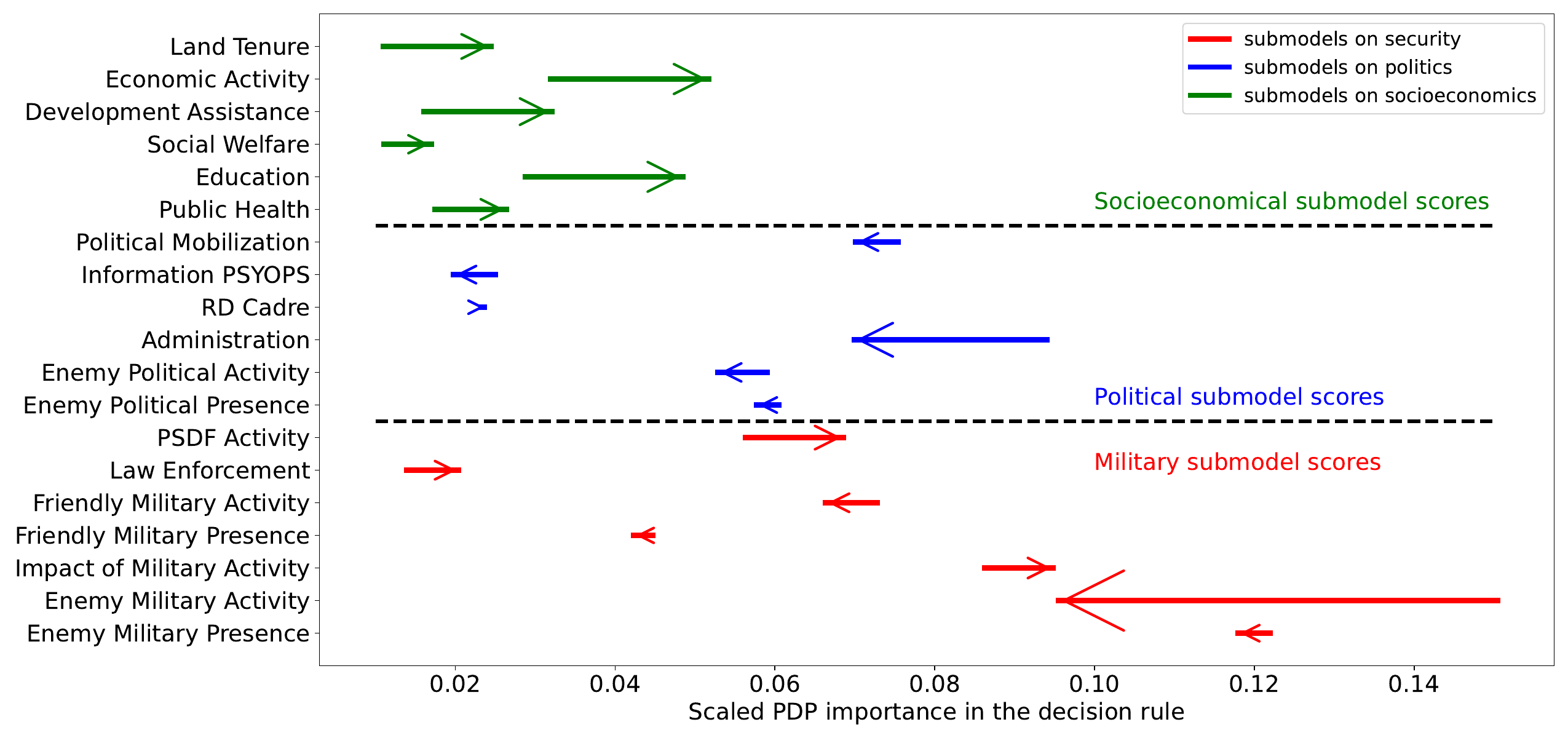}
    \caption{Regional safety as outcome}
    \label{f821-1}
    \end{subfigure}
    \caption{The scaled PD importance of 19 different sub-model scores in the learned policy ($\epsilon=0.1$) and baseline policy with different outcomes. Each arrow indicates the change of the scaled PDP importance from the baseline policy to the learned policy. }
    \label{f821}
\end{figure}

As the policy class $\Delta_3$ is large and contains many different
policies, it is difficult to directly interpret the obtained policy
through the learned decision tables. Therefore, we use the PD
importance measure to understand how the data-derived policy differs
from the HES.  Figure~\ref{f821} presents the relative importance of
the sub-model scores under the baseline and optimal policies with a
safety constraint $\epsilon=0.1$.  Each plot shows the scaled PD
importance of the 19 different sub-model scores under the baseline and
learned policies for the three different outcomes: regional safety,
economy, and civic society.  Each arrow shows how the scaled PD
importance changes from the baseline policy to the learned policy.

We find that when the outcome is either regional economy or civic
society (Figures~\ref{f821-2}~and~\ref{f821-3}), the learned policy
puts less weight on the security sub-model scores than the baseline
policy, with particularly large declines in the relative importance of
enemy military activity and presence.  Note, however, that this is
partially offset by an increase in the relative importance of PSDF
(People's Self-Defense Force) activity and law enforcement.

In contrast, the socioeconomic sub-model scores are more important
when targeting all three outcomes, and particularly so when targeting
the economic outcome.  Finally, the change in the relative importance
of the political sub-model scores is more mixed; their importance only
changes slightly when targeting regional safety (Figure~\ref{f821-1}).
It generally declines when targeting the regional economy, and
increases when targeting regional civic society.

\section{Concluding Remarks}
\label{sec:discussion}

In this paper, we propose a new notion of the average conditional risk
(ACRisk), which represents the population proportion of groups that
would be worse off under a new policy than under the baseline policy.
We then develop a Bayesian safe policy learning methodology that
limits this risk.  We separate the estimation of heterogeneous
treatment effects from the optimization of policies, enabling the use
of flexible Bayesian nonparametric models while considering a complex
policy class.

We apply the proposed methodology to the HES, a military security
assessment used during the Vietnam War to guide airstrike decisions
for USAF commanders.  Substantively, our analysis shows that the
HES---which saw active use during the war---attached too much
importance to the military security scores.  In contrast, the Bayesian
safe policy assigns greater weights to socioeconomic factors.  We also
develop a stochastic optimization algorithm that is generally
applicable to monotonic decision tables.  Given the popularity of such
decision rules in public policy and other settings, we believe that
our optimization algorithm is of independent interest.

There are several directions for further research. First, while we
take a Bayesian approach, one may consider a frequentist approach to
controlling the ACRisk in policy learning.  Second, our simulation
study suggests that the posterior ACRisk constraint plays a role of
regularization in policy optimization and an appropriate level of risk
constraint can improve the average value of the learned policy.
Better understanding how regularization improves policy learning is an
important next step.  Third, it is of interest to extend the proposed
Bayesian policy learning framework to other types of constraints such
as fairness and budget constraints.

\bibliography{BayesSafe}
\end{document}


\spacingset{1.375}

\clearpage
\appendix
\setcounter{equation}{0}
\setcounter{figure}{0}
\setcounter{table}{0}
\setcounter{theorem}{0}
\setcounter{lemma}{0}
\setcounter{section}{0}
\setcounter{definition}{0}
\renewcommand {\theequation} {S\arabic{equation}}
\renewcommand {\thefigure} {S\arabic{figure}}
\renewcommand {\thetable} {S\arabic{table}}
\renewcommand {\thetheorem} {S\arabic{theorem}}
\renewcommand {\theproposition} {S\arabic{proposition}}
\renewcommand {\thelemma} {S\arabic{lemma}}
\renewcommand {\thesection} {S\arabic{section}}
\renewcommand {\thedefinition} {S\arabic{definition}}

\begin{center}
  \spacingset{1}
  \LARGE {\bf Supplementary Appendix for\\ ``Bayesian Safe Policy
    Learning with Chance Constrained Optimization: Application to
    Military Security Assessment during the Vietnam War''}
\end{center}

\section{Two-way and three-way decision tables used in the HES}

\begin{figure}[H]
    \centering
    \includegraphics[scale=0.4]{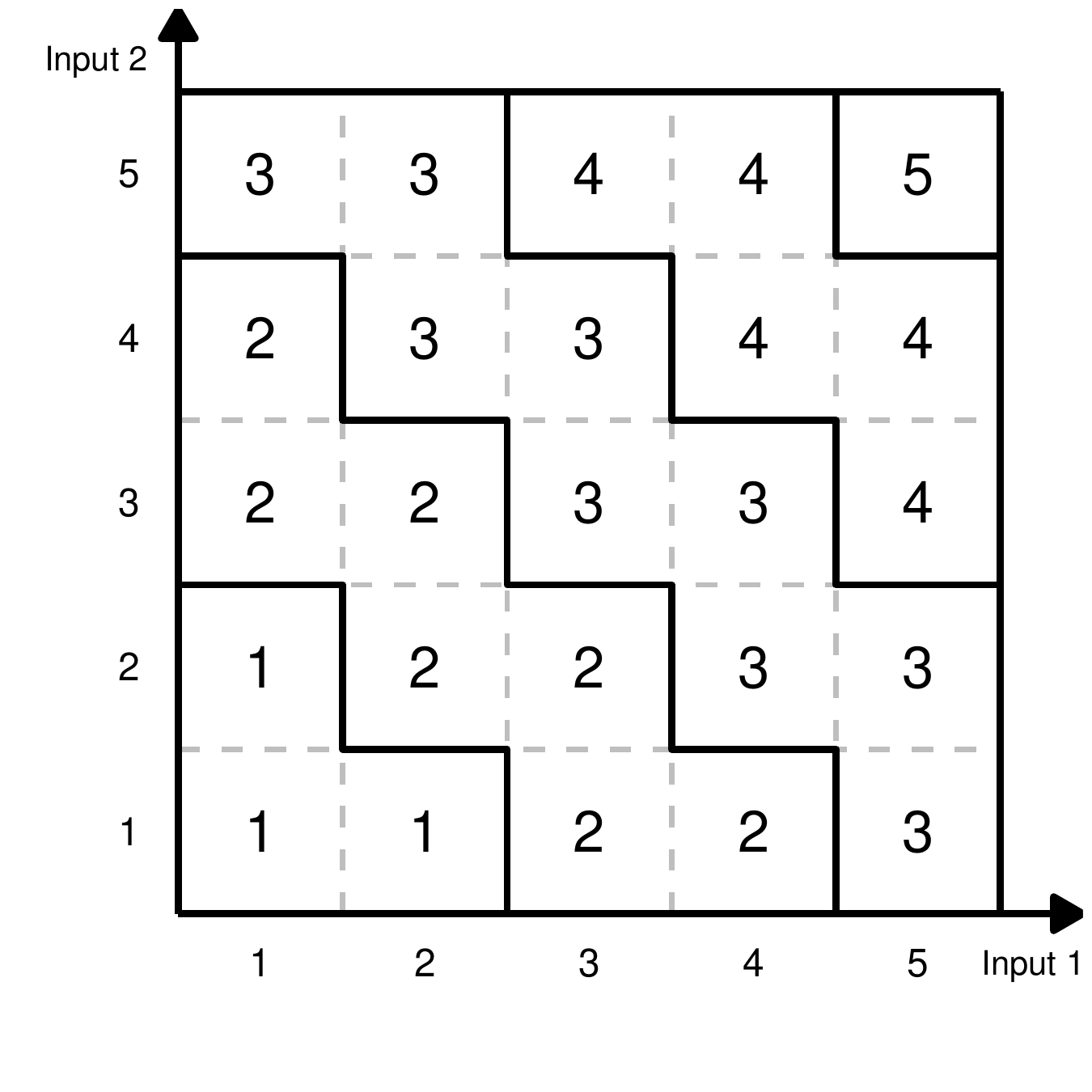}
    \caption{The two-way table used in the Hamlet Evaluation System for
      aggregating two input scores.  The $(i,j)$ element in the table
      above shows the output score when the first input is $i$ and the
      second input is $j$.}
    \label{fig:hes_table}
  \end{figure}

\begin{figure}[H]
  \includegraphics[width=0.4\textwidth, page=1]{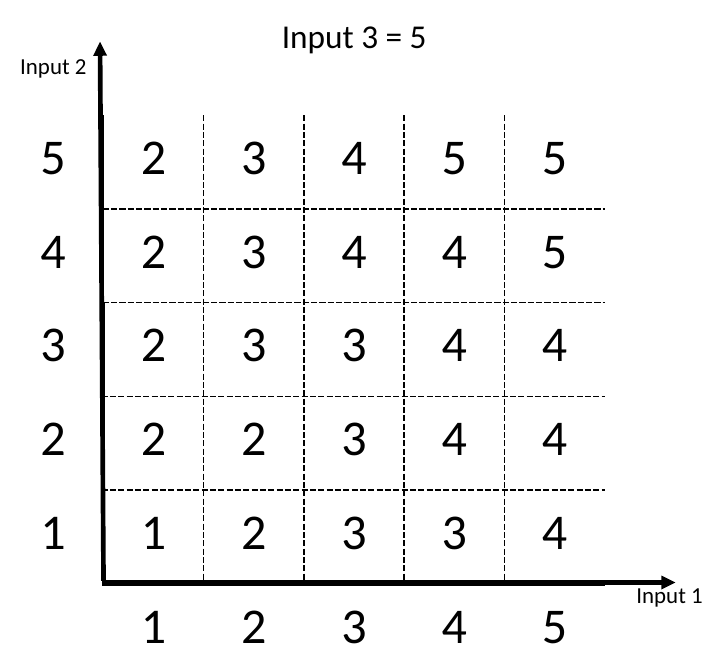}
  \includegraphics[width=0.4\textwidth, page=2]{figs/3waytable.pdf}
  \includegraphics[width=0.4\textwidth, page=3]{figs/3waytable.pdf}
  \includegraphics[width=0.4\textwidth, page=4]{figs/3waytable.pdf}
  \includegraphics[width=0.4\textwidth, page=5]{figs/3waytable.pdf}
  \caption{Three-way decision tables used in the HES. Each figure fixes the third input score and shows the output score for different combination of the first two inputs.}
  \label{fig:3waytable}
\end{figure}

\section{Additional summaries of the HES}

\begin{figure}[H]
  \includegraphics[width=0.5\textwidth]{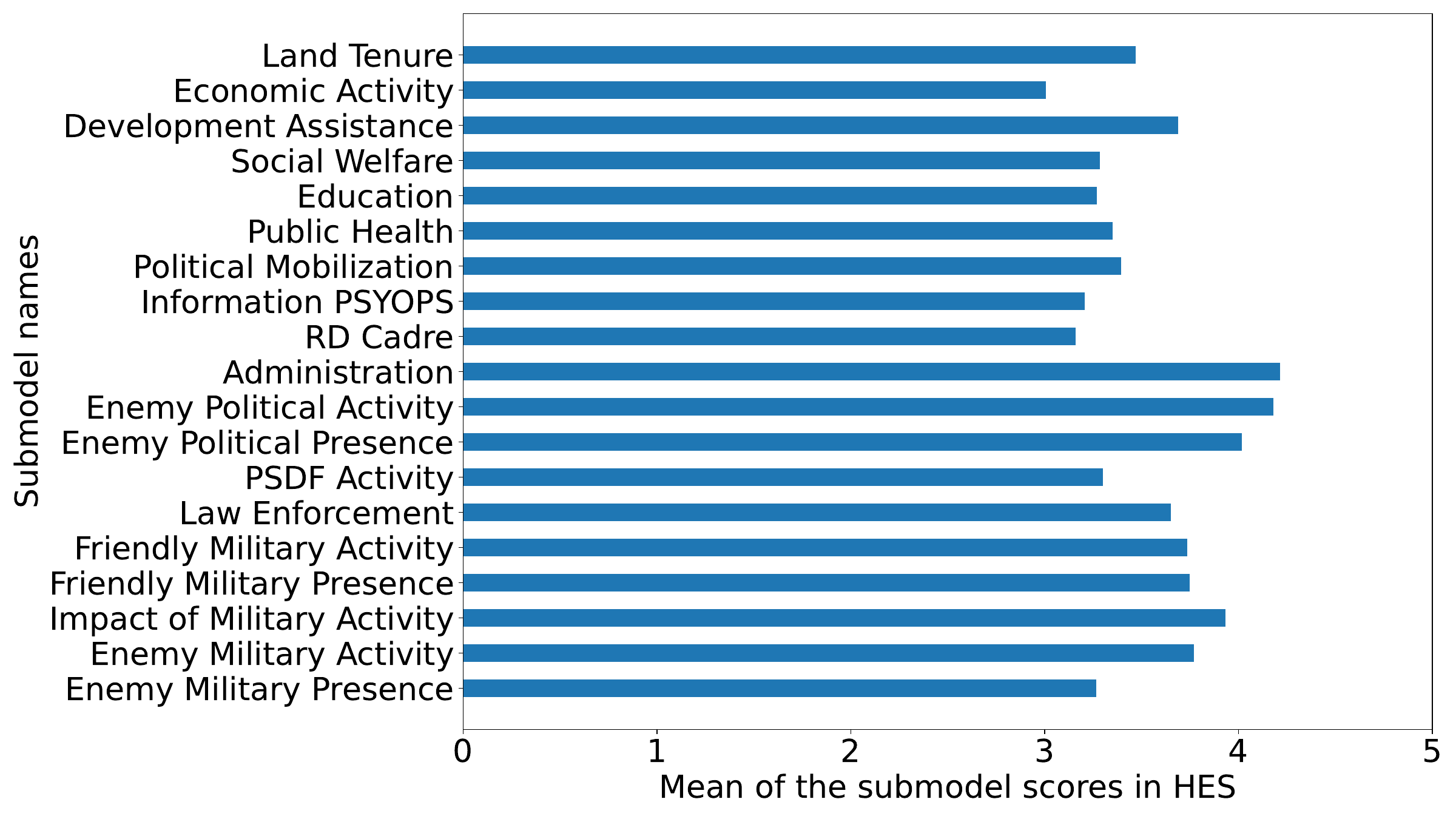}
  \caption{The mean of each sub-model score across the 1954 regions}%
  \label{fig:mean_submodel_scores}
\end{figure}

\begin{table}[H]
    \begin{tabular}{ll}
    Variable                         & Value \\ \hline
    Number of regions                & 1954  \\
    Number of regions being attacked & 1024  \\
    Average security score           & 3.34  \\
    Average regional safety outcome  & 0.53  \\
    Average regional economy outcome & 0.51  \\
    Average civic society outcome    & 0.43  \\ \hline
    \end{tabular}%
      \caption{Summary statistics on key measures}%
      \label{tab:summary}
  \end{table}

  \section{Proof of Theorem~\ref{th1}}
  \label{pf1}
  
 We wish to show
  that the original chance-constraint optimization problem (see
  Equation~\eqref{eq:safe_bayes})
      \begin{align}
          \delta_{\text{safe}} = \argmax_{\delta \in \Delta}\ \ \
        &\mathbb{E}_{}\left[V(\delta;\bTheta) \mid \{Z_i\}_{i=1}^n\right], \label{pf1:11} \\
          \text{subject to\ \ \ }
        &\mathbb{E}\left[R(\delta,\tilde\delta;\bTheta)\mid \{Z_i\}_{i=1}^n\right]\le \epsilon, \label{pf1:12}
      \end{align}
      can be equivalently written as,
          \begin{align}
              \delta_{\text{safe}} = &\argmax_{\delta \in \Delta}\ \ \ \sum_{k=0}^{K-1}\int  I(\delta(\bx)=k)b_k(\bx)dF(\bx) \label{pf1:21}\\
      &\text{subject to \ \ } \sum_{k=0}^{K-1} \int  I(\delta(\bx)=k)r_k(\bx)dF(\bx) \le \epsilon, \label{pf1:22}
          \end{align}
          where $F(\bx)$ be the CDF of the covariate $\bX$, and
  \begin{equation*}
      \begin{aligned}
          \tau_k(\bx,\btheta)&:=\mathbb{E}\left[u(k,Y(k))-u(\tilde\delta(\bx),Y(\tilde\delta(\bx)))
            \ \bigr | \ \bX=\bx,\bTheta=\btheta \right],\\
          b_k(\bx) &:= \mathbb{E}\left[\tau_k(\bx,\bTheta)\mid \{Z_i\}_{i=1}^n \right],\\
          r_k(\bx) &:= \mathbb{P}\left(\tau_k(\bx,\bTheta)<0\mid \{Z_i\}_{i=1}^n \right).
      \end{aligned}
  \end{equation*}
  
  We first show that the constraint given in Equation~\eqref{pf1:12}
  is equivalent to Equation~\eqref{pf1:22}.  By the definition of the
  posterior average conditional risk, we have
  \begin{align}
    R(\delta,\tilde\delta;\bTheta) &= \mathbb{P}\left(
      \mathbb{E}\Big[u( \delta(\bX),Y( \delta(\bX))) \ \bigr | \
      \bX,\bTheta \Big]< \mathbb{E}\left[u(\tilde
        \delta(\bX),Y(\tilde \delta(\bX))) \ \bigr | \ \bX,
        \bTheta\right] \ \biggr | \ \bTheta \right) \nonumber \\ 
        &= \mathbb{P}\left(
          \mathbb{E}\Big[u( \delta(\bX),Y( \delta(\bX))) - u(\tilde
          \delta(\bX),Y(\tilde \delta(\bX)))\ \bigr | \
          \bX,\bTheta \Big]< 0 \ \biggr | \ \bTheta \right) \nonumber \\
        &=  \mathbb{P}\left(\sum_{k=0}^{K-1}\tau_k(\bX,\bTheta)I(\delta(\bX)=k)
            < 0 \ \biggr | \ \bTheta \right) \label{Equ: ACRisk}
  \end{align}
  Therefore, Equation~\eqref{pf1:12} can be written as follows.  For
  clarity, we use subscripts to explicitly indicate the random
  variables with respect to which each expectation is taken.
  \begin{align*}
      &\ \ \ \ \ \mathbb{E}_{\bTheta}\left[R(\delta,\tilde\delta;\bTheta)\ \bigr | \
        \{Z_i\}_{i=1}^n\right] \\
      &= \
      \mathbb{E}_{\bTheta}\left[\mathbb{P}_{\bX}\left(\sum_{k=0}^{K-1}\tau_k(\bX,\bTheta)I(\delta(\bX)=k)<0
          \ \biggl | \
          \bTheta\right)\ \biggl | \ \{Z_i\}_{i=1}^n\right]\\
      &= \ \mathbb{E}_{\bTheta}\left [ \mathbb{E}_{\bX}\left
        [I\left\{\sum_{k=0}^{K-1}\tau_k(\bX,\bTheta)I(\delta(\bX)=k)<0\right\}
        \  \biggr | \ \bTheta\biggl]
         \ \right | \ \{Z_i\}_{i=1}^n\right] \\
      &= \ \mathbb{E}_{\bX,\bTheta}\left[ I\left\{\sum_{k=0}^{K-1}\tau_k(\bX,\bTheta)I(\delta(\bX)=k)<0 \right\}\
      \biggl | \ \{Z_i\}_{i=1}^n\right]\\
      &= \ \mathbb{E}_{\bX}\left [ \mathbb{E}_{\bTheta}\left
      [I\left\{\sum_{k=0}^{K-1}\tau_k(\bX,\bTheta)I(\delta(\bX)=k)<0\right\}
      \ \biggr | \ \{Z_i\}_{i=1}^n, \bX \right]
   \ \Biggl
   | \ \{Z_i\}_{i=1}^n\right] \\
   &= \ \mathbb{E}_{\bX}\left [\sum_{k=0}^{K-1}I(\delta(\bX)=k) \mathbb{E}_{\bTheta}\left
   [I\left\{\tau_k(\bX,\bTheta)<0\right\}
   \ \biggr | \ \{Z_i\}_{i=1}^n, \bX \right]
\ \Biggl
| \ \{Z_i\}_{i=1}^n\right] \\
&=\ \mathbb{E}_{\bX}\left[\sum_{k=0}^{K-1}I(\delta(\bX)=k)r_k(\bX)\ \biggl |
   \ \{Z_i\}_{i=1}^n\right] \\
&= \ \sum_{k=0}^{K-1} \int  I(\delta(\bx)=k)r_k(\bx)dF(\bx),
  \end{align*} 
  where the third and fourth equalities follow from the law of
  iterated expectation, and the fifth equality uses the fact that
  there is only one non-zero term in the summation term
  $\sum_{k=0}^{K-1}\tau_k(\bX,\bTheta)I(\delta(\bX)=k)$, leading to
  the simplification of the indicator function.
  
  Similarly, for the optimization target, we have
  \begin{align}
      &\mathbb{E}_{}\bigr[V(\delta;\bTheta)\mid\{Z_i\}_{i=1}^n\bigr ]\nonumber\\
      = \ &\mathbb{E}_{}\bigr[u(\delta(\bX),Y(\delta(\bX)))\mid \{Z_i\}_{i=1}^n\bigr ] \label{pf1:41}\\
      = \
      &\mathbb{E}_{\bX}\left[\mathbb{E}_{Y}\left\{u(\delta(\bX),Y(\delta(\bX)))\mid
        \{Z_i\}_{i=1}^n, \bX\right\} \ \Bigr | \ \{Z_i\}_{i=1}^n\right ]\nonumber\\
      = \ &\mathbb{E}_{\bX}\left[\mathbb{E}_{Y}\left\{\sum_{k=0}^{K-1}u(\delta(\bX),Y(k))I(\delta(\bX)=k)\
         \biggl | \ \{Z_i\}_{i=1}^n, \bX\right\}\ \biggl | \ \{Z_i\}_{i=1}^n\right]\nonumber\\
      = \ &\mathbb{E}_{\bX}\left[\mathbb{E}_{\bTheta}\left\{\sum_{k=0}^{K-1}\tau_k(\bX,\bTheta)I(\delta(\bX)=k)\
         \biggl | \ \{Z_i\}_{i=1}^n, \bX\right\}\ \biggl | \ \{Z_i\}_{i=1}^n\right] + \text{const.}\nonumber\\
      = \ &\mathbb{E}_{\bX}\left[\sum_{k=0}^{K-1}I(\delta(\bX=k)b_k(\bX)\
            \biggr | \ \{Z_i\}_{i=1}^n\right]+\text{const.} \nonumber\\
      = \ &\sum_{k=0}^{K-1} \int  I(\delta(\bx)=k)b_k(\bx)dF(\bx) +
         \text{const.} \nonumber
  \end{align}
  where $\text{const.}$ represents the term that is not a function of
  $\delta$.  In Equation~\eqref{pf1:41}, the expectation is taken over
  the distribution of $\bX$, the posterior distribution of $\bTheta$,
  and the conditional distribution of $Y$ given $\bX,\bTheta$.  This
  completes the proof.  \qed
  
\section{BART and GP for Bayesian Inference on the CATE}
\label{sec:bartgp}

Below, we consider BART and GP for Bayesian inference on these
parameters $\{f_k\}_{k=0}^{K-1}$.  Specifically, we sample from the
posterior distribution of $\{f_k\}_{k=0}^{K-1}$ and apply
Algorithm~\ref{alg:mcmc} to find a safe policy.

\paragraph{Bayesian Additive Regression Trees (BART).}
BART is a popular Bayesian nonparametric model that is commonly used
for causal inference, especially to estimate the CATE
\citep{taddy2016nonparametric,hahn2020bayesian}.  In general, BART
excels in learning complex nonlinear relations while it is often poor
at extrapolating.  Thus, BART may be a suitable choice when there
exists a substantial covariate overlap between treatment conditions.

We use a BART to model each $f_k$ for $k\in \{0, 1,\ldots, K-1\}$ as
the sum of $L$ regression trees, i.e.,
$f_k(\bx) = \sum_{\ell=1}^L g_{k\ell}(\bx;T_{k\ell},P_{k\ell})$ where
$g_{k\ell}(\cdot)$ is the $\ell$-th regression tree with parameter
$T_{k\ell}$ and $P_{k\ell}$ denoting the structure of the regression
tree and the parameters in the terminal nodes, respectively.  Thus,
the parameter $\bTheta$ consists of
$\{T_{k\ell},P_{k\ell}\}_{1\le \ell\le L, 0\le k\le K-1}$ as well as
$\sigma^2$.  We draw posterior samples for
$\{T_{k\ell},P_{k\ell}\}_{1\le \ell\le L}$ using an MCMC algorithm
once a prior distribution is specified \citep{chipman2010bart}.

\paragraph{Gaussian Process Regression.}  Another popular Bayesian
nonparametric model is Gaussian Process regression (GP).  GP has a
greater degree of smoothness than BART, making it more suitable for
extrapolation \citep{rasmussen2003gaussian,branson2019nonparametric}.
Therefore, we should consider using GP when the overlap of covariates
between treatment conditions is poor.

As in the case of BART, we use a GP to model each $f_k$. Specifically,
$f_k$ is assumed to be a random function based on a collection of
Gaussian processes. To conduct Bayesian inference on $f_k$, we specify
a prior for $f_k$ by giving the mean function $\mu_k(\cdot)$ and
kernel function $K_k(\cdot,\cdot)$ and obtain posterior samples of
$f_k$ using the MCMC algorithm
\citep{rasmussen2003gaussian,branson2019nonparametric}.

When strong prior information is unavailable, we can set
$\mu_k(\cdot)=0$ for $k \ge 1$ which corresponds to no treatment
effect.  For the kernel function, which determines the covariance
between $f_k(\bx_1), f_k(\bx_2)$ for any $\bx_0,\bx_1\in \mathcal{X}$,
we can, for example, use Matern kernels:
\begin{align}
  K_{\text {Matern}}(\bx_1,\bx_2) & =
                                \sigma_0^2\frac{2^{1-\nu}}{\Gamma(\nu)}\left(\frac{\sqrt{2
                                \nu} ||\bx_1-\bx_2||}{\ell}\right)^{\nu}
                                B_{\nu}\left(\frac{\sqrt{2 \nu}
                                ||\bx_1-\bx_2||}{\ell}\right) \label{eq:matern}
\end{align}    
where $l$ is the scale parameter, $\sigma_0^2$ is the variance
parameter, $B_\nu$ is the modified Bessel function of the second kind,
and $\nu$ is the smoothness parameter.  The hyperparameters in the
Matern kernels can be selected based on prior knowledge about the
smoothness of $f_k(\cdot)$.  For example, to make $f_k$ more smooth,
we can increase the scale and smoothness parameters.  In general,
$\ell$ and $\nu$ determine the prior knowledge about the smoothness of
$f$, while $\sigma_0^2$ determines the strength of this prior
knowledge.

Extrapolation based on GP is similar to frequentist
extrapolation methods that specify the model class by assuming a
certain type of smoothness on the CATE under the robust optimization
framework.  For example, \cite{ben2021safe} considers the case with
two arms and assumes a Lipschitz constraint on the CATE, i.e.,
$|f_1(\bx_1)-f_1(\bx_2)|\le c|\bx_1-\bx_2|$.  In our framework, if we
specify the prior of $f_1(\bx)$ as a GP with mean function $m(\bx)$
that is $c_1$ Lipschitz, then the Matern kernel with scale parameter $\ell$ and
smoothness parameter $\nu$ implies the following probabilistic
Lipschitz condition:

$$
\mathbb{P}\left(|f_1(\bx_1)-f_1(\bx_2)|>c_2||\bx_1-\bx_2||\right) \le \sigma_0^2\left\{\left(1+\frac{1}{\nu-1}\right)\frac{1}{c_2^2l^2}+ \frac{c_1^2}{c_2^2}\right\}
$$
Thus, there exists a direct relationship between the prior hyperparameter
of GP and the smoothness of the underlying model.

\newpage

\section{Additional simulation results}
\label{sec:addsimu}

Here, we present additional detailed simulation results.

\subsection{Average value and ACRisk with different signal strength and prior strength}

\begin{figure}[H]
  \centering
  \begin{subfigure}[H]{0.495\textwidth}
    \includegraphics[width=\textwidth]{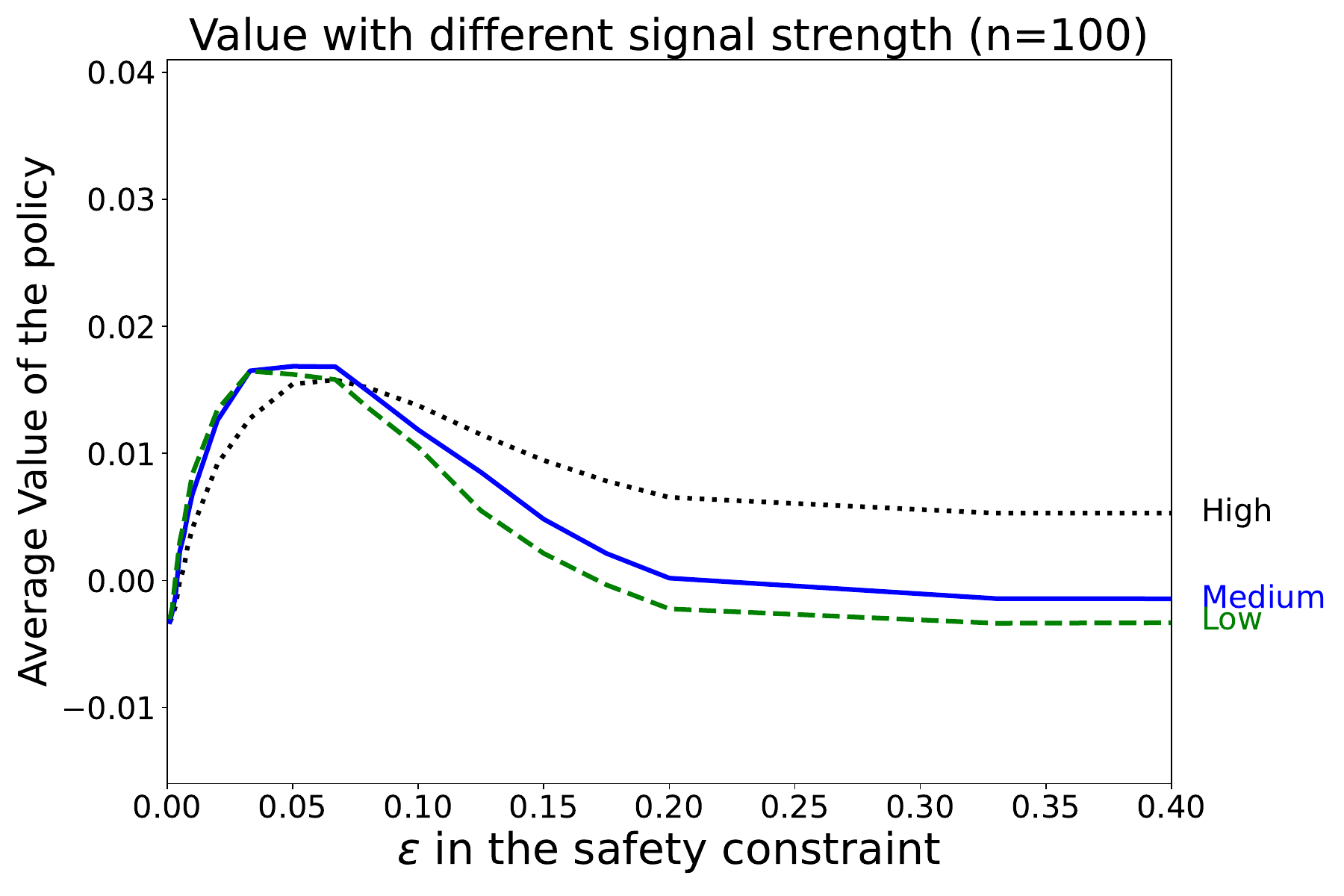} 
    \caption{Average Value}
    \label{fig:simu21}
  \end{subfigure}
  \begin{subfigure}[H]{0.495\textwidth}
    \includegraphics[width=\textwidth]{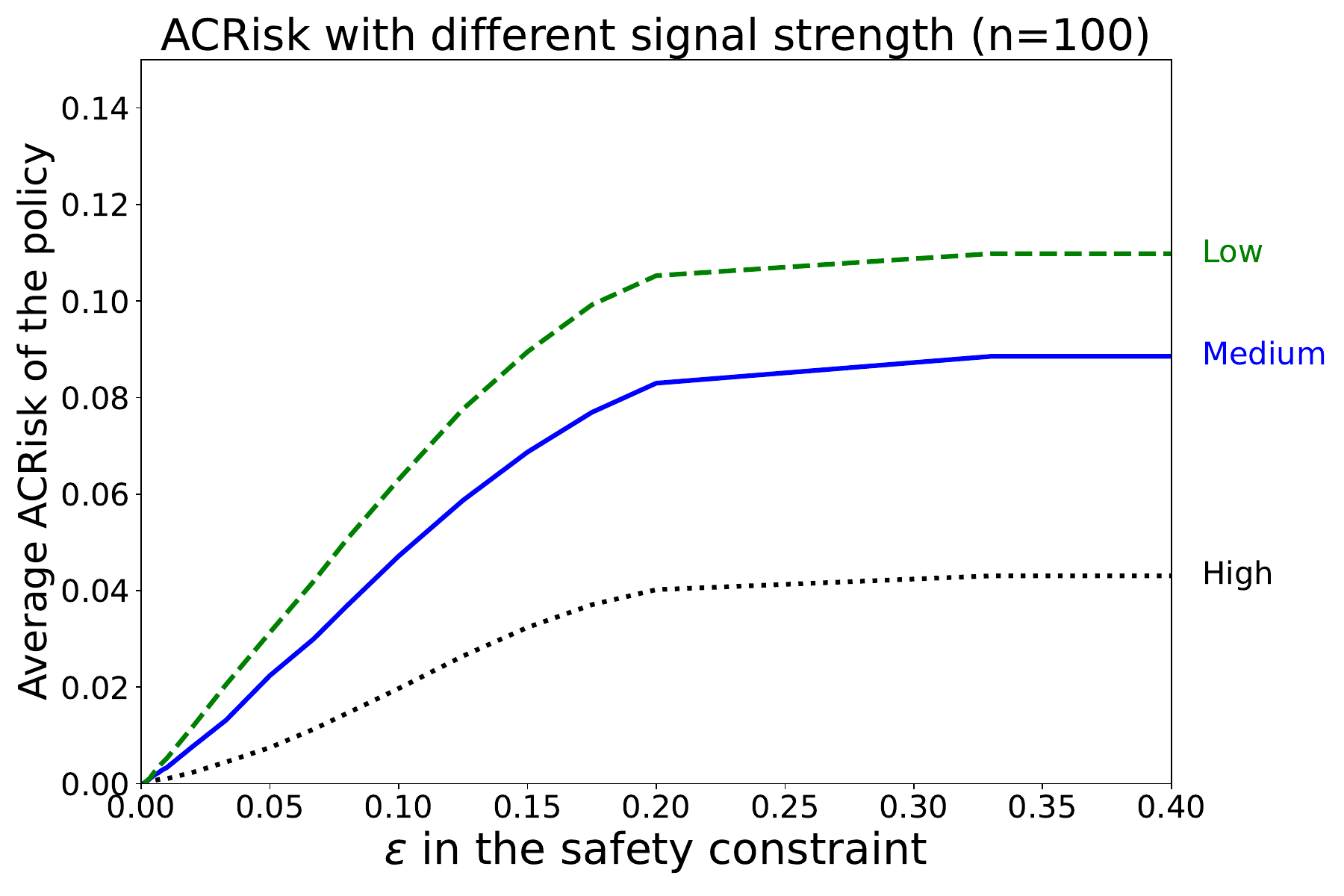}
    \caption{Average ACRisk}
    \label{fig:simu22}
  \end{subfigure}
  \caption{Average Value and ACRisk for learned policies using data with covariate overlap, varying the safety constraint and signal strength.}
  \label{fig:simu2}
\end{figure}

\begin{figure}[H]
  \centering
  \begin{subfigure}[H]{0.495\textwidth}
    \includegraphics[width=\textwidth]{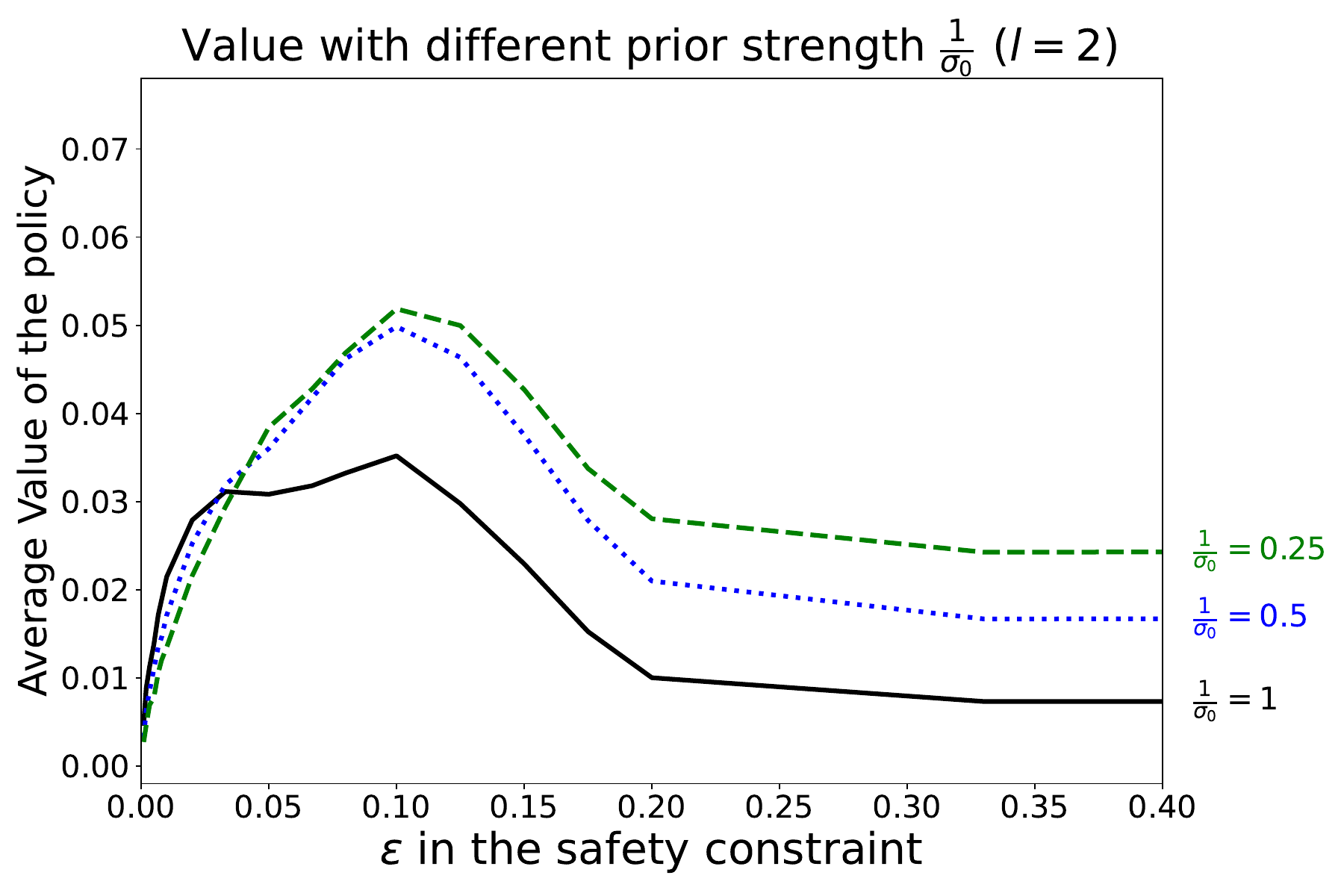}
    \caption{Average Value}
    \label{fig:simu41}
  \end{subfigure}
  \begin{subfigure}[H]{0.495\textwidth}
    \includegraphics[width=\textwidth]{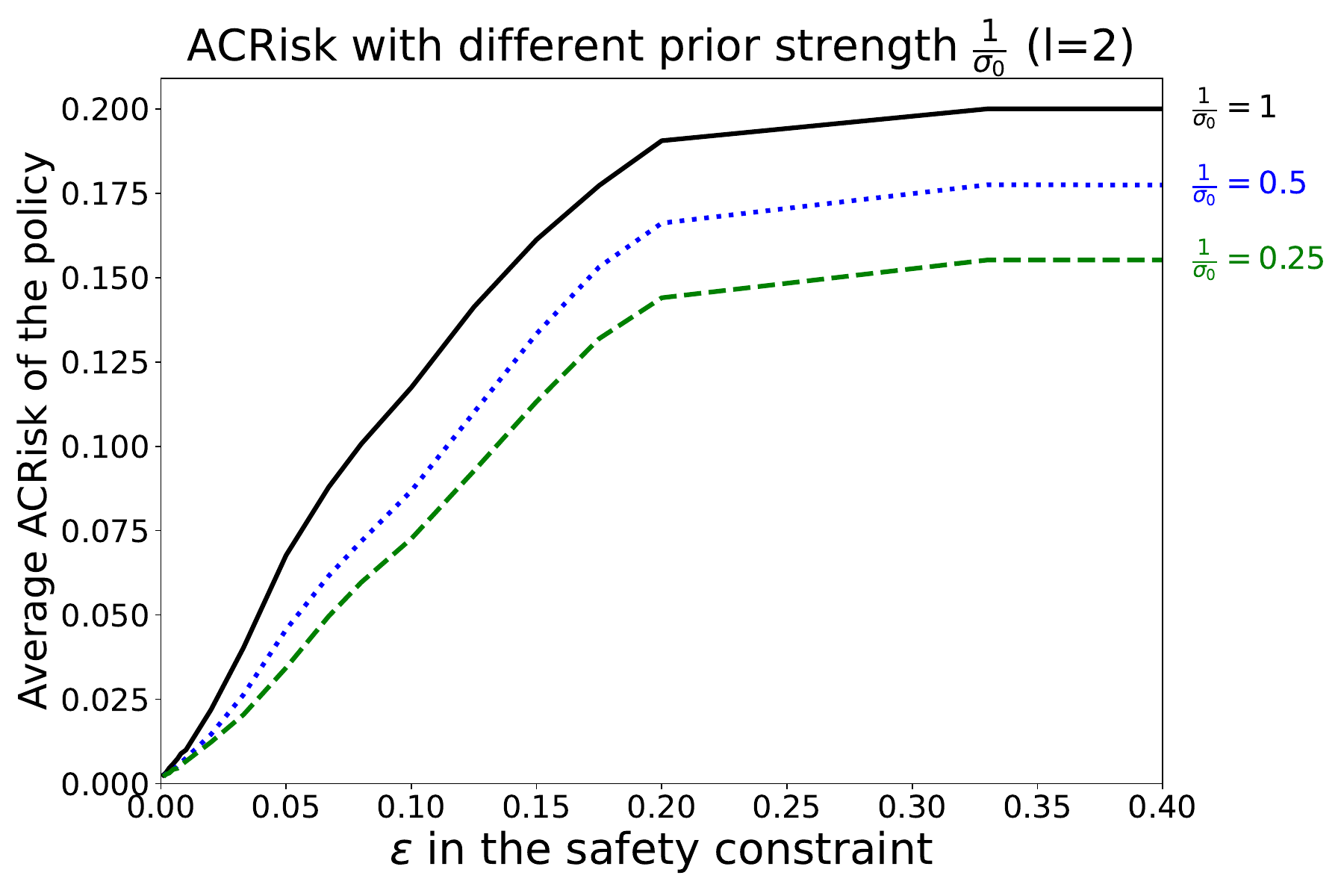}
    \caption{Average ACRisk}
    \label{fig:simu42}
  \end{subfigure}
  \caption{Average Value and ACRisk for learned policies using data without covariate overlap, varying the safety constraint and prior strength for the CATE.}
  \label{fig:simu4}
\end{figure}

\subsection{The tail distribution of the empirical ACRisk}

\begin{figure}[H]
  \centering
  \begin{subfigure}[H]{0.495\textwidth}
    \includegraphics[width=\textwidth]{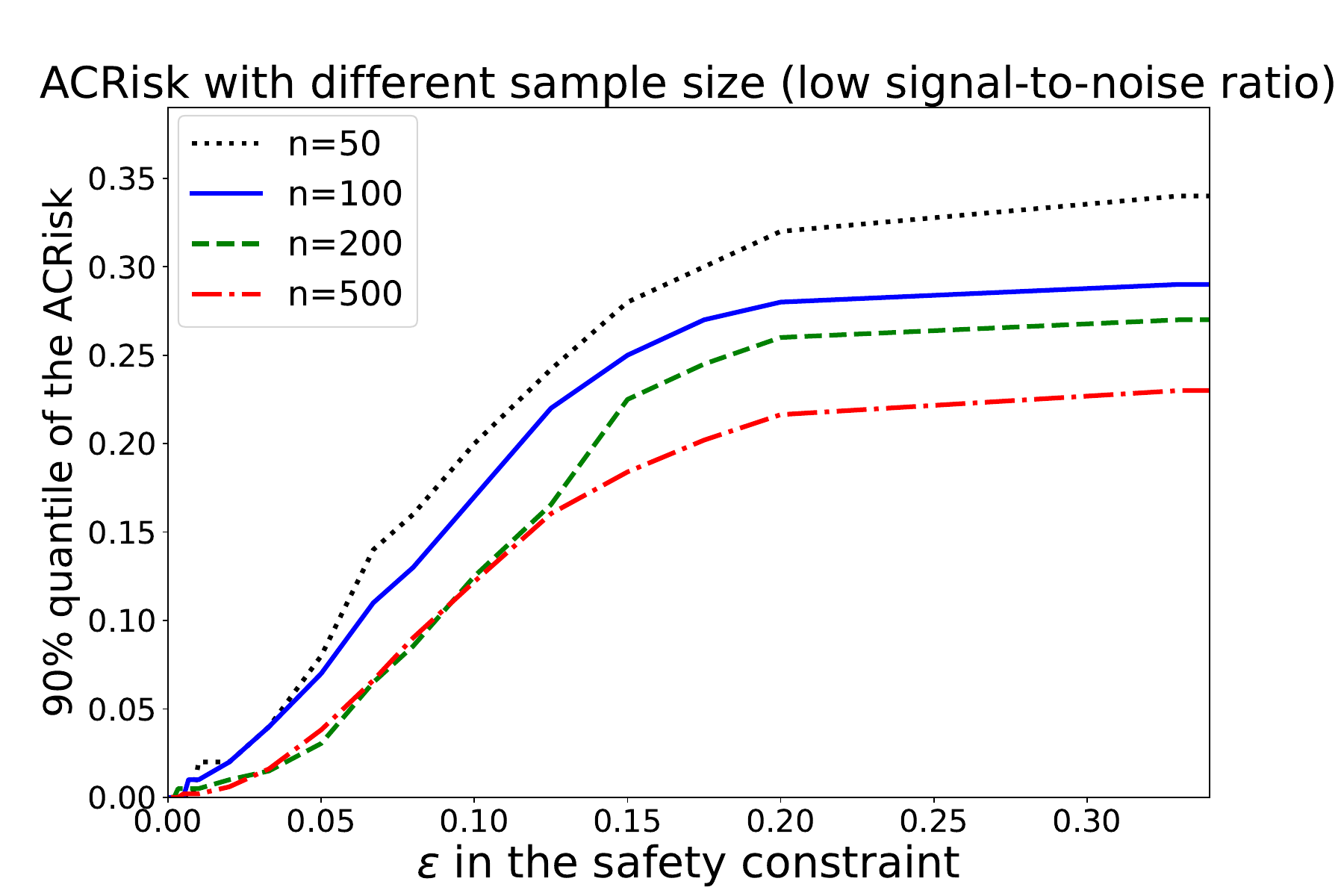}
    \caption{ACRisk with different sample size}
    \label{fig:simus11}
  \end{subfigure}
  \begin{subfigure}[H]{0.495\textwidth}
    \includegraphics[width=\textwidth]{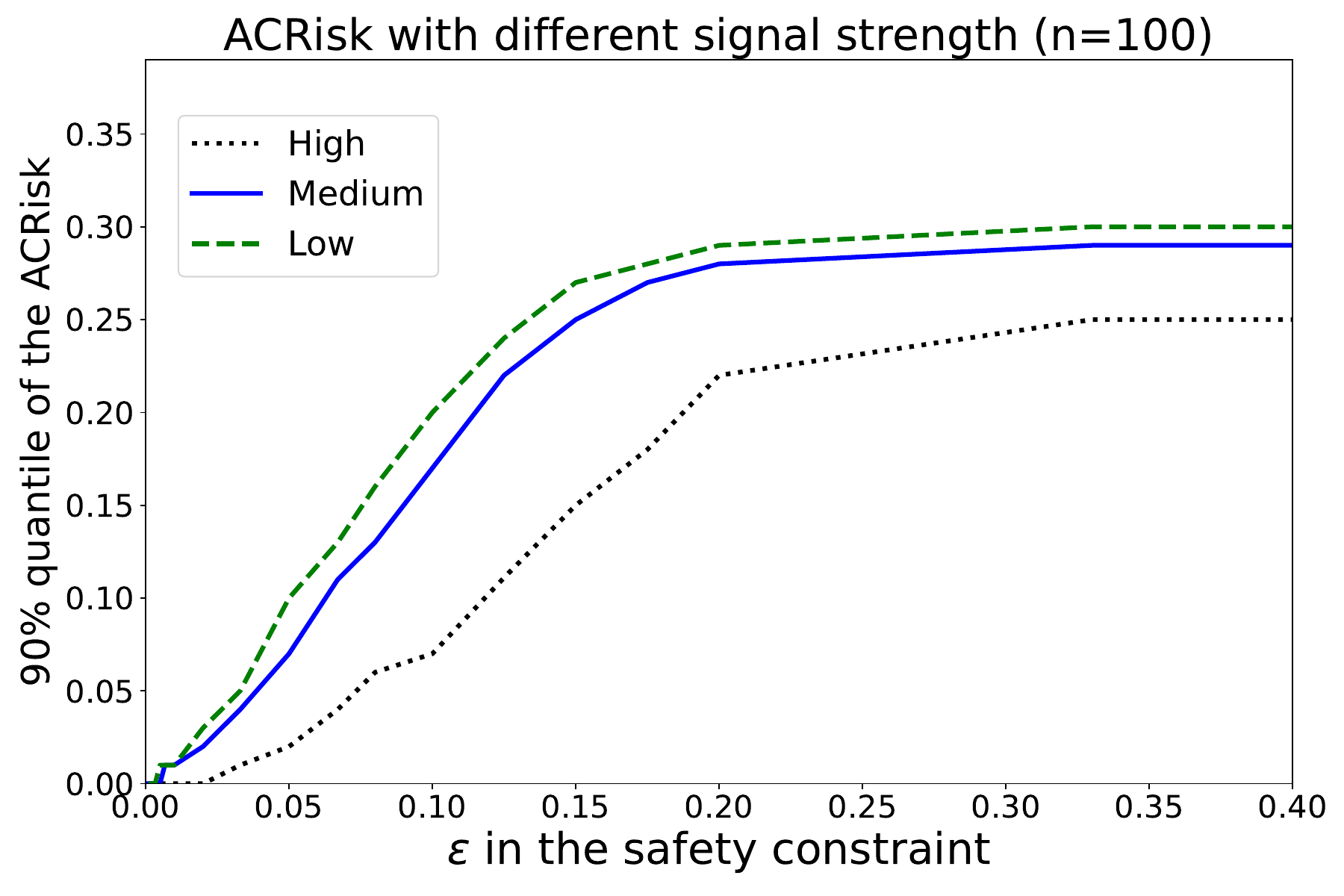}
    \caption{ACRisk with different signal strength}
    \label{fig:simus12}
  \end{subfigure}
  \caption{90 \% quantile of the ACRisk for learned policy among 2000 simulations. The CATE is estimated with BCF and covariates have overlap }
  \label{fig:simus1}
\end{figure}

\begin{figure}[H]
  \centering
  \begin{subfigure}[H]{0.495\textwidth}
    \includegraphics[width=\textwidth]{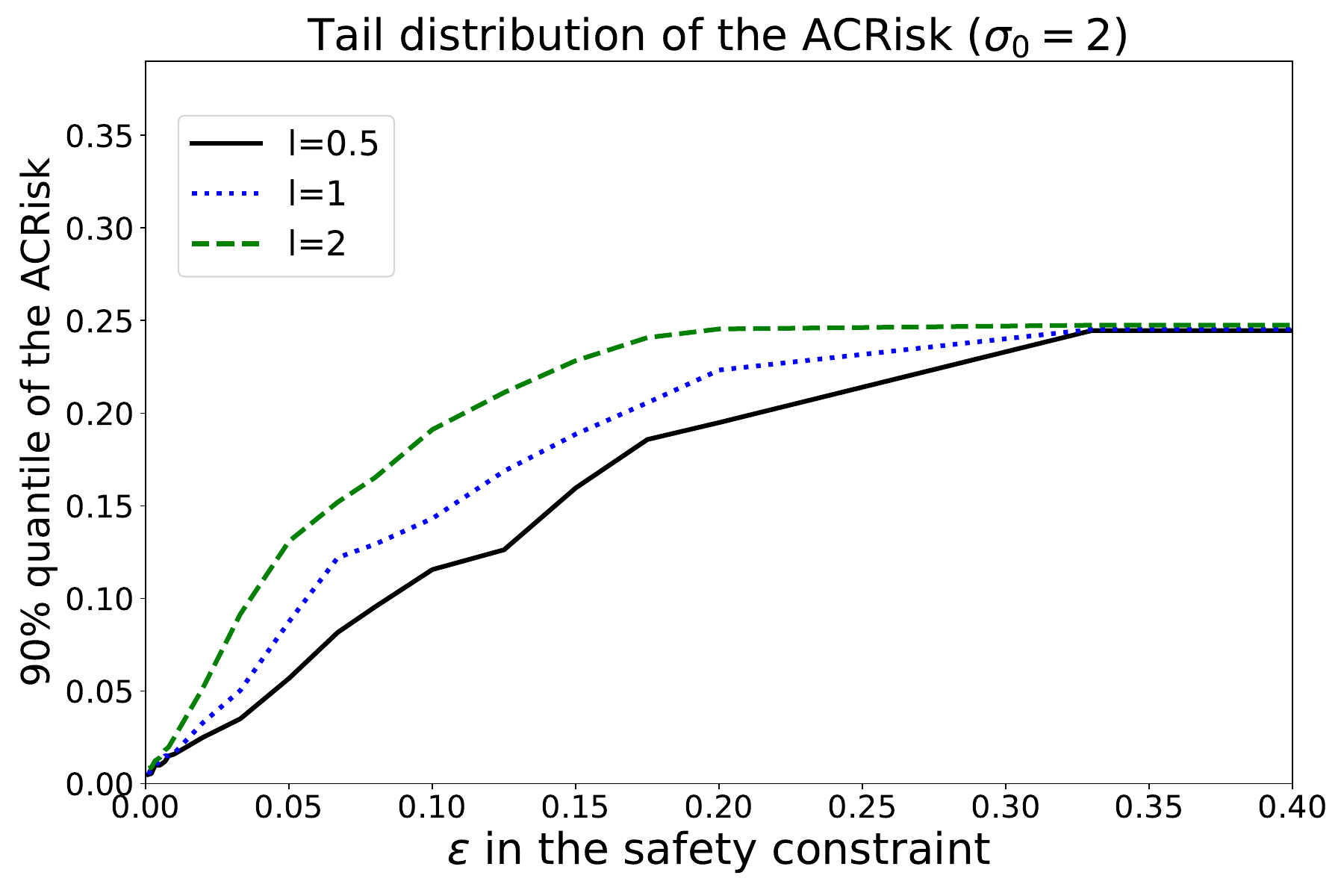}
    \caption{ACRisk with different smoothness}
    \label{fig:simus21}
  \end{subfigure}
  \begin{subfigure}[H]{0.495\textwidth}
    \includegraphics[width=\textwidth]{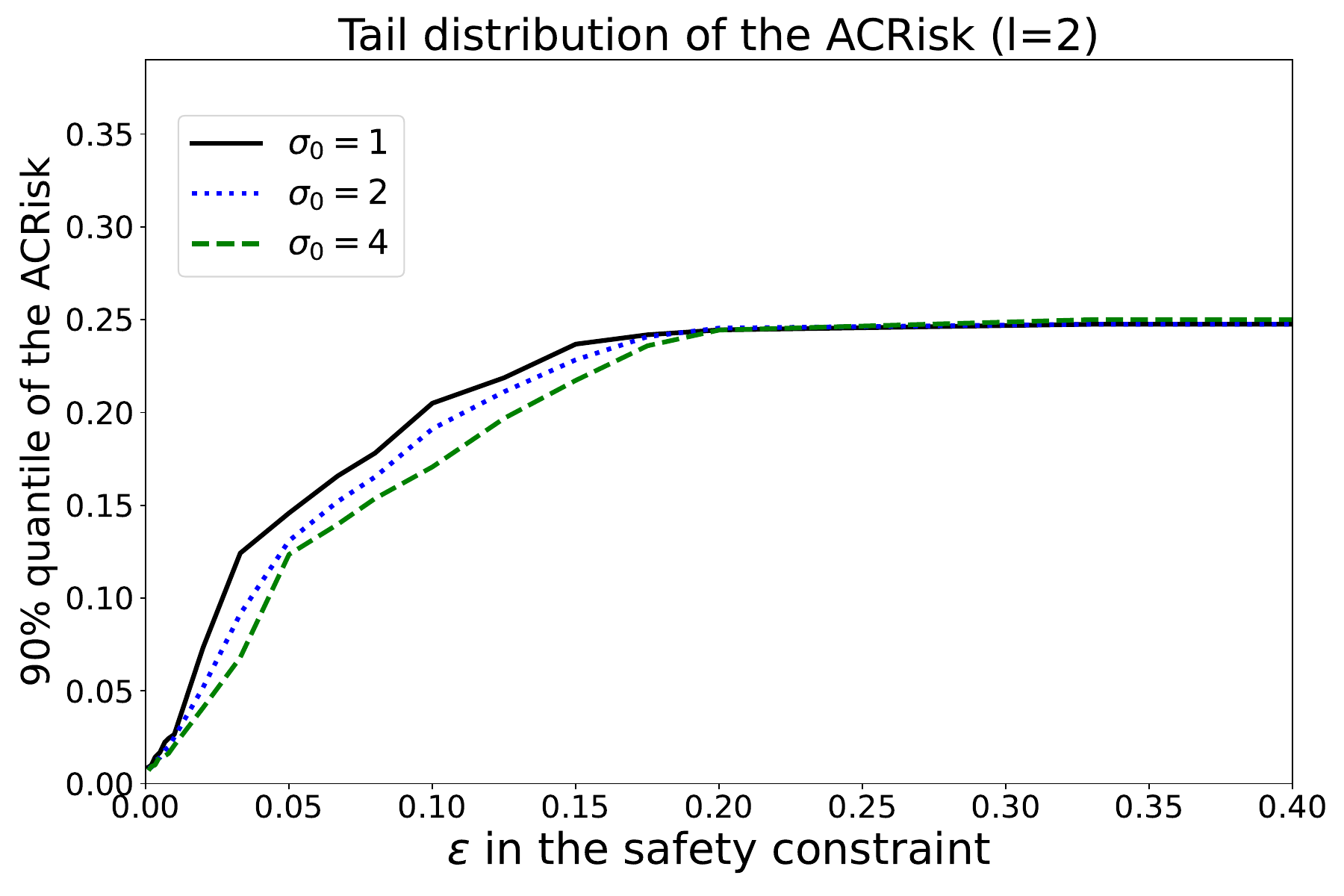}
    \caption{ACRisk with different prior strength}
    \label{fig:simus22}
  \end{subfigure}
  \caption{90\% quantile of the ACRisk for learned policy among 2000 simulations. The CATE is estimated with a GP and there is no covariate overlap. }
  \label{fig:simus2}
\end{figure}

In the main text, we discuss the average ACRisk of the learned policy
in multiple simulations. However, we may also be interested in the
tail distribution for the ACRisk of learned policy. Therefore, we
inspect the 90 percentile of the ACRisk for the learned policy across
2000 simulations. Figures~\ref{fig:simus1}~and~\ref{fig:simus2} show
how the 90 percentile of the ACRisk changes with the sample size,
signal strength, smoothness for prior of the CATE, and the strength of
the prior. 

Overall, we find the results similar to those presented in the main
text. In figure~\ref{fig:simus1}, the
90 percentile of the ACRisk increases as the safety constraint
$\epsilon$ increases before reaching a plateau that corresponds to the
ACRisk obtained by maximizing the posterior expected utility with no
constraint. A smaller sample size and lower signal-to-noise ratio lead
to a greater ACRisk. For the setup without covariate overlap (figure~\ref{fig:simus2}), we see a smoother and stronger prior increase the ACRisk because the extrapolation is more aggressive.

\subsection{Simulation results with binary outcomes}

The previous simulation design focused on continuous outcomes. Here,
we present simulation results when the outcome is binary.  
We use a similar setup as in Section~\ref{sec:numerical} where the
covariates $\bX = (X_1,X_2)$ and
$X_1,X_2 \overset{\text{i.i.d}}{\sim} Uniform[-1,1]$. We use the same
Scenario I (with covariate overlap) and Scenario II (without covariate
overlap) for generating the decision $D$ in the data. For the outcome,
we let
\begin{equation*}
  Y \mid \bX \sim \text{Bernoulli}\left(\text{expit}\left(\frac{X_1}{2} + \frac{X_2}{2} + \gamma\{3I(X_1>0,X_2>0)-\frac{3}{2}\}D|X_1||X_2|\right)\right), 
\end{equation*}
where we consider a strong signal case $\gamma=2$ and a weak signal
case $\gamma=1$. We vary the number of observations
$n\in \{50,100,200\}$, and we use a GP to model the CATE.
  
\begin{figure}[t!]
  \centering
  \begin{subfigure}[H]{0.495\textwidth}
    \includegraphics[width=\textwidth]{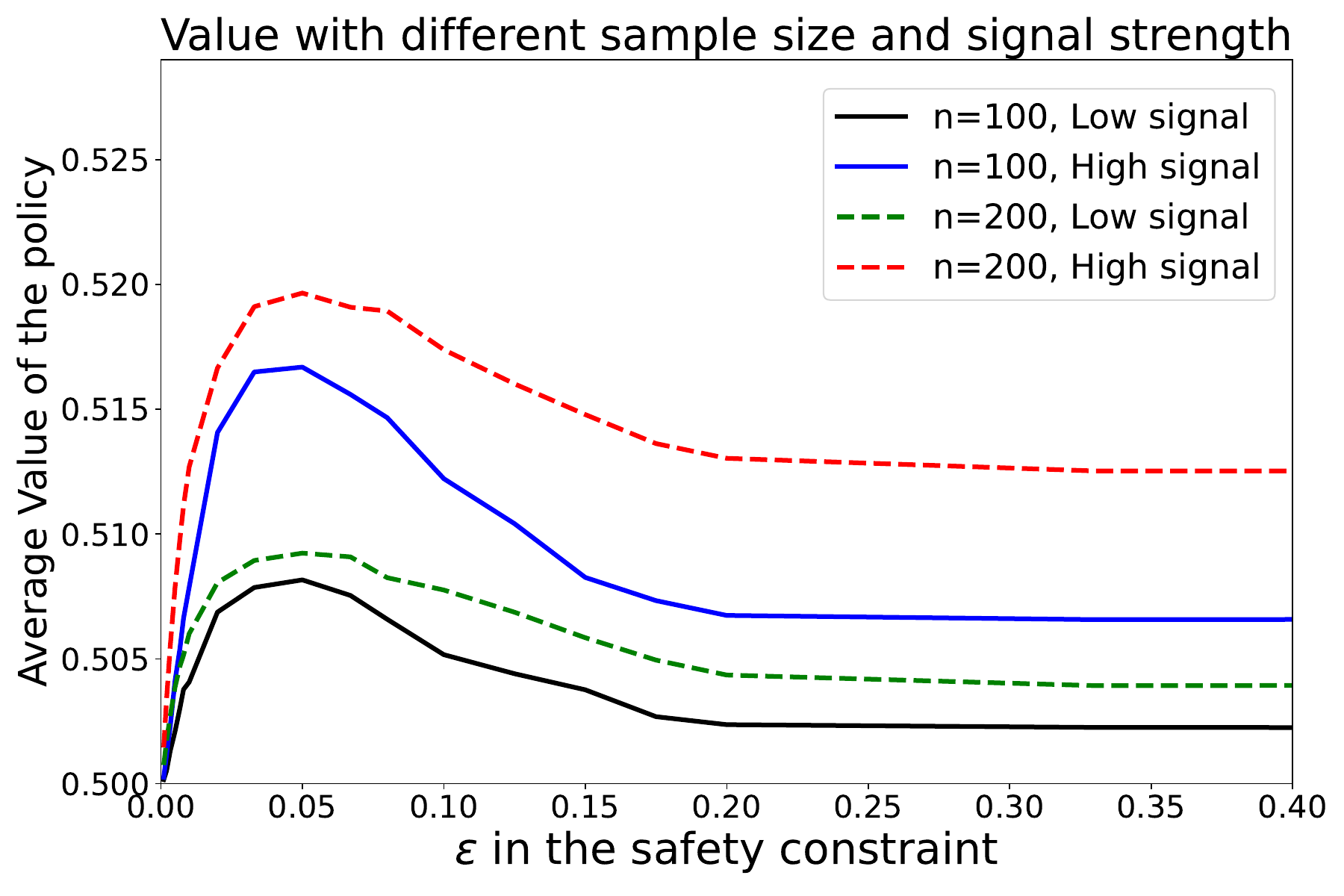}
    \caption{Average Value with different sample size and signal strength}
    \label{fig:simus31}
  \end{subfigure}
  \begin{subfigure}[H]{0.495\textwidth}
    \includegraphics[width=\textwidth]{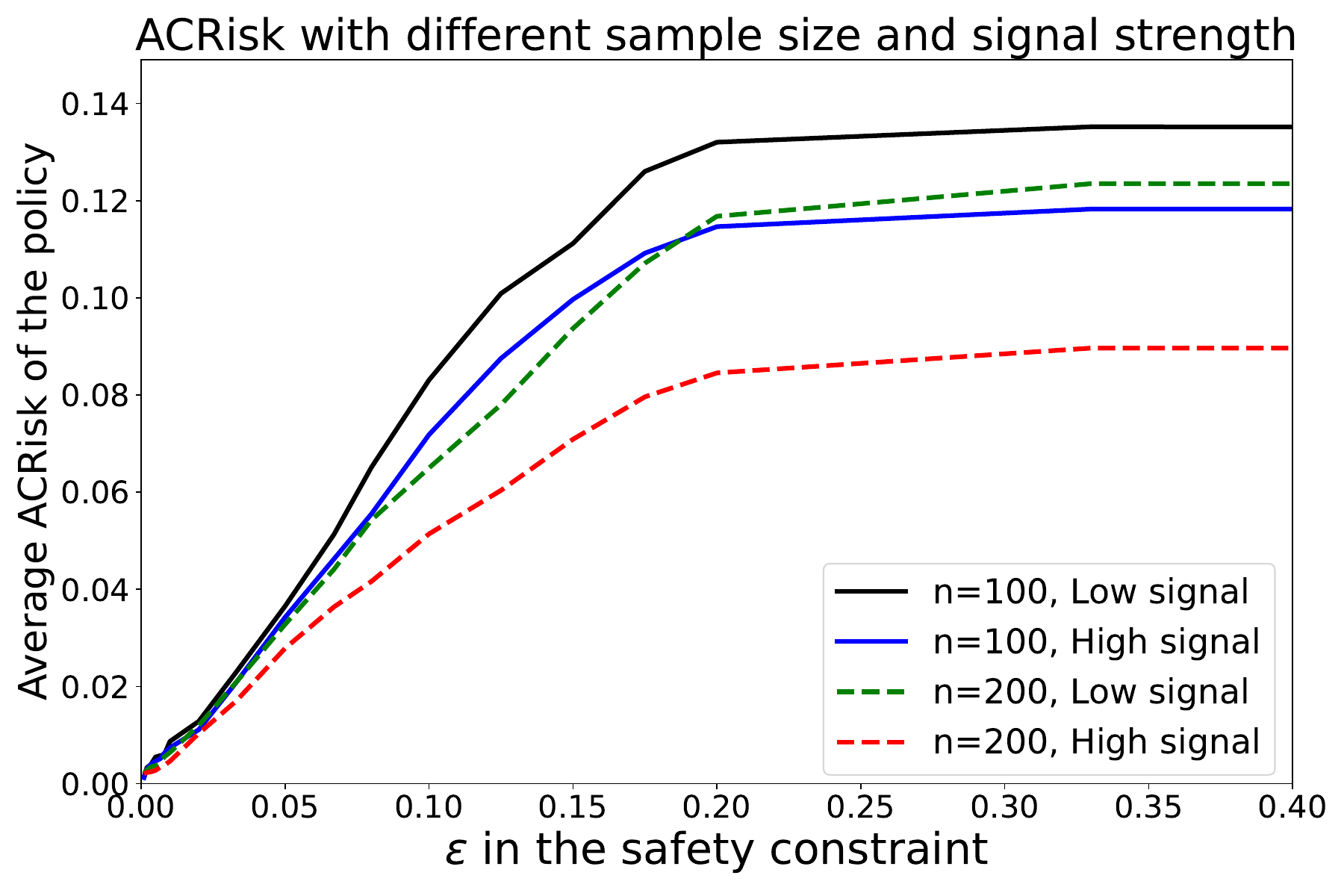}
    \caption{Average ACRisk with different sample size and signal strength}
    \label{fig:simus32}
  \end{subfigure}
  \caption{Average Value and ACRisk for the learned policy among 2000 simulations. The CATE is estimated with GP and covariates have overlap }
  \label{fig:simus3}
\end{figure}

\paragraph{With covariate overlap.} In the case with covariate overlap,
there is no need for extrapolation. Therefore, we use a weak prior for
the GP and specify the mean function $m(x)=0$, the kernel function as
a Matern kernel with $\sigma_0=4, l=0.5$.  We show how the average
ACRisk and the value changes as a function of the safety constraint
$\epsilon$ under different sample size and signal strength in
Figure~\ref{fig:simus3}.  

As shown in Figure~\ref{fig:simus3}, under all settings, the average
ACRisk increases as the safety constraint $\epsilon$ increases. Given
a fixed value of $\epsilon$, a larger sample size and stronger signal
lead to a lower ACRisk. Similar to the results in the main text, we
find a regularization effect of the safety constraint, where under an
appropriate value of the safety constraint, the average value of the
learned policy is greater than that of the policy that maximizes the
posterior expected utility without any safety constraint.

\begin{figure}[t!]
  \centering
  \begin{subfigure}[H]{0.495\textwidth}
    \includegraphics[width=\textwidth]{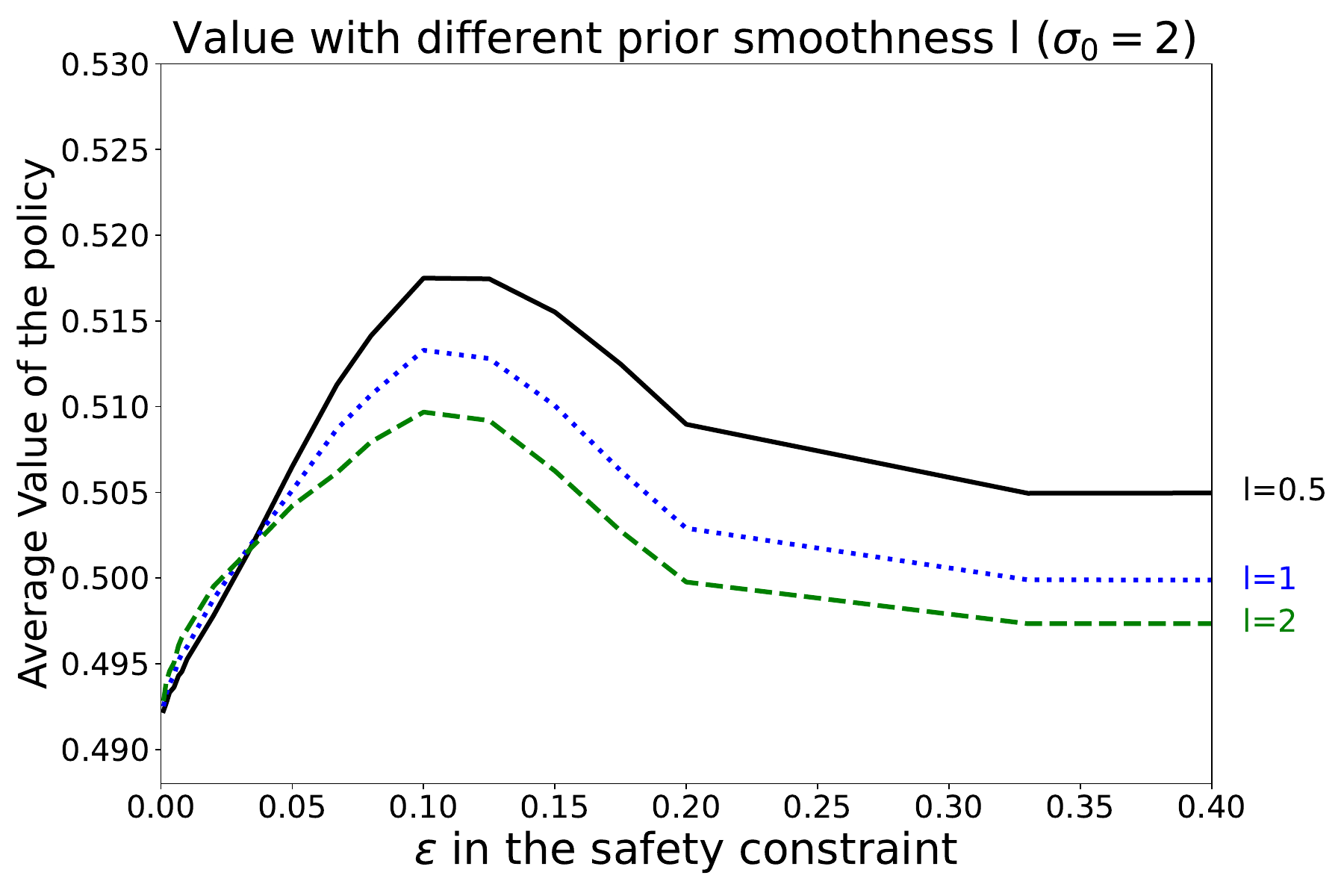}
    \caption{Average Value}
    \label{fig:simus41}
  \end{subfigure}
  \begin{subfigure}[H]{0.495\textwidth}
    \includegraphics[width=\textwidth]{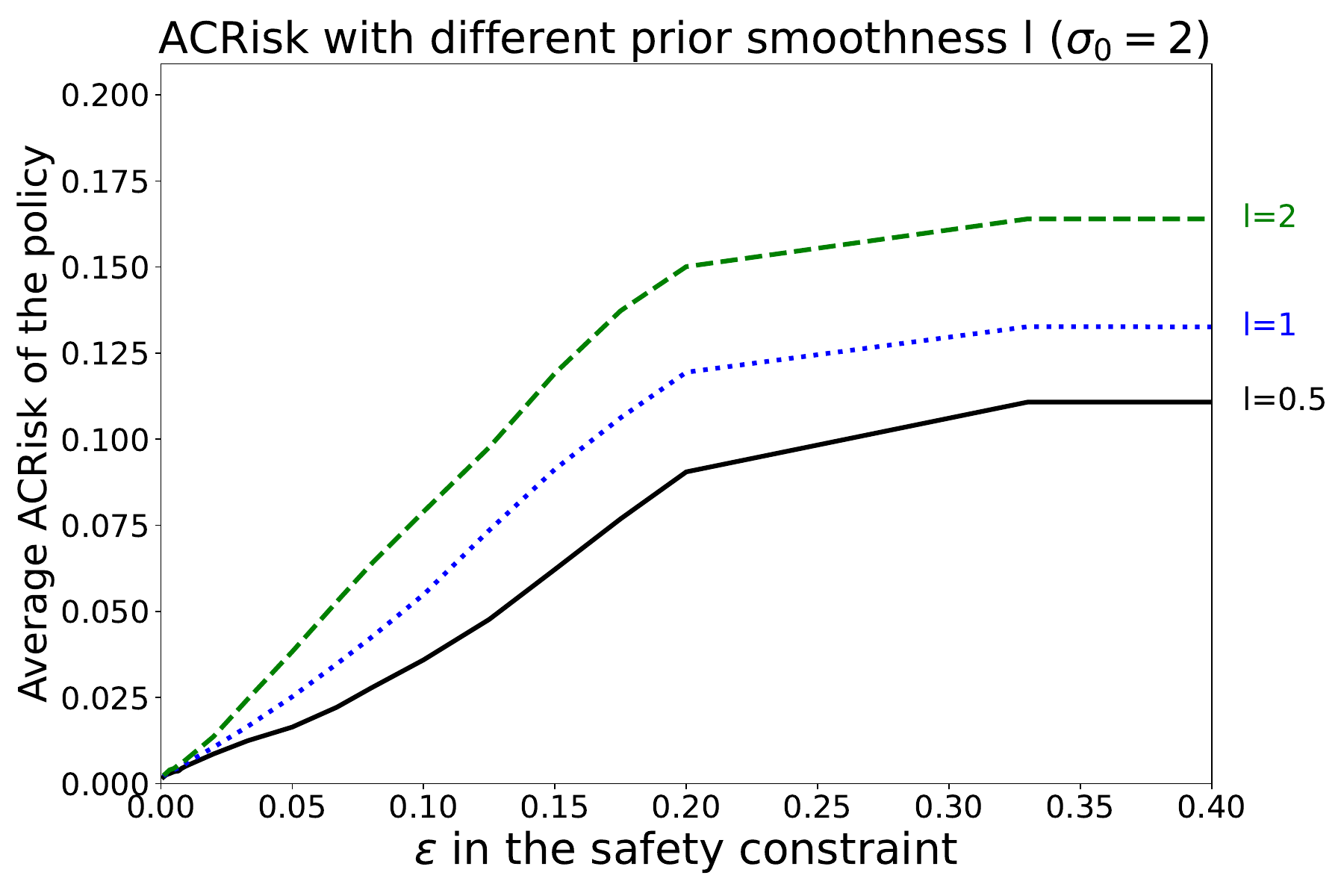}
    \caption{Average ACRisk}
    \label{fig:simus42}
  \end{subfigure}
  \caption{Average Value and ACRisk for learned policies using data without covariate overlap, varying the safety constraint and prior smoothness for the CATE.}
  \label{fig:simus4}
\end{figure}

\begin{figure}[t!]
  \centering
  \begin{subfigure}[H]{0.495\textwidth}
    \includegraphics[width=\textwidth]{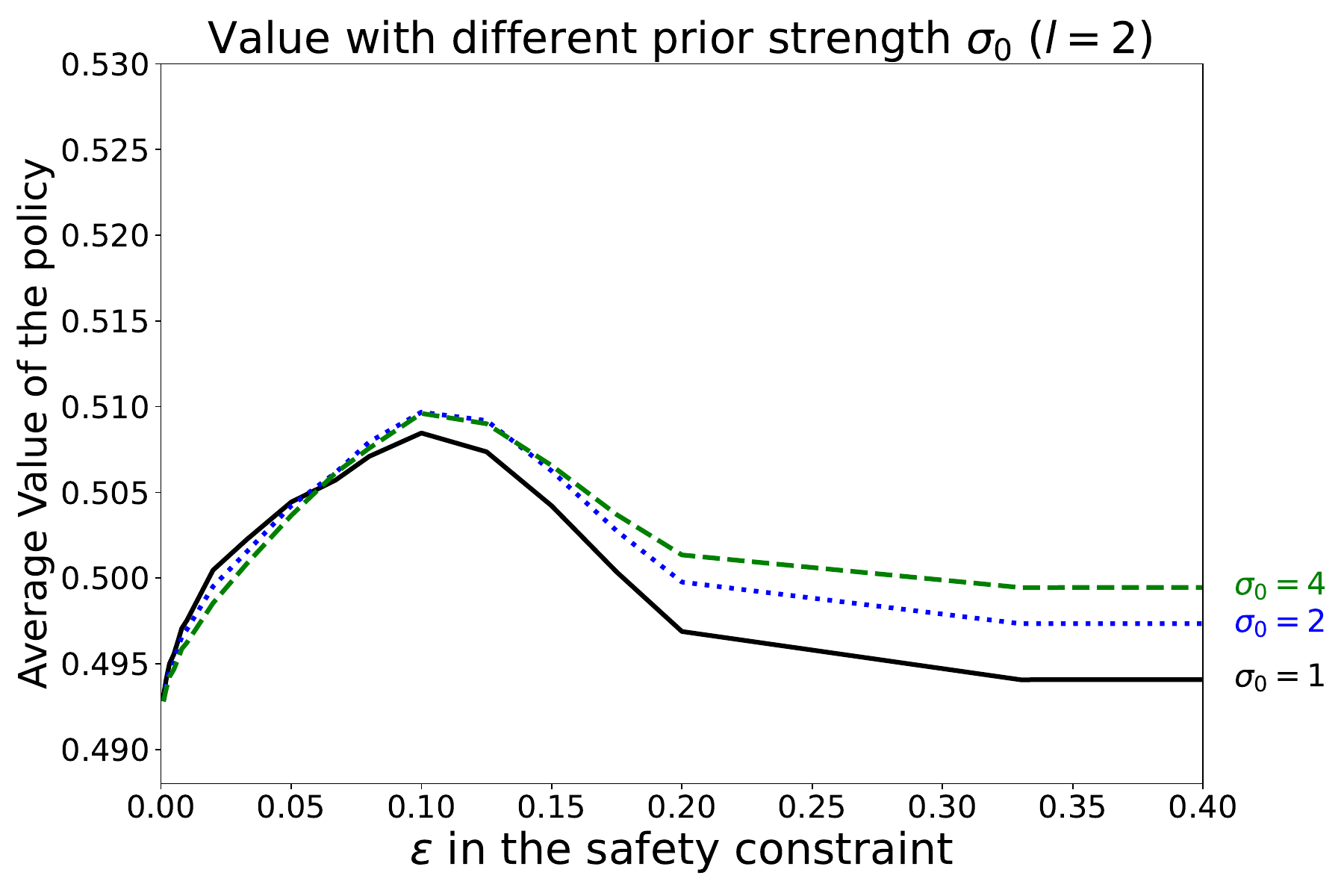}
    \caption{Average Value}
    \label{fig:simus51}
  \end{subfigure}
  \begin{subfigure}[H]{0.495\textwidth}
    \includegraphics[width=\textwidth]{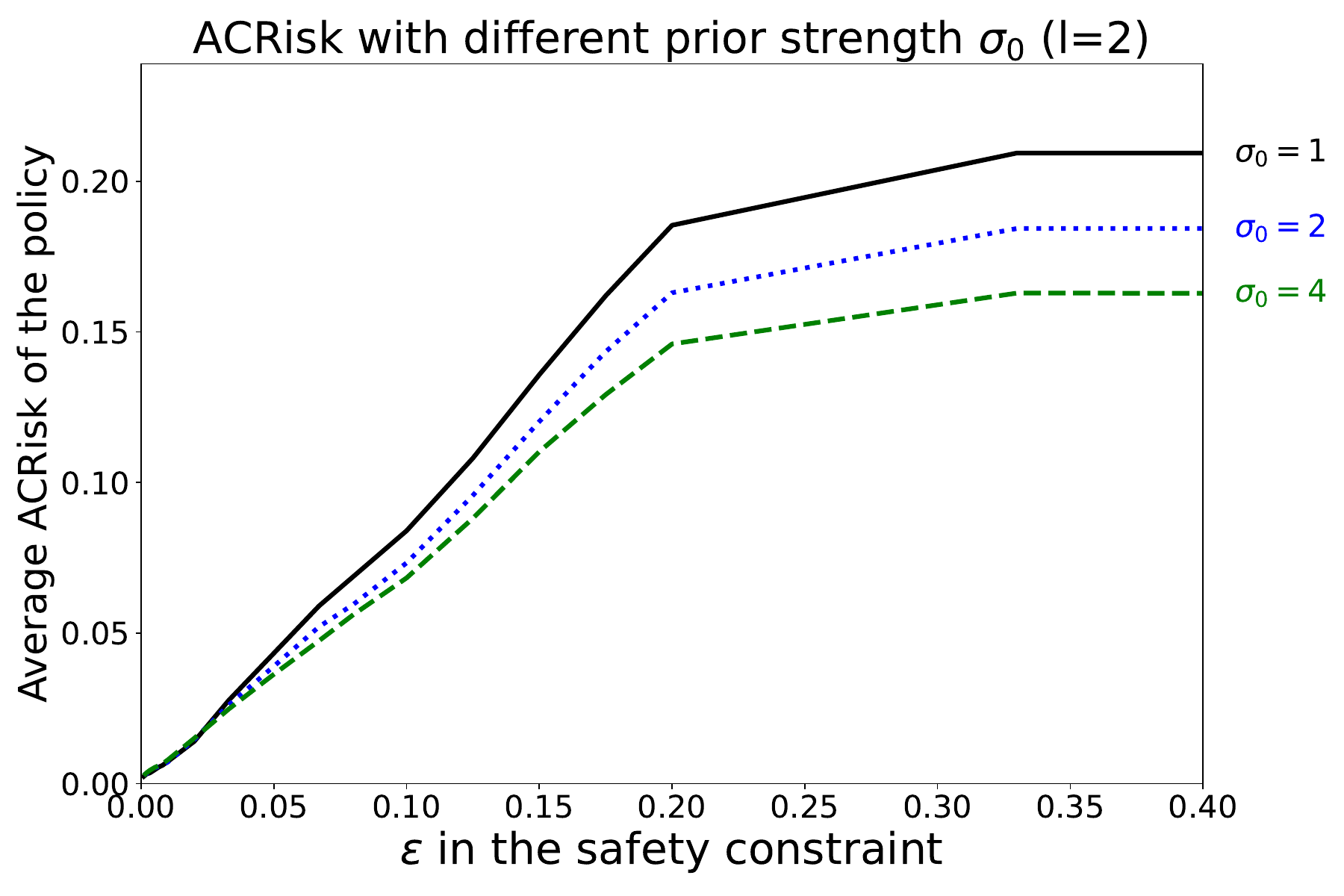}
    \caption{Average ACRisk}
    \label{fig:simus52}
  \end{subfigure}
  \caption{Average Value and ACRisk for learned policies using data without covariate overlap, varying the safety constraint and prior strength for the CATE.}
  \label{fig:simus5}
\end{figure}

\paragraph{Without covariate overlap} When there is no covariate
overlap, the CATE must be extrapolated. We fix the sample size to 200
and the signal strength $\gamma=2$ and investigate the average Value
and ACRisk for learned policies with different
priors. Figure~\ref{fig:simus4} and Figure~\ref{fig:simus5} shows the
average Value and ACRisk of the learned policy as a function of the
safety constraint $\epsilon$ under different prior smoothness and
prior strength. We find that with a smoother or stronger prior, the
learned policy leads to a greater average ACRisk as it extrapolates
more aggressively.
  
\section{Additional application results}

\subsection{Scaled PD importance of level-3 scores}
  
\begin{figure}[H]
  \begin{subfigure}[b]{0.495\textwidth}
    \includegraphics[width=\textwidth]{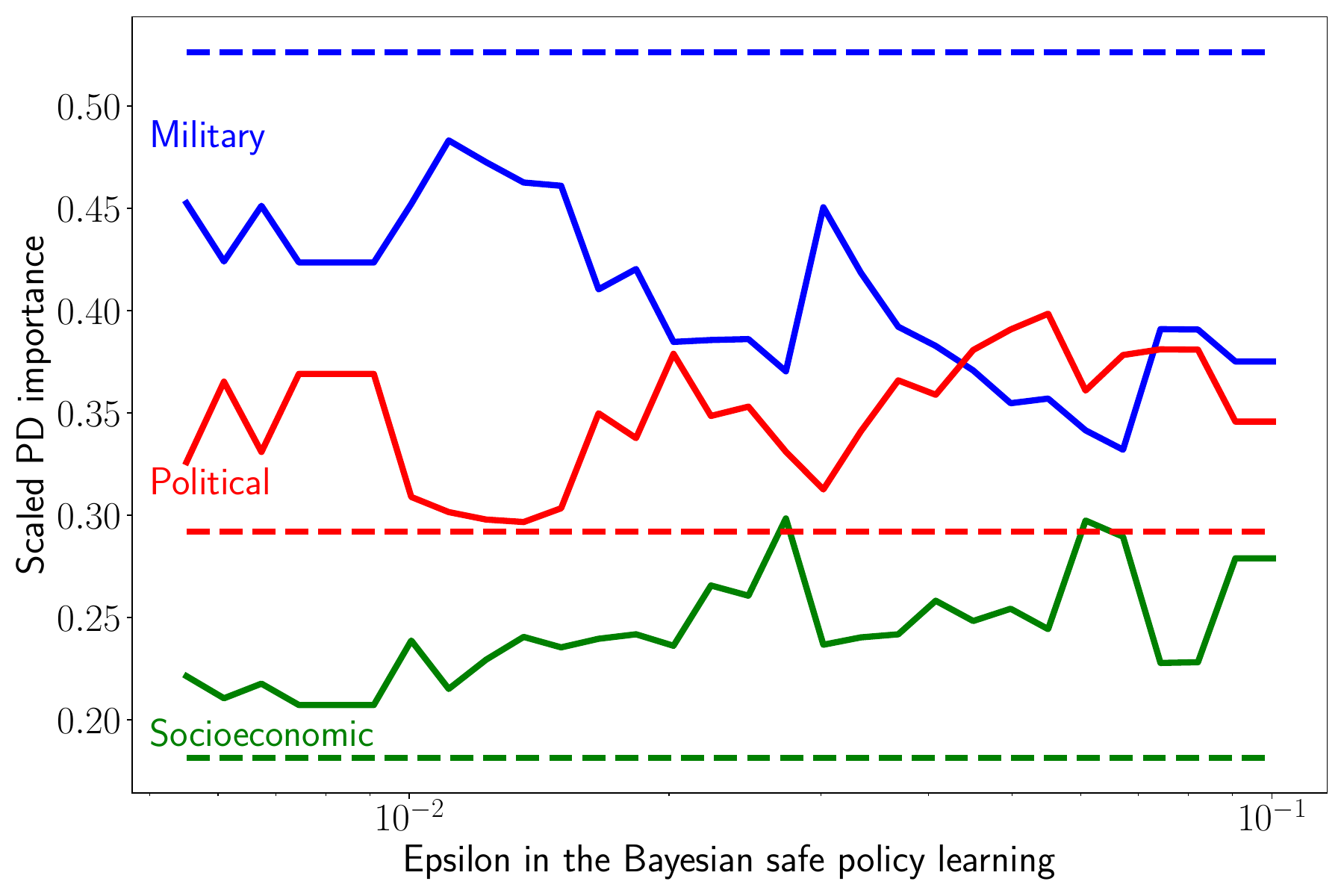}
    \caption{Regional civic society as outcome}
    \label{f817}
  \end{subfigure}   
\end{figure}

\subsection{Sensitivity analysis for prior hyperparameters}
\label{sec:smoothness}

In the main text, we use a GP to estimate the CATE, relying on the GP
prior for extrapolation. Here, we present a sensitivity analysis for
the smoothness of the GP prior, which determines the degree of
extrapolation. Other than the original setup where $l=1$, we also
consider the setups with $l=0.5$ (less extrapolation on the CATE),
$l=2$ (more extrapolation on the CATE).
  
\begin{figure}[t!]
  \centering
  \begin{subfigure}[H]{0.6\textwidth}
    \includegraphics[width=\textwidth]{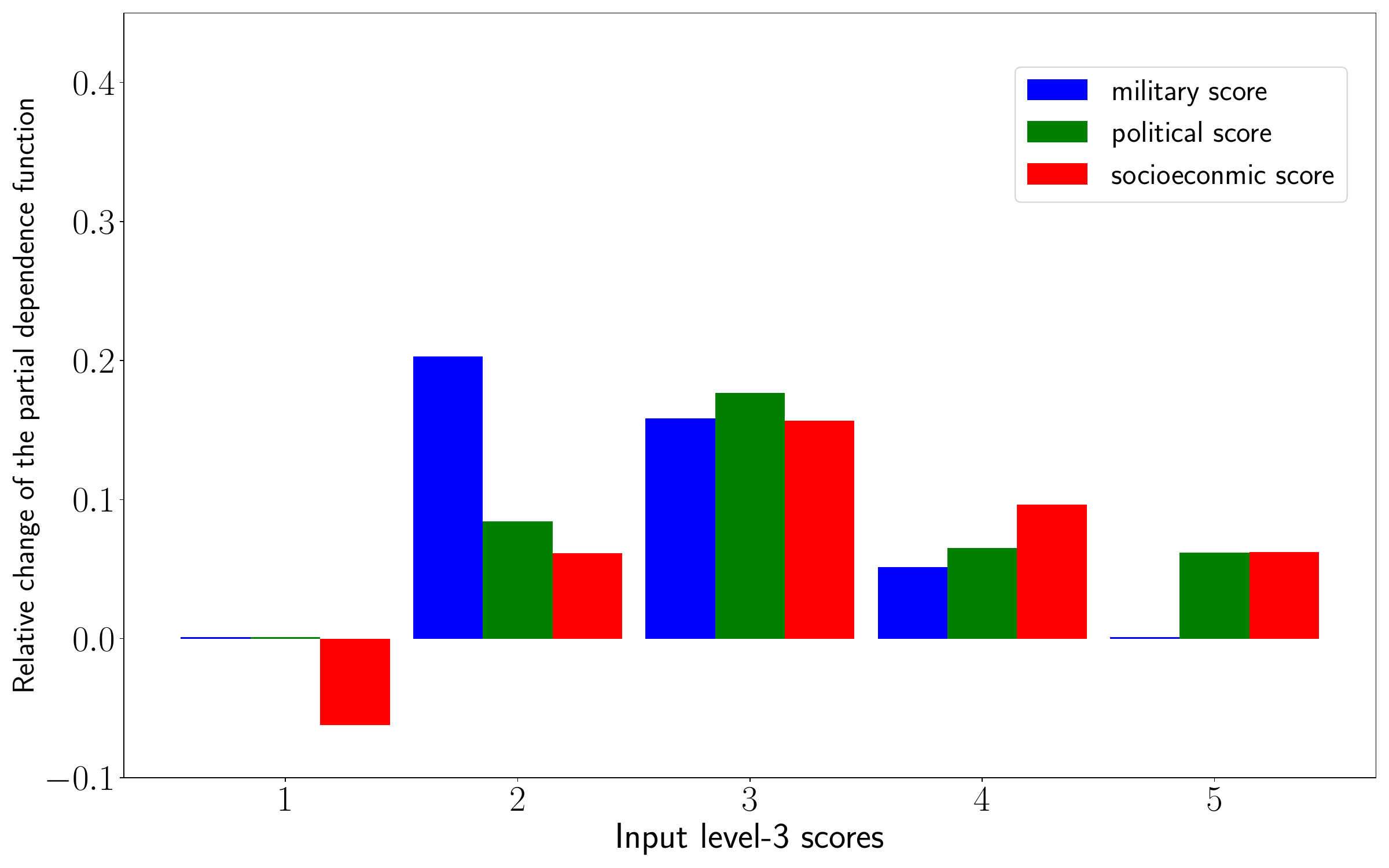}
    \caption{$l=0.5$}
  \end{subfigure}
  \begin{subfigure}[t]{0.6\textwidth}
    \includegraphics[width=\textwidth]{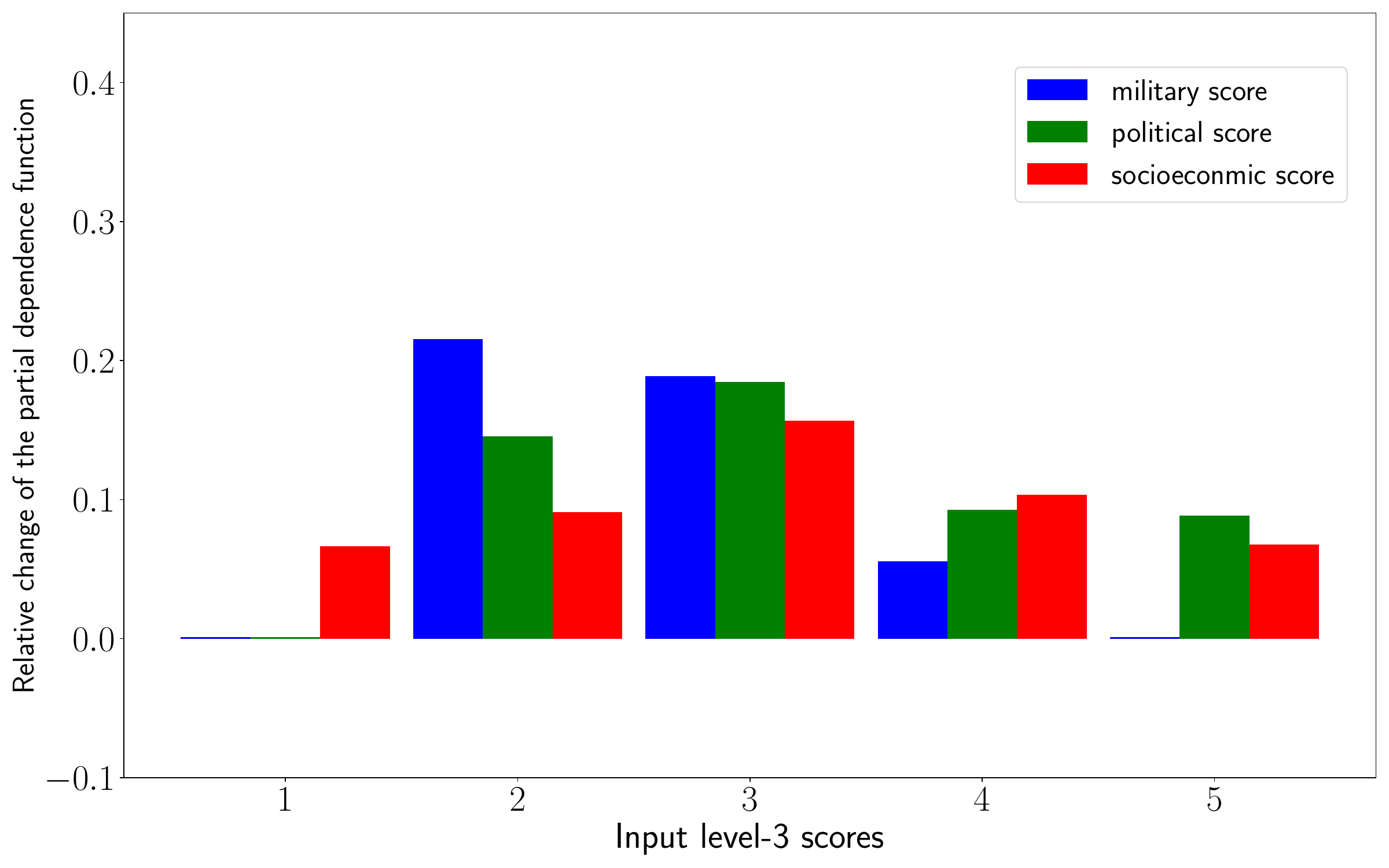}
    \caption{$l=1$}
  \end{subfigure}
  \begin{subfigure}[H]{0.6\textwidth}
    \includegraphics[width=\textwidth]{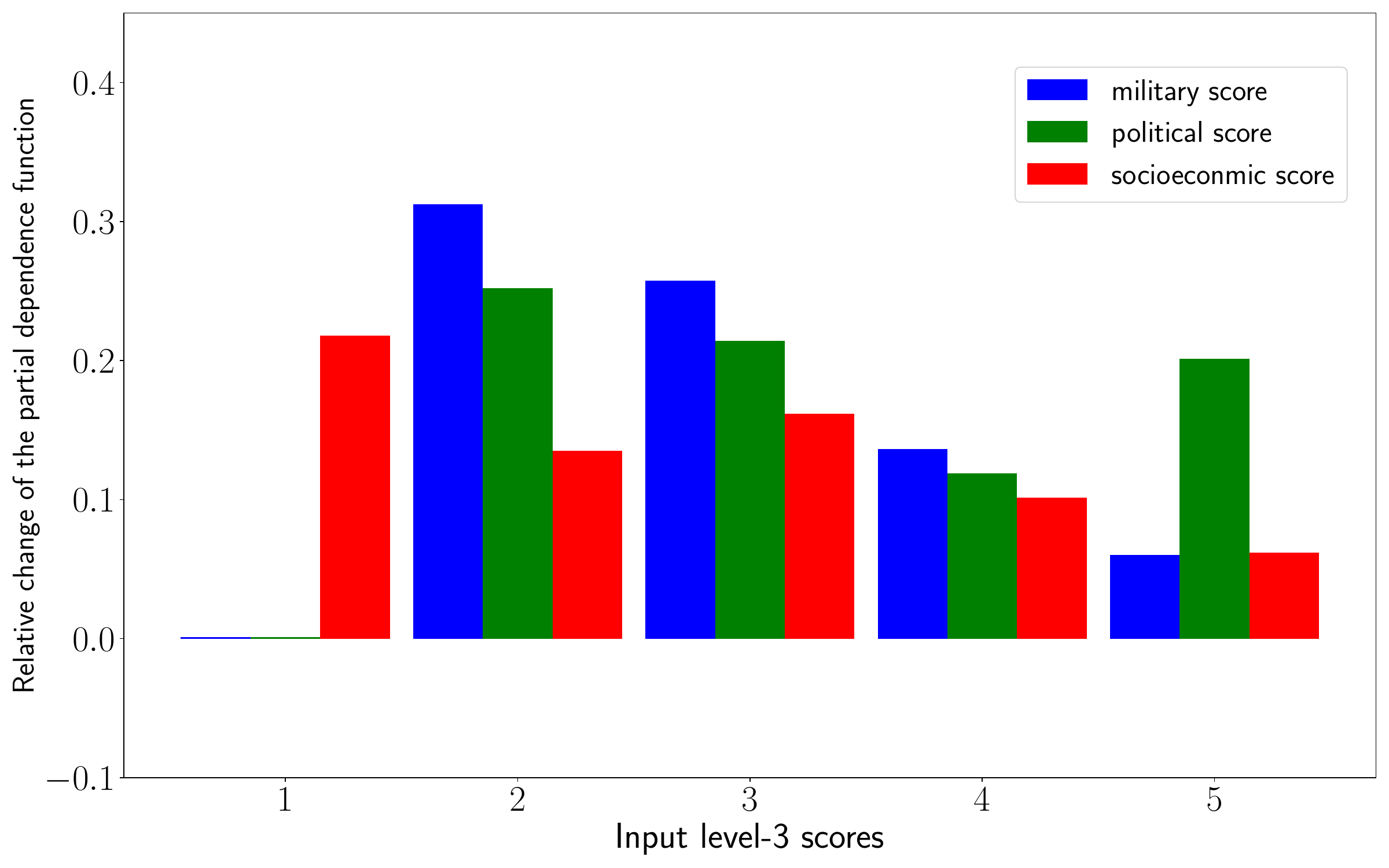}
    \caption{$l=2$}
  \end{subfigure}
  \caption{The relative change of the PD function from the baseline
    policy to the learned policy for $\epsilon = 0.1$. Each block corresponds to the
    different input of the PD function, and different colors
    corresponds to different level-3 scores. } \label{fig:partial_dependence_sensititivity} 
\end{figure}
  
In Figure~\ref{fig:partial_dependence_sensititivity}, we plot the
partial dependence of the output security score as a function of the
input level-3 score, under learned policies with different values of
the prior smoothness parameter $l$.  We use the regional safety as the
outcome for policy learning, and set the safety parameter
$\epsilon=0.1$.  As shown in
Figure~\ref{fig:partial_dependence_sensititivity}, with greater
values of the prior smoothness parameter $l$ for extrapolation, the
obtained policies tend to systematically increase the output security
score. This is consistent to the conclusion of \cite{dell2018nation}:
airstrikes increased the communist insurgency activities and decreases
regional safety. Specifically, the change of the partial dependence is
larger when $l=2$, which corresponds to a stronger smoothness prior
and more extrapolation. When $l=0.5$, there is less extrapolation and
the change of the partial dependence is smaller.
  
We further plot the relative partial dependence importance of each
level-3 scores as a function of the safety constraint $\epsilon$,
varying the prior smoothness parameter $l$. In
Figure~\ref{fig:partial_dependence_sensititivity}, under all the prior
smoothness parameter, the learned policy downweights the military
sub-model scores and upweights the socioeconomic sub-model scores.

\begin{figure}[t!]
  \begin{subfigure}[b]{0.495\textwidth}
    \includegraphics[width=\textwidth]{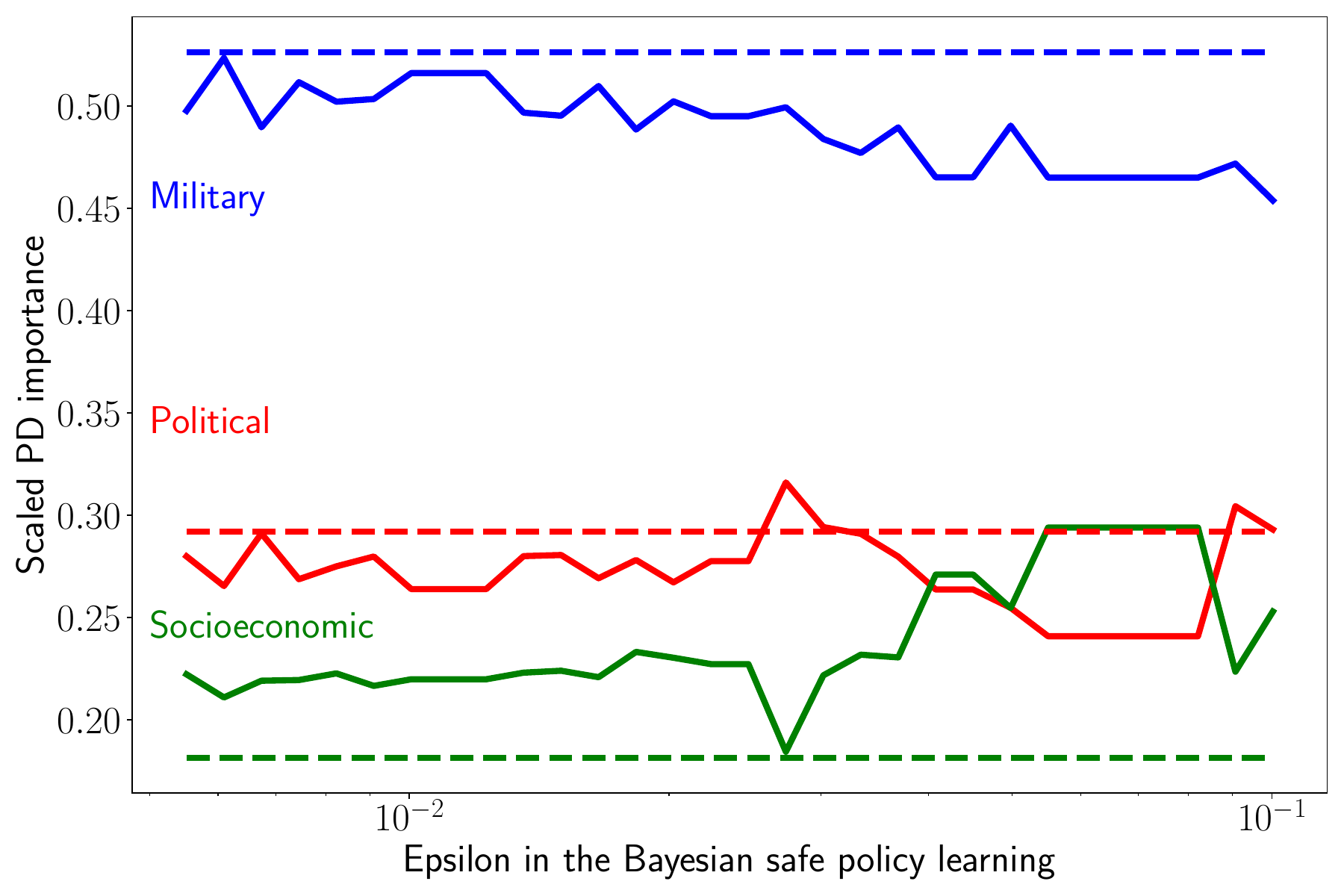}
    \caption{$l=0.5$}
  \end{subfigure}
  \begin{subfigure}[b]{0.495\textwidth}
    \includegraphics[width=\textwidth]{figs/fig1/f615.pdf}
    \caption{$l=1$}
  \end{subfigure}        
  \begin{subfigure}[b]{0.495\textwidth}
    \includegraphics[width=\textwidth]{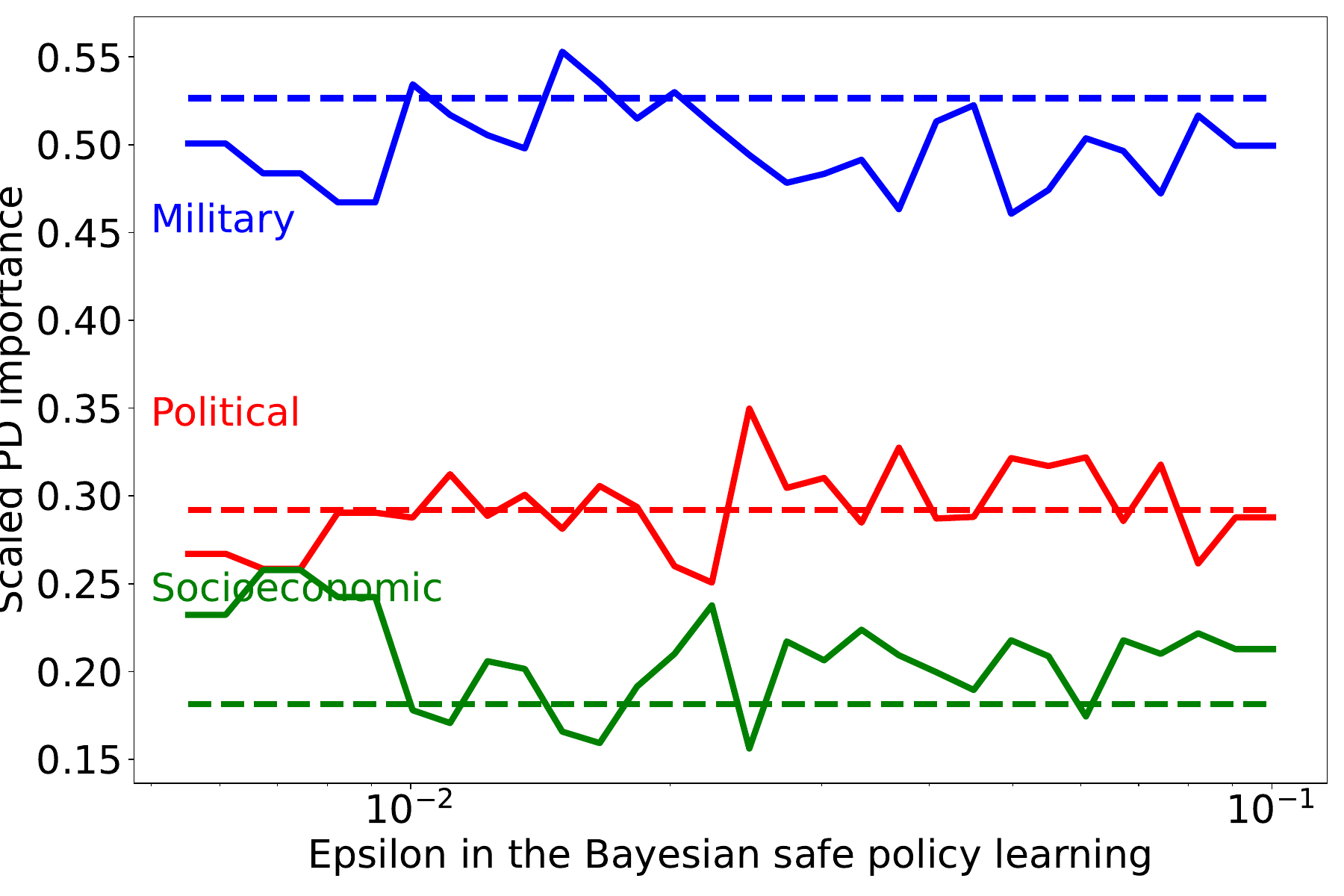}
    \caption{$l=2$}
  \end{subfigure}
  \caption{The scaled Partial Dependence (PD) importance of level-3
  scores of the learned policy with regional safety outcomes, as a function of the $\epsilon$. The
  solid line corresponds to the learned policy, and the dashed
  line indicates the baseline policy. Lines with different colors
  shows the PD importance of different level-3
  scores.  }
\label{fig:partial_dependence_sensititivity2}
\end{figure}
  
\section{Optimization for Monotonic Decision Tables}
\label{sec:opt}

In this section, we develop an optimization algorithm
applicable to monotonic decision tables where the output of a decision
table is non-decreasing in each input dimension.

\subsection{Problem definition}

We define the \textit{monotonic decision table} as below:
\begin{definition}[Monotonic decision tables] Suppose
  $\mathcal{X}_1, \mathcal{X}_2, ..., \mathcal{X}_p,\mathcal{Y}$ are
  finite totally ordered sets, $T$ is a function that maps
  $x\in \mathcal{X}=\mathcal{X}_1\times \mathcal{X}_2\times ... \times
  \mathcal{X}_p$ to $\mathcal{Y}$. Then, $T$ is a monotonic decision
  table if and only if the following condition holds,
\begin{equation*}
    \forall\ x=\{x_i\}_{i=1}^p,x'=\{x_i'\}_{i=1}^p \in \mathcal{X} \ \text{where}\ x_i \le x_i' \ \text{for}\ 1\le i\le p,\  T(x_1,...,x_p)\le T(x_1',...,x_p')
\end{equation*}
\end{definition}

We consider the following general optimization problem over monotonic
decision tables.
\begin{definition}[Optimization with monotonic decision tables]
  Suppose $f,g$ are functions that map a monotonic decision table $T$
  into a real-valued output.  We consider the problem of finding an
  optimal monotonic decision table $T_{opt}$ as defined below:
    \begin{equation}
        \begin{aligned}
            T_{opt}:= \argmin_{T} f(T) \ \text{subject to}\ g(T)\ge 0  
        \end{aligned}
        \label{def83}
    \end{equation}
\end{definition}

In general, optimization over monotonic decision tables with the form
given in Equation~\eqref{def83} is difficult to solve. Although one
could enumerate all possible monotonic decision tables and their
corresponding $f(T),g(T)$, the number of enumerations is equal to
$\mathcal{Y}^{|\mathcal{X}|}=\mathcal{Y}^{|\mathcal{X}_1|\times
  |\mathcal{X}_2|\times \dots \times |\mathcal{X}_p|}$, which grows
exponentially as the size of the decision table increases.  In our
application, we wish to simultaneously learn one two-way decision
table and one three-way decision table, yielding a total of
$5^{25}\times 5^{125}=5^{150}$ enumerations.
We would like to avoid enumerating these many possibilities.

Therefore, we use a Markov chain Monte Carlo (MCMC)-based stochastic
algorithm for optimizing over monotonic decision tables by adopting
ideas from the graph theory.  Specifically, we represent a
monotonic decision tables as an equivalent directed acyclic graph (DAG)
where the directed edges indicates the monotonicity conditions, and
optimize over monotonic decision tables by sampling the DAGs using an
MCMC algorithm.

\subsection{Graph representation for monotonic decision tables}

\begin{figure}[t]
    \centering
    \includegraphics[width=0.85\textwidth]{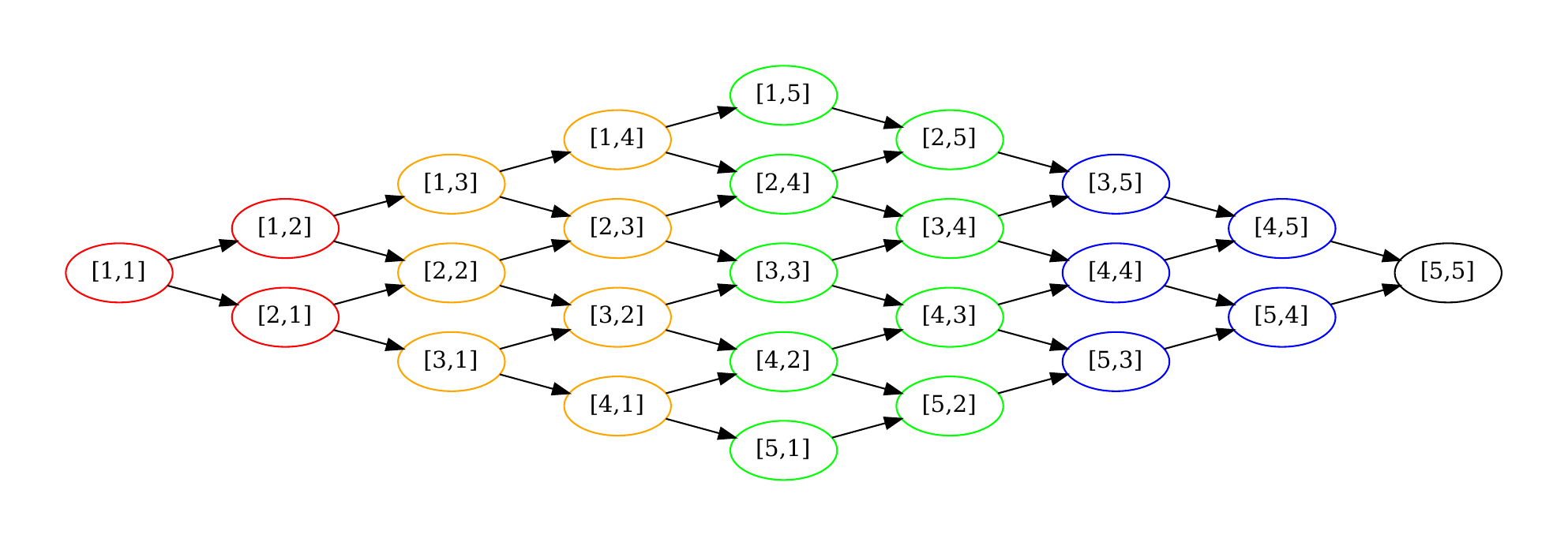}
    \includegraphics[width=0.1\textwidth]{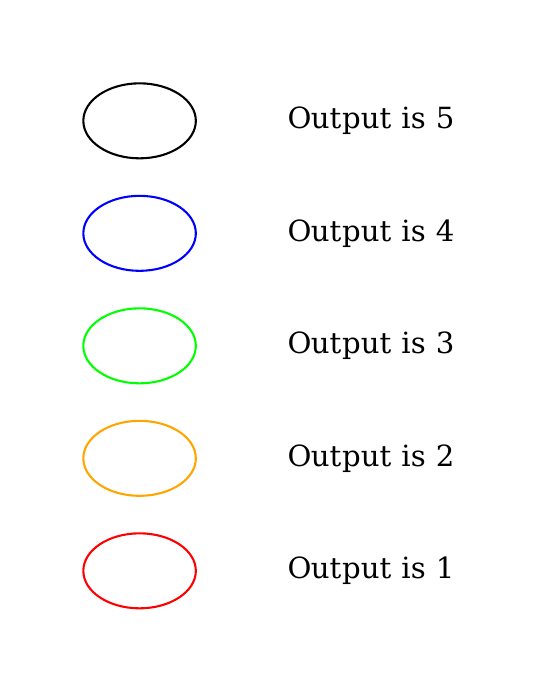}
    \caption{The DAG representation of the original 2-way decision
      table in the HES. Each vertice corresponds to an input of the
      2-way table, and directed edges indicate the relative order the
      corresponding output should satisfy based on the monotonicity
      constraint. The color of the nodes indicate the output of the
      2-way decision table in the original HES.} 
    \label{f831}
\end{figure}

Monotonic decision tables can be equivalently represented as directed
acyclic graphs (DAGs). 
We represent different inputs of
decision tables as vertices of the DAG, and the monotonicity
constraint on the decision table as directed edges in the graph.  For
example, Figure~\ref{f831} shows the graph representation for the
original two-way decision tables in the HES (Figure~\ref{fig:hes_table}). The
two-way decision table has $5\times 5=25$ different inputs, which
corresponds to the $25$ vertices in the graph. The edges in the DAG
indicate the monotonicity constraint that the output should satisfy
according to the decision table. For example, in Figure~\ref{f831},
there is a directed edge from vertex $[1,1]$ to vertex $[1,2]$, which means
that the output of the decision table for input $[1,1]$ should be
no greater than the output for input $[1,2]$.

With this representation, finding a decision table from $\mathcal{X}$
to $\mathcal{Y}$ is equivalent to finding a graph partition that
separate the DAG into $|\mathcal{Y}|$ areas where vertices in the same
area have the same output scores.  A monotonicity constraint on
the decision table is equivalently translated to an \textit{acyclic}
constraint on the partition, where we forbid any directed edges from
a node with larger output to aa node with a smaller outcome
\citep{herrmann2017acyclic,doi:10.1137/18M1176865}.

\begin{figure}[t]
    \includegraphics[width=0.85\textwidth]{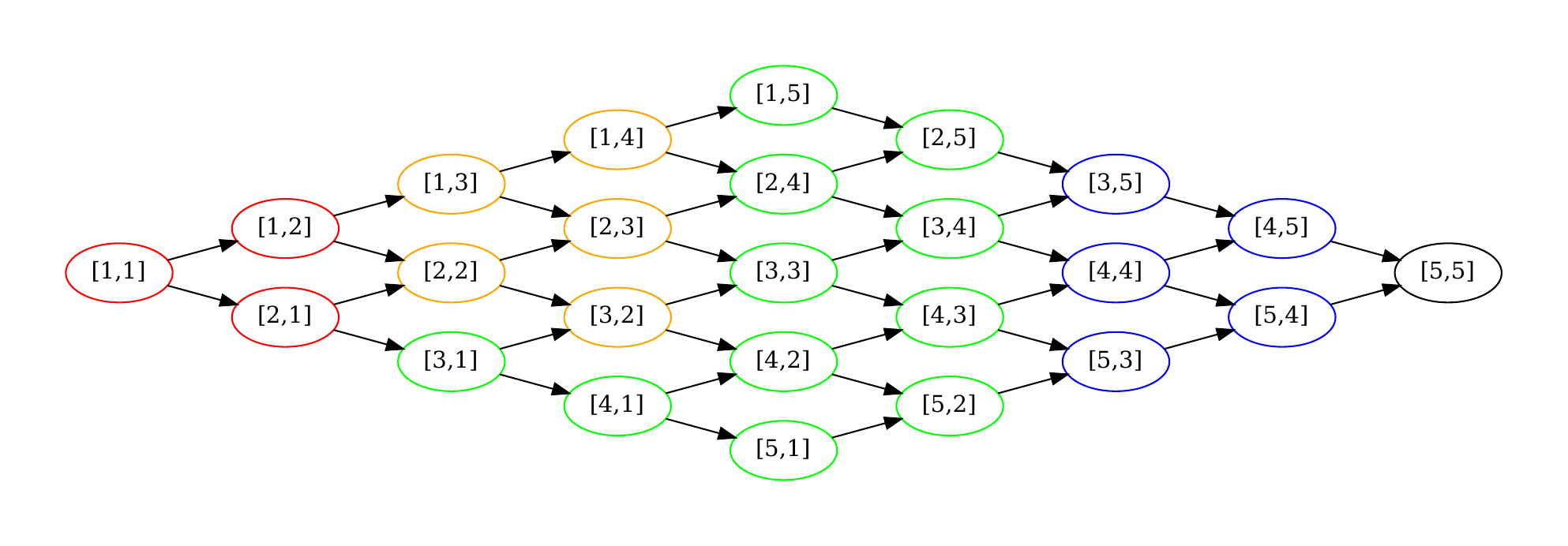}
    \includegraphics[width=0.1\textwidth]{figs/legend2.pdf}
    \caption{An example of partition for a decision table that does
      not satisfy the monotonicity constraint. There is a directed
      edge from $[3,1]$ to $[3,2]$, but the output for $[3,1]$ is
      larger than the output for $[3,2]$.}
    \label{f832}
\end{figure}

Figure~\ref{f831} shows the graph partition corresponding to
the original two-way decision table in the HES, where the color
of a node indicates the partition.  This satisfies the
monotonicity constraint. In contrast, Figure~\ref{f832} shows a
different partition that does not satisfy the monotonicity constraint.
There is a directed edge from $[3,1]$ to $[3,2]$, but the output for
$[3,1]$ is larger than the output for $[3,2]$.

\subsection{Optimization by sampling partitions of the DAGs}

We optimize over monotonic decision tables by finding optimal acyclic
partitions of the corresponding DAGs.  We propose an MCMC-based
stochastic algorithm for sampling acyclic partitions of the DAGs and
optimize over it.  To do this, we first sample a \textit{topological
  sort} of a DAG \citep{karzanov1990conductance} and then segment the
topological sort into $|\mathcal{Y}|$ pieces to obtain a partition of
the DAG.
\begin{definition} \spacingset{1}
  (Topological sort on DAG) A topological sort of a directed acyclic
  graph (DAG) is a linear ordering of its vertices such that for every
  directed edge $uv$ from vertex $u$ to vertex $v$, $u$ comes before
  $v$ in the ordering.
\end{definition}

We take advantage of the fact that any segmentation of a topological
sort corresponds to an acyclic partition and produces a decision table
satisfying the monotonicity constraint. Conversely, for any decision
table satisfying the monotonicity constraint, there will be a
corresponding topological sort and segmentation that produces the
decision table.  We use the MCMC algorithm of
\cite{karzanov1990conductance} to sample a topological sort from the
DAG and a random walk to sample segmentations of the topological
sorts.  This gives us a Markov chain that generates random acyclic
partitions.  Finally, we use a \textit{short-burst} algorithm to use
the MCMC sampling algorithm for optimization \citep{cannon2023voting}.
Algorithm~\ref{alg1} formally describes our optimization procedure.

\begin{algorithm}[t]
  \spacingset{1}
    \caption{Stochastic Optimization Algorithm for Monotonic Decision Tables}\label{alg1}
    \KwData{DAG $\mathcal{G}$ that contains all the potential inputs of the decision table and relations from monotonicity constraints; Markov chain $\mathcal{M}$ that generates an acyclic partial partition of $\mathcal{G}$ with a uniform stationary distribution; Initial monotonic decision table $T_0$ and corresponding state $S_0$ of $\mathcal{M}$.} 
    \For{$ 0\le r\le R$}{
      $S_{r0} \gets S_r$\\
      $T_{r0} \gets T_r$\\
      \For{$1\le k \le K$}{
        $S_{rk} \gets \mathcal{M}(S_{r(k-1)})$\\
        $T_{rk} \gets$ Decision table that corresponds to $S_{rk}$\\
      }
      $S_{r+1}\gets \arg \min_{{T\in\{T_{r0},T_{r1},...,T_{rK}\}}} f(T)$ subject to $g(T)\ge 0$\\
      $T_{r+1}\gets$ Decision table corresponds to $S_{r+1}$;
    }
     Return $T_{R+1}$      
    \end{algorithm}
 
In the application, we run the above algorithm 2000 times to obtain 2000 corresponding decision tables. Then we choose the one that achieves the highest posterior expected value.

\bigskip

\bibliography{BayesSafe}